\documentclass[fleqn,10pt]{wlscirep}
\usepackage[utf8]{inputenc}
\usepackage[T1]{fontenc}
\usepackage{svg}
\usepackage{graphicx}
\usepackage{algorithm}
\usepackage{hyperref}
\usepackage{caption}
\usepackage{subcaption}
\usepackage{algpseudocode}
\usepackage{amsmath}
\usepackage{subcaption}
\usepackage{csvsimple}
\usepackage{multirow} 
\usepackage{tabularx}
\usepackage{xr}
\usepackage{marvosym}  
\usepackage{array}
\usepackage{tikz}
\usepackage{wrapfig}
\usepackage{xcolor}
\usepackage{listings}
\usepackage{hyperref}
\usepackage{xcolor}

\definecolor{lightgay}{HTML}{d3d3d3}
\definecolor{green}{HTML}{66c2a5}
\definecolor{purple}{HTML}{8da0cb}

\title{
MetaMP: Seamless Metadata Enrichment and AI Application Framework for Enhanced Membrane Protein Visualization and Analysis
}

\author[1,*]{Ebenezer Awotoro}
\author[1]{Chisom Ezekannagha}
\author[2]{Florian Schwarz}
\author[2]{Johannes Tauscher}
\author[2,3]{Dominik Heider}
\author[1]{Katharina Ladewig}
\author[4]{Christel Le Bon}
\author[4]{Karine Moncoq}
\author[4]{Bruno Miroux}
\author[1,5,*]{Georges Hattab}
\affil[1]{Center for Artificial Intelligence in Public Health Research (ZKI-PH), Robert Koch Institute, Berlin, 13353, Germany}
\affil[2]{Department of Mathematics and Computer Science, University of Marburg, Marburg, Germany}
\affil[3]{Institute of Medical Informatics, University of Münster, Münster, Germany}
\affil[4]{Université Paris Cité, Centre National de la Recherche Scientifique (CNRS), Biochimie des Protéines Membranaires, UMR7099, Paris, France}
\affil[5] {Department of Mathematics and Computer Science, Freie Universität Berlin, Berlin, 14195, Germany}

\affil[*]{awotoroe@rki.de}


\begin{abstract}
Structural biology has made significant progress in determining membrane proteins, leading to a remarkable increase in the number of available structures in dedicated databases.
The inherent complexity of membrane protein structures, coupled with challenges such as missing data, inconsistencies, and computational barriers from disparate sources, underscores the need for improved database integration.
To address this gap, we present MetaMP, a framework that unifies membrane-protein databases within a web application and uses machine learning for classification. 
MetaMP improves data quality by enriching metadata, offering a user-friendly interface, and providing eight interactive views for streamlined exploration. 
MetaMP was effective across tasks of varying difficulty, demonstrating advantages across different levels without compromising speed or accuracy, according to user evaluations. Moreover, MetaMP supports essential functions such as structure classification and outlier detection.

We present three practical applications of Artificial Intelligence (AI) in membrane protein research: predicting transmembrane segments, reconciling legacy databases, and classifying structures with explainable AI support.
In a validation focused on statistics, MetaMP resolved 77\% of data discrepancies and accurately predicted the class of newly identified membrane proteins 98\% of the time and overtook expert curation.
Altogether, MetaMP is a much~---~needed resource that harmonizes current knowledge and empowers AI-driven exploration of membrane-protein architecture.

\end{abstract}
\begin{document}
\flushbottom
\maketitle
%
%


\section*{Introduction}

Membrane Proteins (MPs) are essential components of cells, involved in various biological processes, and the target of over 50\% of modern medicinal drugs~\cite{aguayo2021multiscale, errey2020production}.  
Membrane proteins are defined as proteins that are associated with or attached to the cellular membranes of cells or organelles. They can be classified into two main categories: integral (or transmembrane) proteins, which are permanently embedded in the lipid bilayer and often span the membrane one or multiple times, and peripheral proteins, which are temporarily associated with the membrane surface or with integral proteins without spanning the bilayer themselves ~\cite{alberts2002membrane, MPcubeBiotech}. 
These proteins perform a wide range of functions, including acting as receptors, enzymes, and transporters, and are crucial for processes such as signal transduction and cell communication~\cite{MPcubeBiotech, sun2023machine}.
The structural biology of MPs has advanced significantly in the past decade, with breakthroughs in purification techniques and structure determination methods~\cite{yao2020cryo} leading to an exponential increase in the number of MP structures deposited in databases such as the Membrane Proteins of known 3D Structure (MPstruc)~\cite{newport2019memprotmd} database.
Since the determination of the first membrane protein structure in 1985~\cite{li2021highlighting}, over 1,700 unique MP structures have been resolved, providing crucial molecular insights into MP function. 
This is observed in the crystal structure list of White~\cite{White_Membrane}.
Despite significant advancements in X-ray crystallography~\cite{kermani2021guide}, NMR~\cite{reif2021solid, hu2021nmr}, and electron microscopy~\cite{yip2020atomic}, as well as improvements in MP production and stabilization~\cite{andrell2013overexpression}, MP structural biology remains challenged by the difficulties in producing and purifying recombinant proteins in a functional state~\cite{hattab2014membrane,hattab2015escherichia}, limiting study efficiency and reproducibility. Addressing this requires broader adoption of standardized, reliable methods for structure determination.

However, data-related issues, such as missing data, inconsistencies in data collection and processing, and the presence of pending MP structures, make the complex nature of membrane protein structure databases a daunting challenge. 
Computational barriers arise from the use of multiple data sources with different information and metadata, requiring pre-processing techniques such as removing sparse data (highly empty columns) to ensure data quality and consistency. 
While current efforts to maintain membrane protein-related databases are commendable and biologists see them as a much-needed resource, the landscape is not accurate enough to perform machine learning experiments.
Indeed, machine learning methods cannot be applied out of the box to data exported from current databases.
Our rationale is to build a database for the seamless use of machine learning methods and visualization techniques for the benefit of the membrane protein community.

In recent years, related work has focused on the evaluation and validation of various MP databases such as MPstruc, OPM, TCDB (Transporter Classification Database), and PDBTM (Protein Data Bank of Transmembrane Proteins)~\cite{choy202110, aleksandrova2024encompass, dobson2024unitmp, tsirigos2010ompdb, newport2019memprotmd}. 
Several database curators and providers are working to ensure that each membrane protein entry in these databases remains consistent, stable, and accurate.
A comparative analysis was performed on multiple MP structure databases, including MPstruc, OPM, and PDBTM. 
The study aimed to assess the degree of overlap and consistency in structural and functional classifications, as well as the assignment of transmembrane domains across these databases.
The study revealed significant differences in database coverage, protein annotation criteria, and classification~\cite{aleksandrova2024encompass}. 
A noteworthy mention is UniTmp~\cite{dobson2024unitmp} which offers a tailored solution for transmembrane protein (TMP) research by integrating various databases such as Topology Data Bank of Transmembrane Proteins (TOPDB)~\cite{dobson2015expediting}, database of conservatively located domains and motifs in proteins (TOPDOM)~\cite{varga2016topdom}, Protein Data Bank of Transmembrane Proteins (PDBTM)~\cite{kozma2012pdbtm}, and Human Transmembrane Proteome (HTP)~\cite{dobson2015human}. 
This integration provides a unified view of TMPs, facilitating the exploration of protein structure, topology, post-translational modifications, and linear motifs. 
However, UniTmp focuses specifically on structural aspects of transmembrane proteins with very limited metadata and currently has no automated update system for database synchronization.

To address these challenges and empower the membrane protein research community, we propose MetaMP, a web application designed to dynamically curate structure determination metadata for resolved MPs. 
MetaMP generates a continuously updated dataset containing rich information, including structure determination methods, taxonomic domains, expression systems, and more. 
This web application emphasizes the importance of spatial, topological, and functional annotations for each MP and serves as a critical and novel resource for researchers.

MetaMP uses a three-tiered approach to efficiently integrate metadata from MPstruc~\cite{White_Membrane}, RCSB PDB~\cite{burley2019rcsb}, OPM~\cite{lomize2012opm}, and UniProt~\cite{uniprot2023uniprot}.
At the data layer, MetaMP leverages these databases for enrichment, ensuring that the manually curated MPstruc database serves as the source for PDB accession IDs and categorical attributes such as groups and subgroups. 
The application layer uses state-of-the-art technologies to process and consolidate the integrated data, while the presentation layer provides a user-friendly interface with a landing page that features eight different views.

By integrating and monitoring disparate data from multiple MP databases, MetaMP establishes a comprehensive resource for the membrane protein research community. 
MetaMP's interactive visualizations and machine learning capabilities empower experts to identify patterns, trends, and correlations across experimental and functional data. 
Its effectiveness has been validated via AI use cases and user evaluation, demonstrating benefits in improving performance and assisting experts in classifying structures, detecting outliers, and providing a data-rich mosaic of what is usually a fragmented outlook.



\begin{table}[htbp]
\centering
\caption{
\textbf{Proportional contribution of each dedicated protein database to MetaMP.}
Number of observations or membrane protein structures, nominal and quantitative attributes or features are reported. The increase in attribute number and diversity in MetaMP marks a key advancement for membrane protein research.
}
\begin{tabular}{lcccc}
\toprule
\textbf{Database} & \textbf{Rows/Observations/MPs} & \textbf{Attributes/Features} & \textbf{Nominal} & \textbf{Quantitative} \\
\midrule
MPstruc & 3795 & 10 & 10 & 0 \\
PDB & 3569 & 228 & 92 & 136 \\
OPM & 2966 & 27 & 19 & 8 \\
UniProt & 3425 & 36 & 34 & 2 \\ 
\midrule
\bf MetaMP & 3569 & 301 & 155 & 146 \\
\bottomrule
\end{tabular}
\label{tab:databases}
\end{table}

\clearpage
\section*{Results}

\begin{wrapfigure}{R}{0.55\textwidth}
\vspace*{-6mm}

\centering
\begin{tikzpicture}
\def\radius{1.8cm}
\def\shift{1.3*\radius}

\coordinate (ceni);
\coordinate[xshift=\shift] (cenii);
\coordinate[yshift=-\shift] (ceniii);
\coordinate[xshift=\shift, yshift=-\shift] (ceniv);

\draw[fill=white, opacity=.3] (ceni) circle (\radius);
\draw[fill=green, opacity=.5] (cenii) circle (\radius);
\draw[fill=purple, opacity=.3] (ceniii) circle (\radius);
\draw[fill=lightgray, opacity=.3] (ceniv) circle (\radius);

\node[xshift=-0.5*\radius, yshift=0.5*\radius] at (ceni) {MPstruc};
\node[xshift=0.5*\radius, yshift=0.5*\radius] at (cenii) {OPM};
\node[xshift=-0.5*\radius, yshift=-0.5*\radius] at (ceniii) {UniProt};
\node[xshift=0.5*\radius, yshift=-0.5*\radius] at (ceniv) {PDB};

\node at (ceni) {10};
\node at (cenii) {27};
\node at (ceniii) {36};
\node at (ceniv) {228};

\node at (barycentric cs:ceni=1,cenii=1,ceniii=1,ceniv=1) {1};
\node at (barycentric cs:ceniii=1,ceniv=1) {2};
\node at (barycentric cs:ceni=1,ceniii=1) {1};
\node at (barycentric cs:ceni=1,cenii=1) {3};
\node at (barycentric cs:cenii=1,ceniv=1) {2};

\end{tikzpicture}
\caption{
\textbf{Venn diagram of the four dedicated protein databases integrated in MetaMP.}
The numbers inside each circle represent the total attributes for each database integrated in MetaMP. 
The number in the middle (1) represents the common attribute shared by all four sources: the PDB accession code or (\texttt{pdb\_code}).
The diagram visually shows that PDB has the most attributes (228), followed by UniProt (36), OPM (27), and MPstruc (11).
}
\label{figure:venn}
\end{wrapfigure}
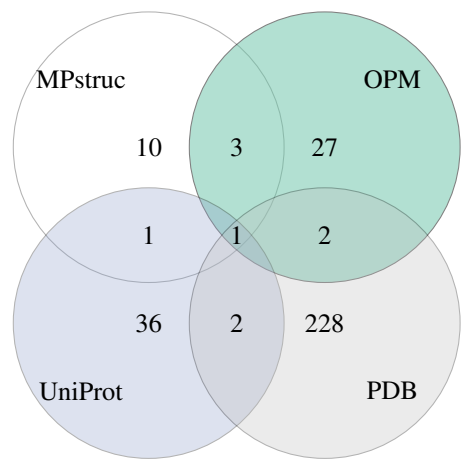


This section begins by showcasing MetaMP’s real-world impact with two key use cases~---~Legacy Database Reconciliation and High-Throughput Screening \& Predictor Benchmarking. It then provides a comprehensive overview of our findings, organized into seven thematic areas: 
(1) Database Overview, 
(2) Artificial Intelligence Use Cases, (3) Eight Interactive Views, 
(4) improvement on Cryo-Electron Microscopy, 
(5) Geographic Distribution of Research Contributions,
(6) Quality Control, with a focus on outlier detection and data‐discrepancy resolution,
and (7) Task-Oriented User Evaluation, combining quantitative performance metrics with qualitative feedback.

\subsection*{Database Overview} 
The initial release of the MetaMP web application is subject to version control and comprehensive documentation. 
The corresponding MetaMP database contains 3,569 entries of MP structures out of 3,795. 
This comprehensive collection was created by selectively combining data from four source databases: MPstruc, PDB, OPM, and UniProt.

\begin{figure}[htbp]
\centering
\includegraphics[width=\textwidth]{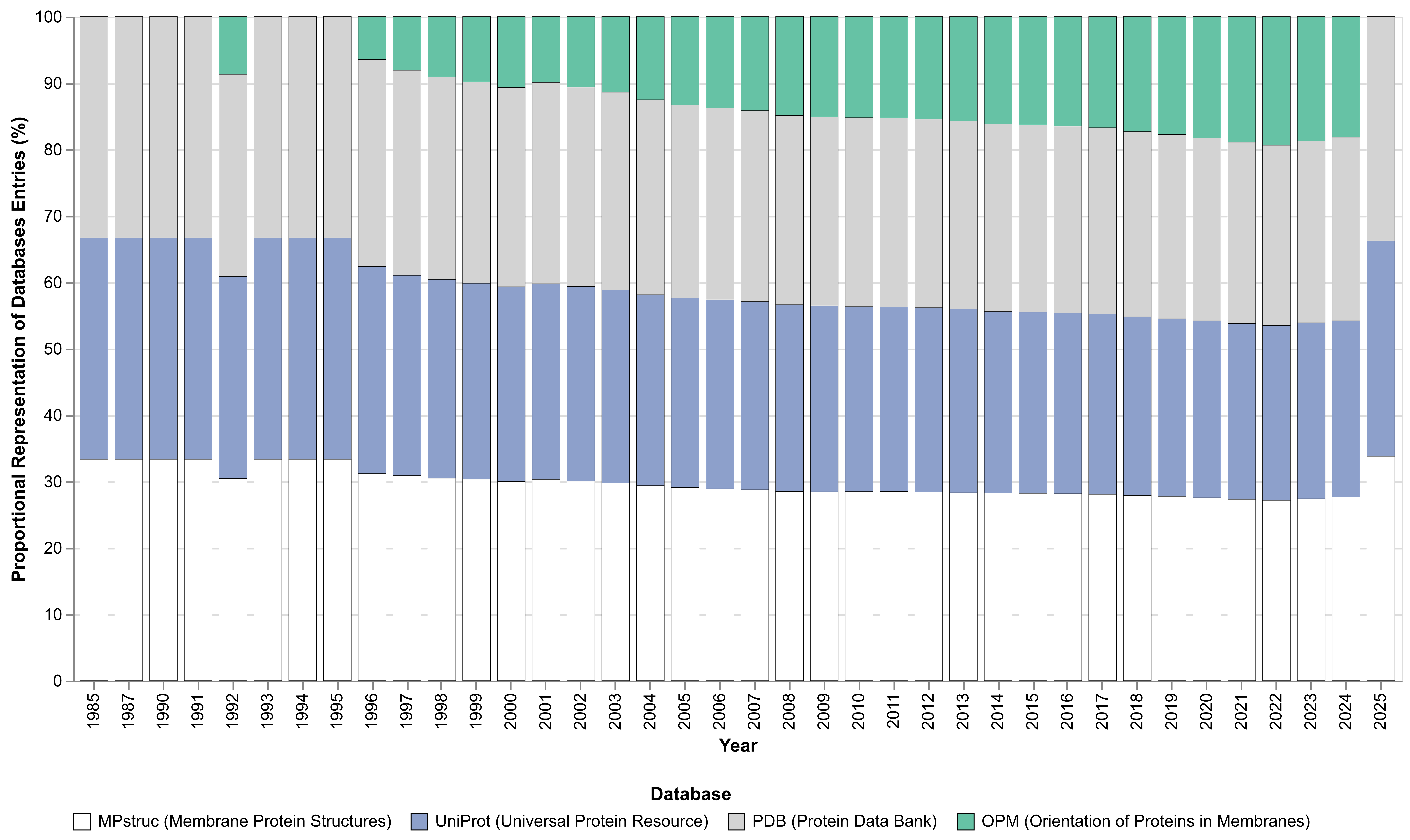}
\caption{
\textbf{
Comparative Annual Representation of Membrane Protein Entries from the PDB, OPM UniProt, and MPstruc Databases.}
The chart shows how the proportion of data for membrane protein entries has changed. 
The same number of entries in two databases corresponds to the same size bars.
For example, in 2025, MPstruc, UniProt, and PDB databases have equal bar sizes, indicating the same number of entries, while OPM shows none.
}
\label{fig:various-databases}
\end{figure}

\clearpage 
Table~\ref{tab:databases} shows the proportional contribution of each database to MetaMP.
While Figure~\ref{figure:venn} shows the attribute overlap of the four source protein databases integrated into MetaMP, Figure~\ref{fig:various-databases} showcases the proportional representation of MP entries from each database over time. 
The MetaMP database release excludes entries that are under review or embargoed in the PDB database.
This is currently the case for two entries with the identifiers 7ROW and 7UUV. 

\begin{figure}
    \centering
\includegraphics[width=1\textwidth]
{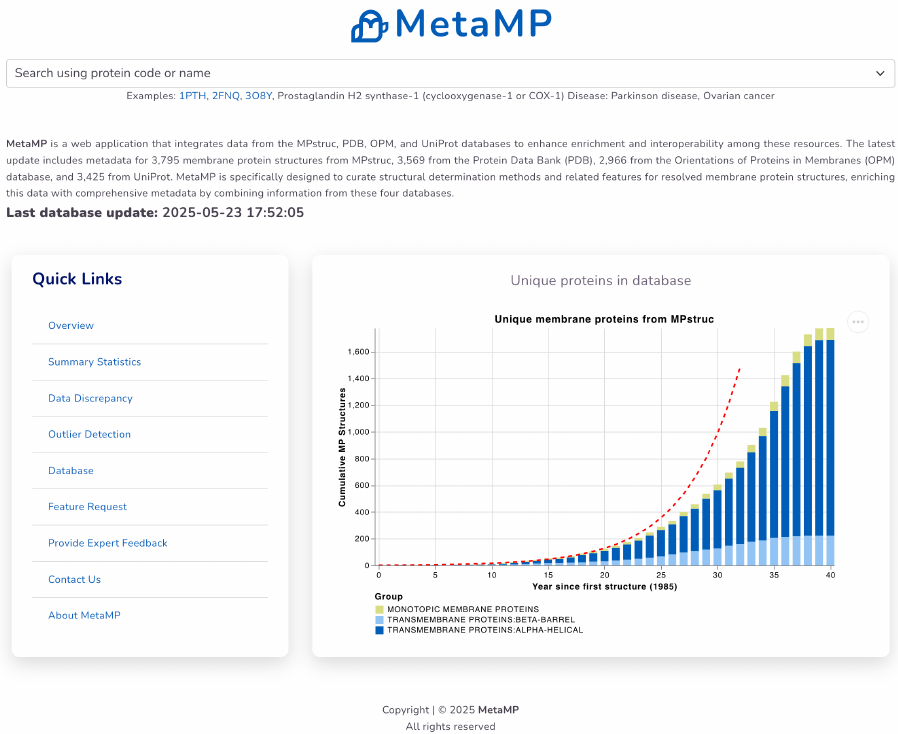}
\caption{\textbf{MetaMP Landing page}. Featuring a search field and eight distinct views. MetaMP offers context-focused views to support experts in their tasks: Overview, Summary Statistics, Data Discrepancy, Outlier Detection, Database, Exploration, and Grouping. These views provide a comprehensive perspective on membrane protein (MP) structures.}
\label{fig:homepage}
\end{figure}

To obtain the MetaMP database, the data preparation process combined automated curation, prioritizing specific attributes from four databases, with careful manual review to fully understand each attribute. 
Data curation refers to the process of organizing, managing, and refining data to ensure it is of high quality, relevant, and accessible. 
In our case, curation is a critical step to ensure the data is machine readable for data science, artificial intelligence, and visualization.
All attributes are listed in the supplementary material Tables 10 through 13.
The bibliography information was excluded because it did not have a direct relationship to the structure information of the MPs.
The MetaMP homepage enables quick searching of its database via a Google-like query field as shown in Figure~\ref{fig:homepage}.

\subsection*{Artificial Intelligence Use Cases}

Use cases demonstrate how membrane protein annotations can be made more accurate and consistent using AI. 
Use Case 1 uses AI to find discrepancies between old and new annotations to improve accuracy and consistency across databases, such as MPstruc and OPM. 
Use Case 2 shows how segment count helps classify proteins and select targets, with MetaMP predicting them and creating a reproducible process. 
Use Case 3 reveals internal logic of classification models and provides justifications
and insights with the help of XAI.

\subsubsection*{1. AI-assisted Legacy Database Reconciliation and Topology-Based Classification}

Historical databases like OPM and MPstruc contain curated entries that were classified before modern AI-based topology predictors. These may contain inconsistencies or gaps due to low resolution, partial models, and early curation.
We applied AI to address the issue of reconciling legacy records and reclassifying the topology. We used the platform's discrepancy detection engine to compare transmembrane segment counts predicted by TMbed~\cite{bernhofer2022tmbed} and DeepTMHMM~\cite{hallgren2022deeptmhmm} with those stored in the databases. This AI-assisted discrepancy check automatically flags entries with mismatched segment numbers or structural categories, facilitating expert review.

Beyond segment comparison, we trained an AI-based classification model to assign each protein to one of three structural groups defined by MPstruc: monotopic, alpha-helical transmembrane, and beta-barrel transmembrane. Unlike traditional sequence-based approaches, this model leverages structural metadata from OPM, including helix tilt angles, membrane thickness, subunit span, etc. These features capture the physical characteristics of membrane integration and allow the model to distinguish topological classes with high accuracy.

The Data Discrepancy view shows each protein's predicted class, original annotations, and the predicted segment counts. Proteins with discrepancies are automatically highlighted. The Selection View enables filtering of inconsistent entries, while the Ranking View orders entries by discrepancy magnitude, streamlining expert triage.
This AI framework bridges structural data with modern predictive capabilities, providing a scalable and transparent approach for refining membrane protein annotations. 

Supplementary Table~14, lists the full list of entries and contrasts classifications from four sources—OPM, MPstruc, MetaMP predictions and expert evaluations. 
Results showed 93 matches (76.86\%) between Expert and Predicted labels, 79 matches (65.29\%) between Predicted and OPM, 94 matches (77.69\%) between Predicted and MPstruc, 96 matches (79.34\%) between Expert and OPM, and 85 matches (70.25\%) between Expert and MPstruc. 
Although these figures show substantial concordance across resources, rigorous, expert-driven consistency checks are necessary.
This effort relies entirely on the MetaMP platform to link AI-derived TM-segment predictions to validated ground-truth counts. MetaMP integrates annotations from OPM, MPstruc and UniProt, and applies automated validations at every step. 
A central repository of both human annotations and model outputs is also maintained. This unified infrastructure was key to a systematic cross-resource evaluation.

\subsubsection*{2. AI prediction of the number of Transmembrane Segments}

The number of transmembrane (TM) segments in a protein is crucial because it determines the protein’s functional class, how it integrates into the lipid bilayer, and its role in signaling, transport, or structural stabilization. 
Proteins with multiple TM helices often form channels or transporters, while single-pass proteins typically function as receptors or anchors ~\cite{kihara1998prediction,remm2000classification,chou1999prediction}. Accurately predicting the number and position of transmembrane (TM) segments is a foundational requirement in both structural biology and bioinformatics, as the number of TM segments not only determines the protein’s topology but also plays a critical role in its classification within membrane protein families.

Motivated by this need, the MetaMP AI Annotation module supports large-scale topology screening and helix-predictor benchmarking in a unified workflow. The module applies both TMbed and DeepTMHMM to membrane protein sequences, extracts each tool’s predicted segment count, compares it against expert or expected values, and flags proteins whose predictions deviate beyond user-defined thresholds. This streamlines target triage in structural genomics or integrative modeling pipelines.
Simultaneously, the same interface computes benchmarking metrics (exact-match rate, MAE, Spearman’s $\rho$, Pearson’s $r$) for any selected predictor and displays results in the Benchmark View (Table~\ref{tab:benchmark}). This consolidated approach accelerates practical screening and the quantitative evaluation of new helix-prediction methods within MetaMP’s reproducible framework.

Building on Use Case 1, we applied two state-of-the-art predictors, TMbed~\cite{bernhofer2022tmbed} and DeepTMHMM~\cite{hallgren2022deeptmhmm}, to all entries in MetaMP, concentrating our analysis on the expert-annotated subset. 
Figure \ref{figure:datadiscrepancyview} compares 10 representative MPs drawn from 3 structural classes: bitopic alpha-helical, beta-barrel, and monotopic.
For the well-characterised bitopic set (1FDM, 1AFO, 2CPB), all sources—OPM, MPstruc, expert curation, TMbed, and DeepTMHMM—converge on a single transmembrane (TM) helix.
In contrast, proteins that MPstruc and domain experts classify as monotopic (1B12, 1KN9, 1OJA) are predicted by both AI models (TMbed, and DeepTMHMM) to contain two TM segments, mirroring OPM’s transmembrane assignment despite the zero-segment architecture supported by MPstruc and our experts.

\begin{figure}
    \centering
    \begin{minipage}[t]{.285\linewidth}
        \vspace*{0pt}
        \begin{subfigure}[t]{.9\linewidth}                \includegraphics[width=\textwidth]{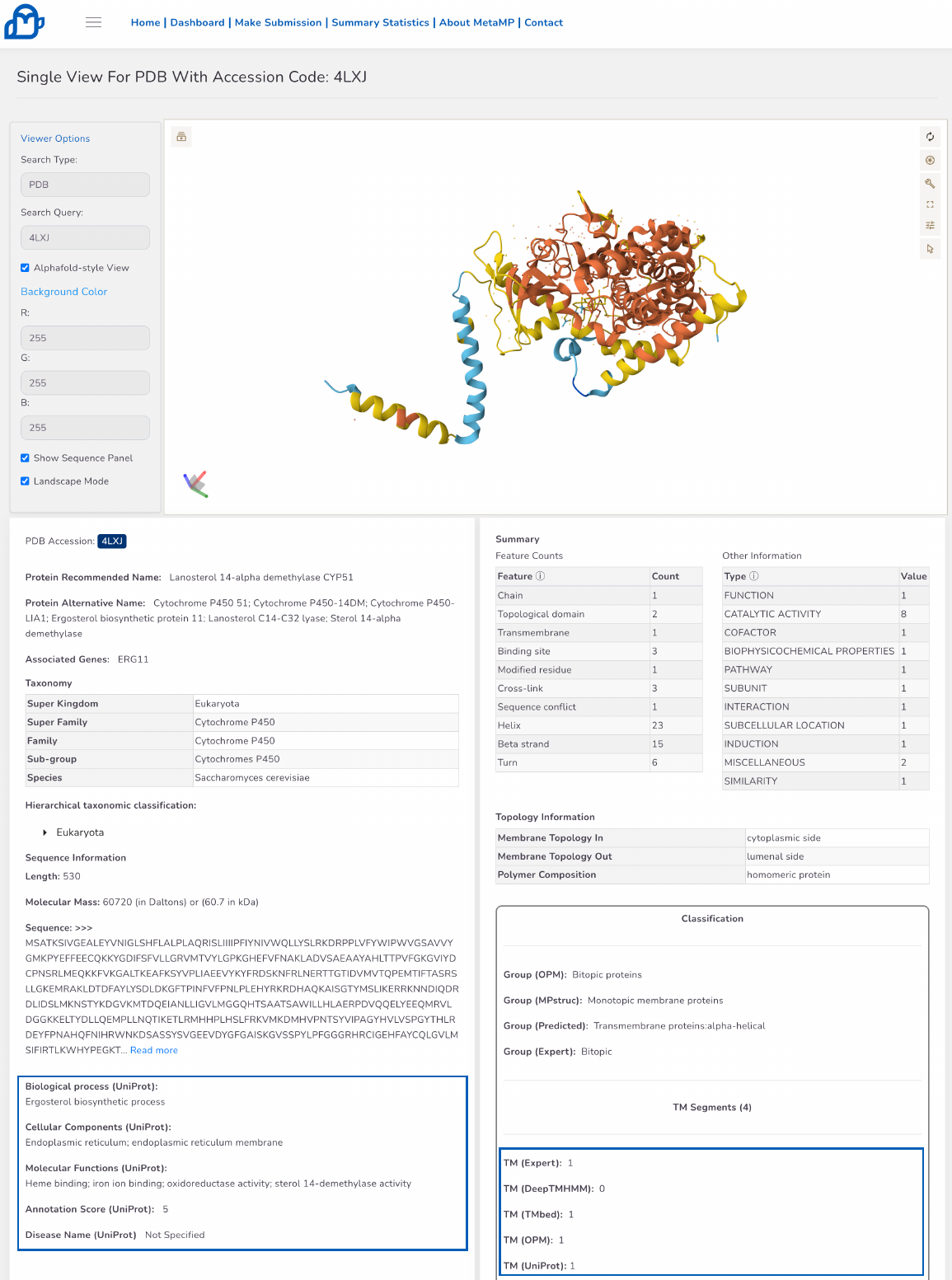}
            \caption{Screenshot of the view}
            \label{figure:singleview}
        \end{subfigure}
    \end{minipage}
    \hfill
\begin{minipage}[t]{.6\linewidth}
    \vspace*{0pt}
    \begin{subfigure}[t]{\linewidth}
        \frame{\includegraphics{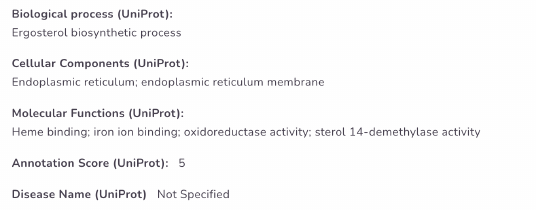}}
        \caption{Relevant Annotation and Functional metadata}
        \label{figure:singleview1}
    \end{subfigure} \\

    \begin{subfigure}[b]{\linewidth}
        \frame{\includegraphics{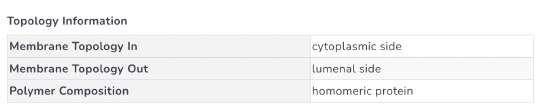}}
        \caption{Relevant Topological metadata}
        \label{fig:singleview2}
    \end{subfigure} 
\end{minipage}

    \caption{\textbf{The Single-Entry Structural view} combines protein structural, functional, and sequence information. The center panel shows a 3D molecular structure as a ribbon diagram, highlighting secondary structure elements and overall folding. Summary tables on the right count annotated structural features and functional annotations. Additional panels summarize the protein's taxonomy, sequence characteristics, topology, and curated biological annotations. This view integrates diverse annotations to facilitate comprehensive interpretation of protein features. The full screenshot of this view is available as Supp.~Fig.~14.}
    \label{figure:singlePageViewFull}
\end{figure}

Table~\ref{figure:datadiscrepancyviewtab} illustrates that while TMbed and DeepTMHMM generally align with some annotations made by the experts, there are notable exceptions. For example, TMbed predicts two TM segments in the beta-barrel protein 1PFO, whereas both the expert and DeepTMHMM assign zero. 
The Single-Entry Structural view of Lanosterol 14-alpha demethylase CYP51 (PDB: 4LXJ) is shown as an example in Figure~\ref{figure:singleview}.
Conversely, both models fail to detect single-pass helices in proteins like 1GOS, 1OJA, and 1O5W, where experts annotate one TM segment. These discrepancies may reflect differences in algorithmic interpretation or limitations in the original expert annotations, rather than fundamental uncertainty about the proteins' classification.

Across the full benchmark, the TM segment counts from TMbed and DeepTMHMM follow OPM’s assignments more closely than those from MPstruc or our expert curation. This likely reflects the fact that many public training sets (e.g.,~PDB-derived compilations) draw their membrane-boundary labels from OPM, whereas MPstruc and our experts apply stricter topological criteria~\cite{hallgren2022suppl, bernhofer2022tmbed}. By bringing all four sources into a single MetaMP-backed database and running automated consistency checks, our platform makes such cross-resource discrepancies explicit and provides a rigorous basis for improving future segment-prediction algorithms.

To generalize the case-by-case observations above, we quantified
predictor accuracy on the entire expert reference data subset
(Table~\ref{tab:benchmark}). DeepTMHMM reproduces the expert TM count in
74.4~\% of proteins, marginally outperforming TMbed (71.1~\%).
Nevertheless, both methods display broad error distributions
(std.\,$\approx$\,12.8), underscoring that a minority of predictions
still deviate by double-digit segment counts.

\begin{table}[ht]
\centering
\caption{\textbf{Performance of AI-assisted TM segment on the
expert annotated data subset ($n=121$).} Exact match\,=\,predicted TM count identical
to the expert annotation; MAE\,=\,mean $|\Delta TM|$; std.\,=\,standard
deviation of $|\Delta TM|$.}
\label{tab:benchmark}
\begin{tabular}{lccccc}
\toprule
\textbf{Predictor pair} &
\textbf{Exact matches} &
\multicolumn{1}{p{2.9cm}}{\centering $\mathbf{MAE \pm STD}$\\($\Delta$ TM segments)} &
\textbf{Spearman $\rho$} &
\textbf{Pearson $r$} \\
\midrule
TMbed $\rightarrow$ Expert        & \textbf{86 / 121\,(71.1\%)} & \textbf{3.36 $\pm$ 12.82} & \textbf{0.268} & \textbf{0.192} \\
DeepTMHMM $\rightarrow$ Expert    & \textbf{90 / 121\,(74.4\%)} & \textbf{3.32 $\pm$ 12.77} & \textbf{0.373} & \textbf{0.194} \\
DeepTMHMM $\leftrightarrow$ TMbed & 106 / 121\,(87.6\%) & 0.18 $\pm$ 0.56 & 0.739 & 0.938 \\
\bottomrule
\end{tabular}
\end{table}

\noindent DeepTMHMM and TMbed agree with each other in 87.6~\% of cases,
with an average difference of only 0.18±0.56 segments (Spearman
$\rho$ = 0.74, Pearson $r$ = 0.94), indicating that they share
systematic tendencies. Accepting only those proteins where both
predictors concur therefore yields a high-confidence subset
($\approx$ 88 \%), whereas the remaining $\approx$ 12 \% of proteins benefit from additional evidence such as cryo-EM density or biochemical topology
assays (See Supplementary~Figure~11).

\subsubsection*{3. Explainable AI for Structural Classification of Membrane Proteins}

\begin{figure}[htbp]
\centering
\includegraphics[width=\textwidth]{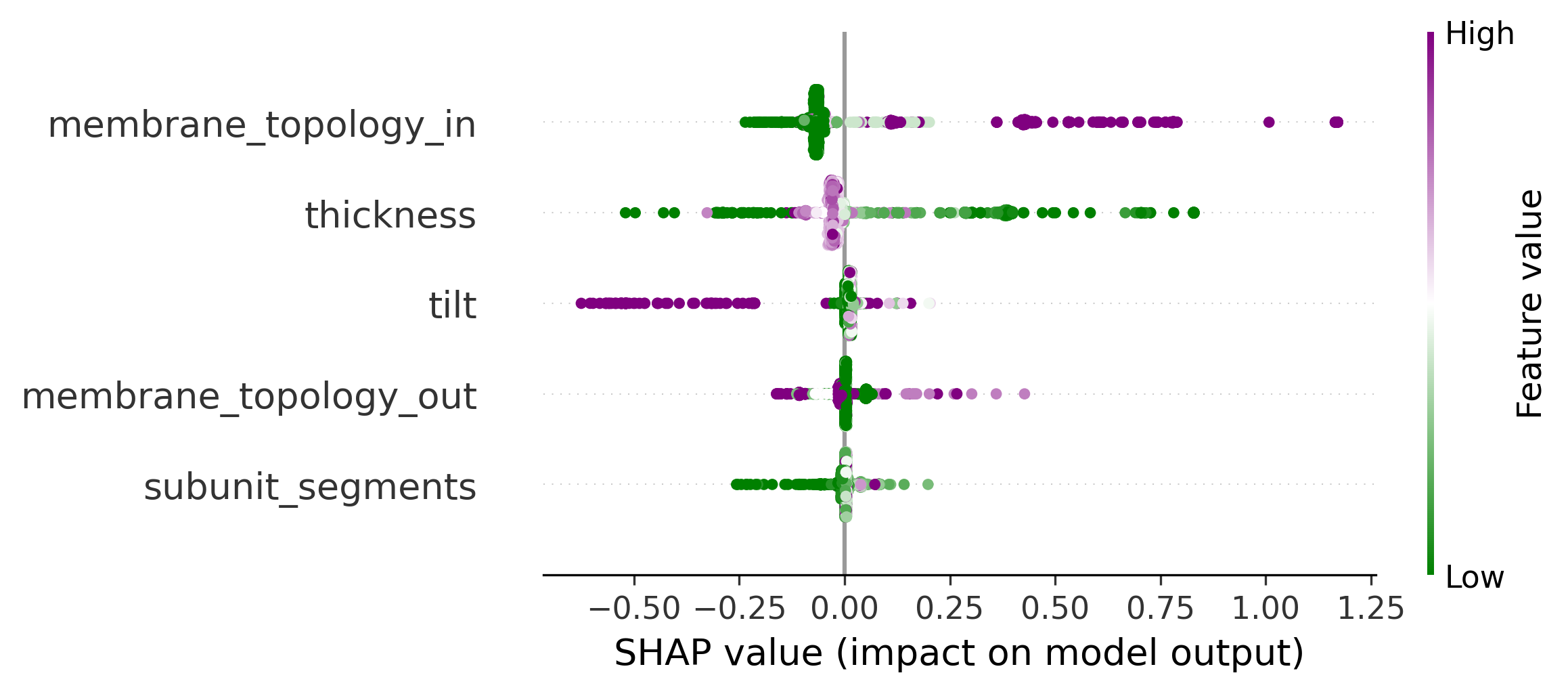}
\caption{\textbf{Shapley summary chart}. This chart depicts the contribution of each feature to the prediction model, illustrating the feature importance and their respective impact on the target variable. Each dot represents a single observation in the dataset, where the position along the x-axis shows the SHAP value (effect on the prediction), and the color gradient indicates the feature value (from low to high). Features with higher SHAP values have a more substantial influence on the prediction. This plot not only ranks the features by importance but also provides insights into how different values of each feature drive model predictions. (green = low, purple = high)}
\label{figure:shap}
\end{figure}

We built an explainable AI (XAI) workflow for interpreting structural classifications of membrane proteins using the capabilities in Use Case 1 and the framework in Use Case 2.
Our classification model, adopted from Use Case 2, groups proteins into three OPM topological classes: thickness, tilt, and subunit segments (numerical), and membrane topology in/out (categorical). The model achieved high accuracy, but understanding the predictions' drivers is key for interpretability, trust, and scientific insight.
To this end, we applied SHAP (SHapley Additive explanations) to quantify the contribution of each feature. The summary plot in Figure~\ref{figure:shap} highlights five key features. Each protein instance is colored by feature value and positioned by Shapley value, showing the feature's marginal impact on class assignment.

Several trends emerged: proteins with low helix tilt and fewer subunit segments were strongly linked to the monotopic class, while those with higher membrane thickness and tilt were favored alpha-helical classifications. The membrane topology features provided more context. Topological types affected the predictions, showing the importance of structural cues in class membership.
This example illustrates how MetaMP's models make accurate predictions and offers interpretability, hence strengthening trust in the underlying models. 

\subsection*{Eight Interactive Views}
MetaMP offers eight rich and context-focused views to support experts in their tasks: Overview, Summary Statistics, Membrane Insight View, Data Discrepancy, Outlier Detection, Database, Exploration, and Grouping.
Altogether, these interactive views provide a comprehensive understanding of metadata for MP structures. 
Indeed, this unified web application improves understanding of the specific protein class of MP while providing broader insights, effectively streamlining the process that traditionally required extensive manual curation by domain experts.
Two example views are shown: Data Discrepancy and Exploration.

\begin{figure}
\begin{subfigure}[t]{\textwidth}
\centering
\includegraphics[width=\textwidth]  {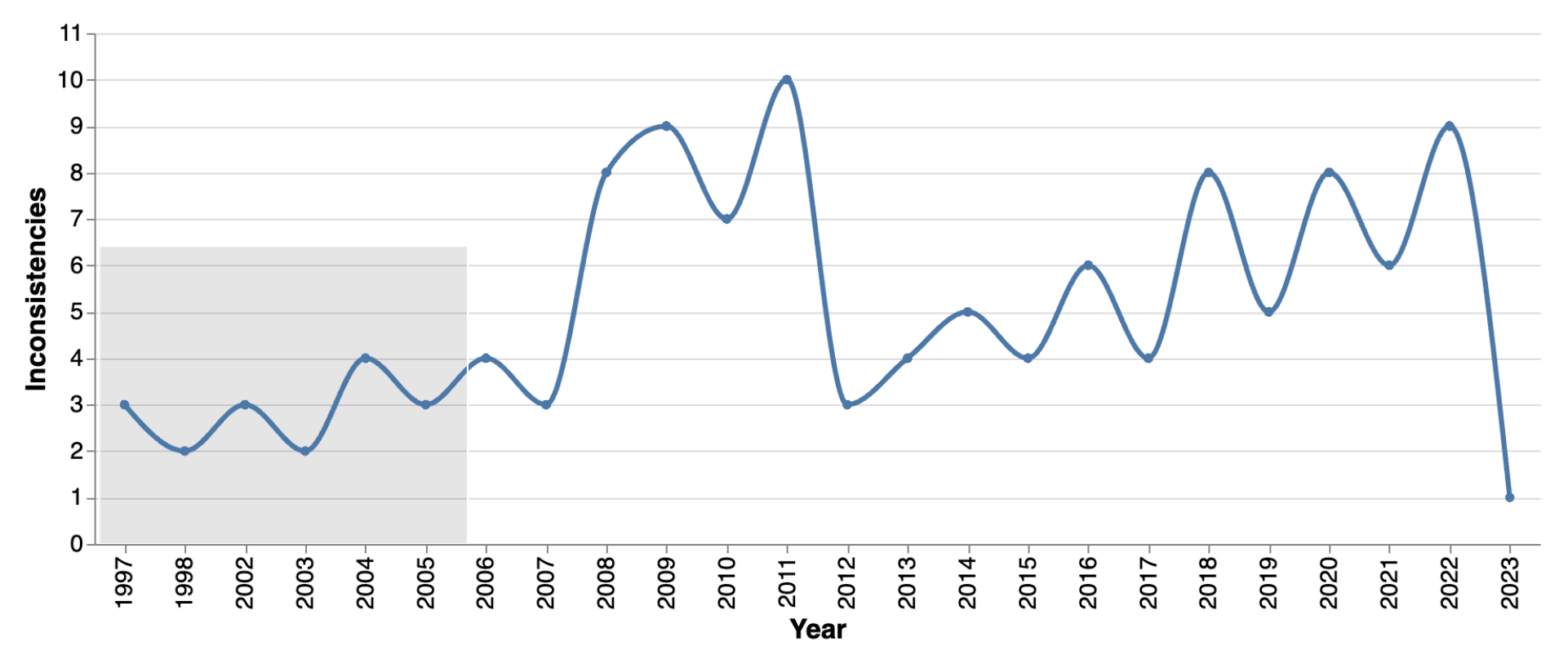}
  \caption{Data Discrepancy Line Chart.}
\label{figure:datadiscrepancyviewchart}
\vspace*{3mm}
\end{subfigure}
\hfill
\begin{subfigure}[t]{\textwidth}
\centering
\resizebox{\textwidth}{!}{%
\tiny
\begin{tabular}{ccp{2cm}p{2.5cm}p{2cm}p{2cm}p{1cm}p{1cm}p{1cm}}
\hline
\textbf{Year} & \textbf{PDB Code} & \textbf{Group (OPM)} & \textbf{Group (MPstruc)} & \textbf{Group (Predicted)} & \textbf{Group (Expert)} & \textbf{TM (Expert)} & \textbf{TM (TMbed)} & \textbf{TM (DeepTMHMM)} \\
\hline
1997 & 1PFO & Monotopic membrane proteins & Transmembrane proteins:beta-barrel & Transmembrane proteins:beta-barrel & Transmembrane proteins:beta-barrel & 0** & 2 & 0\\
1997 & 1FDM & Bitopic proteins & Transmembrane proteins:alpha-helical & Transmembrane proteins:alpha-helical & Bitopic & 1 & 1 & 1\\
1997 & 1AFO & Bitopic proteins & Transmembrane proteins:alpha-helical & Transmembrane proteins:alpha-helical & Bitopic & 1 & 1 & 1\\
1998 & 2CPB & Bitopic proteins & Transmembrane proteins:alpha-helical & Transmembrane proteins:alpha-helical & Bitopic & 1 & 1 & 1\\
1998 & 1B12 & Transmembrane proteins:alpha-helical & Monotopic membrane proteins & Monotopic membrane proteins & Monotopic membrane proteins & & 2 & 2 \\
2002 & 1GOS & Bitopic proteins & Monotopic membrane proteins & Monotopic membrane proteins & Bitopic & 1 & 0 & 0\\
2002 & 1MT5 & Bitopic proteins & Monotopic membrane proteins & Monotopic membrane proteins & Monotopic membrane proteins &  & 0 & 0\\
2002 & 1KN9 & Transmembrane proteins:alpha-helical & Monotopic membrane proteins & Monotopic membrane proteins & Monotopic membrane proteins & & 2 & 2 \\
2003 & 1OJA & Bitopic proteins & Monotopic membrane proteins & Monotopic membrane proteins & Bitopic & 1 & 0 & 0\\
2003 & 1MZT & Bitopic proteins & Transmembrane proteins:alpha-helical & Transmembrane proteins:beta-barrel & Bitopic & 1 & 1 & 1 \\
2004 & 1O5W & Bitopic proteins & Monotopic membrane proteins & Transmembrane proteins:alpha-helical & Bitopic & 1 & 0 & 0\\
2004 & 1PJF & Bitopic proteins & Transmembrane proteins:alpha-helical & Transmembrane proteins:alpha-helical & Bitopic & 1 & 1 & 1\\
2004 & 1UUM & Bitopic proteins & Monotopic membrane proteins & Monotopic membrane proteins & Bitopic & 1 * & 0 & 0 \\
2004 & 1T7D & Transmembrane proteins:alpha-helical & Monotopic membrane proteins & Monotopic membrane proteins & Monotopic membrane proteins &  & 2 & 2\\
2005 & 2BXR & Bitopic proteins & Monotopic membrane proteins & Transmembrane proteins:alpha-helical & Bitopic & 1 & 0 & 0 \\
2005 & 1YGM & Monotopic membrane proteins & Transmembrane proteins:alpha-helical & Transmembrane proteins:alpha-helical &  Not a Membrane Protein &  & 0 & 0 \\
2005 & 1ZLL & Bitopic proteins & Transmembrane proteins:alpha-helical & Transmembrane proteins:alpha-helical & Bitopic & 1 & 1 & 1\\
\hline
\end{tabular}%
}
\caption{Data Discrepancy Table.}
\label{figure:datadiscrepancyviewtab}
\end{subfigure}
\caption{
\textbf{Data Discrepancy view from 1997 to 2005.}
One of the eight views available in MetaMP, the Data Discrepancy view illustrates classification inconsistencies in membrane protein structures across the four integrated databases from 1997 to 2005. Discrepancies are primarily observed between OPM and MPstruc, highlighting differences in structural categorization and shifts in classification trends over time. This view is implemented as a line chart linked to a dynamic table—interacting with the chart updates the corresponding table content. Subfigure b is interactive: users can sort columns, and row-based highlights reveal a second layer of discrepancy related to the number of transmembrane (TM) segments. This eight-year sample provides insight into how membrane protein representations and groupings have evolved across data sources.
}
\label{figure:datadiscrepancyview}
\end{figure}

The Data Discrepancy view is shown in Figure~\ref{figure:datadiscrepancyview}.
The 11 discrepancies observed from 1997 to 2005, out of a total of 121, highlight the need to resolve such data inconsistencies. 
After years of experience, domain experts carefully review and resolve the list of data inconsistencies present in databases.
To resolve such inconsistencies, undertook a comprehensive re-evaluation of each three-dimensional structure, by directly counting the number of trans-membrane (TM) segments from a visualization of each protein structure.
Three exemplary cases of this process and the associated rationale for each MP structure are reported.
1PFO or perfringolysin O is originally misclassified, and is clearly a transmembrane beta-barrel protein. This classification is based on the number of transmembrane segments (TM), pore-forming activity, and high-resolution crystallographic evidence for a membrane-spanning beta-barrel structure. 
1B12 is \textit{E. coli}'s signal peptidase, initially classified as transmembrane (OPM) or monotopic (MPstruc), but is now definitively categorized as monotopic. 
Structural and biochemical studies~\cite{paetzel2014structure} confirm its interaction with only one face of the membrane, without spanning the entire lipid bilayer. 
1YGM is not a membrane protein itself, but rather a unique protein that supports the expression of other membrane proteins. 
Originally identified in \textit{Bacillus subtilis}, Mistic functions as a fusion partner to enhance the production of integral membrane proteins in bacterial expression systems, particularly in \textit{E. coli}. While Mistic associates with membranes, it does not insert into or span the lipid bi-layer like typical membrane proteins. Its unusual properties, including a surprisingly polar surface, allow it to bypass the cellular translocon machinery and facilitate the expression of challenging membrane proteins.
The full list of these 121 expert-curated corrections appears in Supplementary Table 14. 

The Single-Entry Structural view, Figure~\ref{figure:singleview}, portrays the metadata enrichment and AI capabilities of MetaMP, which facilitate access to metadata and the visual investigation of the three-dimensional structure of an MP.

The remaining views are available on the MetaMP website and are visually documented in the Supplementary.

\subsection*{Improvement on cryo-Electron Microscopy} 
We observe that cryo-Electron Microscopy (cryo-EM) has seen a rise of resolved structures, while X-ray crystallography has consistently been used for structural determination.
To validate our database, we extracted known emerging techniques to resolve MPs such as cryo-EM from our database. 
As expected we found that the average resolution of MP structures determined by cryo-EM has significantly improved, rising from 7.95 ± 2.47 Angstroms (Å) in 2012 to 3.17 ± 0.39 Angstroms (Å) in 2024. 
In contrast, X-ray crystallography has consistently resolved MP structures with an average resolution of approximately 2.7 Angstroms (Å). Further information is available in the supplementary material. 


\subsection*{Geographical Distribution of Research Contributions} 
Geographical analysis highlighted that most research contributions originate from the United States and the United Kingdom, which collectively represent over 95\% of the dataset 
(see~Supp.~Fig.~1).    

\subsection*{Quality Control}
MetaMP employs a comprehensive Quality control (QC) mechanism to address inconsistencies and enhance the reliability of MP structure data.
The QC process comprises outlier and consistency analysis. 
It is essential for ensuring that the data used in research is accurate, consistent, and of high quality. 
As a direct result of implementing this process, sixteen outdated PDB codes were found and automatically updated to the official accession codes in the PDB database (see ~Supp.~Table~2).
The complete list of old and updated accession codes is reported in the supplementary material.
This process ensures quality for subsequent applications in high-stakes domains like artificial intelligence and medicine~\cite{dubey2024nested}.  

On one hand, the outlier analysis identified a notable entry, with protein code 6ZG5, which has a low resolution of 40 Angstroms (Å) due to cryo-EM subtomogram averaging technic on the complex assembled in membrane. 
Supp.~Fig.~7 showcases this outlier. 
The QC process prompted further investigation into the outlier's structural and functional implications.
On the other hand, the consistency analysis revealed significant discrepancies in the classification of MP structures: several proteins were classified differently in the OPM database compared to the MPstruc database.
This prompted discussions over the MP types and classes~\cite{shimizu2018comparative}.

    

\subsection*{Feature Selection and Machine Learning Model Evaluation for the Classification of MPs} 

The Random Forest (RF) feature selection process identified five important attributes from the OPM database: three numerical and two categorical. 
The numerical attributes are \texttt{Thickness}, \texttt{Tilt}, and \texttt{Subunit Segment}. 
The categorical attributes are \texttt{Membrane topology in} and \texttt{Membrane topology out}. 
These attributes were found to be the most important in the MetaMP database and were used in machine learning.
The attributes all originated from the OPM database.
 
Newly resolved MPs are usually manually curated and assigned to one of the three main types on MPstruc: monotopic membrane proteins, transmembrane alpha-helical proteins, and transmembrane beta-barrel proteins.
We compared the performance of 7 supervised and 7 semi-supervised learning models to assist human experts for the classification task.
The supervised learning models trained on labeled data alone served as baseline benchmarks for comparison with the semi-supervised models. 
The semi-supervised models, which incorporated labeled and unlabeled data, outperformed their supervised counterparts in most cases. 
Across all classifiers, the semi-supervised models exhibited a notable improvement in accuracy compared to their supervised counterparts. 
On average, the accuracy of the semi-supervised RF model increased by approximately 0.92\%, rising from 97.6\% to 98.5\%. Additionally, the F1-score saw an increase of about 0.82\%, from 97.7\% to 98.5\%.
The semi-supervised RF model demonstrated superior performance across all metrics, achieving the highest mean accuracy of 0.977 (±0.005) and F1-score of 0.976 (±0.004), outperforming all other six models in overall consistency and effectiveness.
The performance metrics of all machine learning models are reported in Supp.~Table~3.

While this model demonstrates strong performance for newly resolved membrane proteins, its application can also be extended to address data discrepancies and expedite their resolution.
To assess the real-world applicability and robustness of the trained model, the model predictions were compared with the expert evaluations.
A detailed breakdown can be found in Supp.~Table~14.
In total, 93 entries out of 121 are correctly predicted, which accounts for 76.86 or approximately 77\%.
Although bitopic is reported in this table as a MP group, bitopic proteins are transmembrane alpha-helical proteins.
Trying to interpret the results in Supp.~Table~14, there are some prediction errors for which a reason or explanation can be found. 
For example, in the case of 1MZT and 1FDM, the trained model misclassified 1MZT as a beta-barrel. However, both proteins are structurally similar, probably due to its higher alpha-helical content influencing the prediction.
For 1OJA, which the model classified as monotopic, the partial visibility of its transmembrane domain suggests flexibility that may have obscured its membrane-spanning properties. 
These examples illustrate the challenges of model predictions in accurately reflecting protein topology in the midst of structural dynamics.

\subsection*{Task-oriented User Evaluation}
A task-oriented user evaluation was conducted, comprising three consecutive tasks with training and testing phases, and varying degrees of difficulty. 
The tasks included generating summary statistics and finding outliers in a subset of the data.
The tasks mapped well to two views -- Summary Statistics view and Outlier Detection view -- and allowed for explicit evaluation of the features contained therein.
A total of 24 participants took part in the user study and completed all tasks in full (see Supp.~Table~6). 
One participant was excluded from the study for failure to complete the requisite tasks.
All participants were volunteers and received no compensation for their participation.
The following sections present an overview of the quantitative and qualitative results of the user study.

\subsubsection*{Quantitative Results}
The participants were identified as male (n=13), female (n=10), or declined to disclose their gender (n=1).  
The supplementary material provides a detailed overview of the socio-demographic characteristics, domain expertise, and years of experience of the participants.

The combined training and testing phases, conducted on separate datasets for all tasks, were completed by participants in less than ten minutes.
The participants mean score was 4.21 ± 0.98 out of 6. 
The average time to complete the training and test tasks was 9.34 minutes.
On average, users completed the testing tasks in 41.47\% less time than the training tasks. 
This improvement, where participants became approximately 41\% faster, is indicative of the typical learning curve, whereby individuals enhance their efficiency following an initial training phase.
Similarly, participants dedicated approximately 70.86\% more time to training activities. 
This notable discrepancy is likely attributable to the learning nature of training. 
Results indicate that participants required a significantly longer time to become acquainted with these tasks during the training phase as opposed to the testing phase.
Examination of task completion times revealed clear patterns of central tendency and variability.
The mean times for the tasks ranged from approximately 0.7 minutes for Task 2 to about 3 minutes for Task 3, indicating variation in task complexity and duration. 
Notably, Task 2 had the shortest mean time, suggesting that it was completed more quickly on average, while Task 3 had the longest mean time, reflecting greater complexity or difficulty.

Three hypotheses were formulated in advance and subsequently tested in order to gain a deeper understanding of participant behavior. 
The hypotheses focused on three key areas: 
(1) the effectiveness of the training or learning process, 
(2) the difficulty of the task, and 
(3) the optimal balance between speed and accuracy.

\textbf{Hypothesis 1: Learning effectiveness.}  
\textit{Null hypothesis (H\textsubscript{01}):} There is no difference in completion times between training and testing phases.  
\textit{Alternative hypothesis (H\textsubscript{11}):} Participants complete the task significantly faster during the testing phase compared to the training phase.

To assess the appropriateness of statistical testing, we evaluated the distribution of differences in completion times between training and testing using the Shapiro–Wilk test. The result (W = 0.803, $p < 0.001$) indicated a significant deviation from normality. Given this violation of the normality assumption, we did not use the paired t-test. Instead, we applied the Wilcoxon signed-rank test, a non-parametric alternative suited for non-normal data.

The average time required for the training phase was approximately 2 minutes, while the average time for the testing phase was about 1.2 minutes, indicating a notable reduction in time.  
The Wilcoxon test yielded a test statistic of $W = 821.000$ with a $p$-value of $0.006$.  
This significant result supports the hypothesis that training effectively enhances efficiency in task completion.

\textbf{Hypothesis 2: Task difficulty.}  
\textit{Null hypothesis (H\textsubscript{02}):} There is no difference in completion times across tasks.  
\textit{Alternative hypothesis (H\textsubscript{12}):} Task difficulty significantly affects completion time.

The results of the difficulty of the task, hypothesis (2), and the ANOVA test yielded an F-statistic of 3.824 and a p-value of less than 0.001.  
Given that the p-value is below the standard significance threshold of 0.05, we can conclude that there are statistically significant differences in completion times between tasks.  
In particular, Task 3, which had the highest average completion time of 3 minutes, was identified as the most challenging.  
A note was made for this task as it involved getting accustomed to the interaction with various interactive charts in a large and intricate view composition.  
This view supported the task of outlier detection and included a Scatterplot matrix (SPLOM), a whisker plot, and a scatter plot.  
In contrast, Task 2 had the lowest average time of 0.682 minutes and was determined to be the least challenging.  
These findings confirm that task difficulty varies significantly and impacts the time users need to complete them.  

\textbf{Hypothesis 3: Speed–accuracy trade-off.}  
\textit{Null hypothesis (H\textsubscript{03}):} Time taken to complete a task does not significantly affect the likelihood of correctness.  
\textit{Alternative hypothesis (H\textsubscript{13}):} There is a statistically significant trade-off between speed and accuracy.

The results of the logistic regression model, which was fit to the data, showed that the intercept had a coefficient of 1.1128 (p < 0.001), while the coefficient for time taken was -0.1555 (p-value < 0.05).  
The model's log-likelihood was -85.857, with a pseudo R-squared value of 0.022.  
The correlation between time taken and correctness was -0.170.  
These findings suggest that the relationship between speed and accuracy is weak and not statistically significant.  
Therefore, the hypothesis that a trade-off exists between speed and accuracy is not strongly supported by the data.  
This indicates that in this context, the speed of task completion does not significantly affect the likelihood of errors.  
These results suggest that Task 2 was relatively straightforward, with both low average completion time and minimal variability, while Task 3 posed greater challenges, as evidenced by its high mean time and substantial variability.  

For a further statistical analysis and detailed metrics, please refer to the supplementary material.

\subsubsection*{Qualitative Results}
Our questionnaire included an optional text box for users to provide feedback about the system, the study, or any inconsistencies they encountered. 
User feedback has been instrumental in refining MetaMP. 
Positive aspects such as speed and reliability were appreciated, while constructive criticism led to improvements in chart positioning, drop-down functionality and system responsiveness. 
System usability was rated positively by most participants as seen in Supp.~Fig.~11 showcasing the results of the system usability scale (SUS) as a violin plot.
Further results can be found in the supplementary material including the feedback shared by participants for data visualizations, and interactive features.

\section*{Methodology}

\subsection*{Materials}
MetaMP obtained its data from four databases: MPstruc~\cite{MPstrucpdb}, PDB~\cite{bittrich2023rcsb}, OPM~\cite{lomize2012opm} and UniProt~\cite{uniprot2023uniprot}. 
This section presents the necessary information about each of these databases.
The MPstruc data were downloaded from the MPstruc website in XML format.
A Python script was then used to extract information from this data file, including protein group, subgroup, name, species, taxonomic domain, and resolution. 
MetaMP uses unique identifiers, such as PDB codes and UniProt IDs to systematically retrieve records, ensuring comprehensive data extraction and accurate representation.

The MPstruc database provides a structured classification system for MPs that includes three hierarchical levels: groups, subgroups, and individual proteins~\cite{MPstrucpdb}. 
At the group level, proteins are categorized based on their interaction with the membrane. 
For example, proteins may be monotopic, interacting with only one side of the bilayer membrane or span the membrane using structures such as alpha helices or beta barrels. 
Subgroups further organize proteins by function and taxonomy. 
The most specific level, the individual protein, corresponds to different PDB structures within each subgroup~\cite{shimizu2018comparative, hatami2023preparing}.
MPstruc serves as our primary source because of its human-curated nature, which helps mitigate many problems arising with fully automated procedures. 
However, it is important to recognize that human error can still affect the accuracy of its content.
The RCSB Protein Data Bank (PDB) is a fundamental repository for the 3D structural data of biological molecules and provides metadata describing the biological context of protein structures, including resolution, molecular weight, source organism, experimental techniques, and relevant literature references~\cite{burley2019rcsb}.
This database can be further explored using the RCSB PDB Structure Search Attributes.
The Orientations of Proteins in Membranes (OPM) database  offers metadata on the spatial orientation of MPs within lipid bilayers and topological data on transmembrane helices~\cite{lomize2012opm}.
The UniProt database provides detailed information on molecular functions, cellular components, and biological processes, protein-protein interactions, and taxonomic information about the proteins and their species of origin~\cite{uniprot2023uniprot}.


MetaMP is built on a three-tiered architecture that includes the Data, Application, and Presentation layers~\cite{nestler2011end}.
The architecture is illustrated in Supp.~Fig.~12. 

\subsubsection*{Data Layer}
The data layer is fundamental to the functionality and effectiveness of MetaMP.
To build this layer, we follow the Extract, Transform, and Load (ETL) methodology~\cite{khan2024overview}, as shown in Figure~\ref{fig:ETL}.
The ETL process begins with the extraction of data from the databases. 
The extracted data is transferred to the staging area, where it is temporarily stored and processed. 
This staging area acts as a buffer, allowing the data to be verified, validated, and, if necessary, transformed before being loaded into the MetaMP database using PostgreSQL. 
The staging area maintains the integrity and quality of the data during the transfer while ensuring performance.
Staging area operations include data cleansing, filtering, data normalization, verification of data transfer, data restructuring, and combining data with lookups. 
Each of the six operations is described below:

\begin{figure}
\centering
\includegraphics[width=\textwidth]{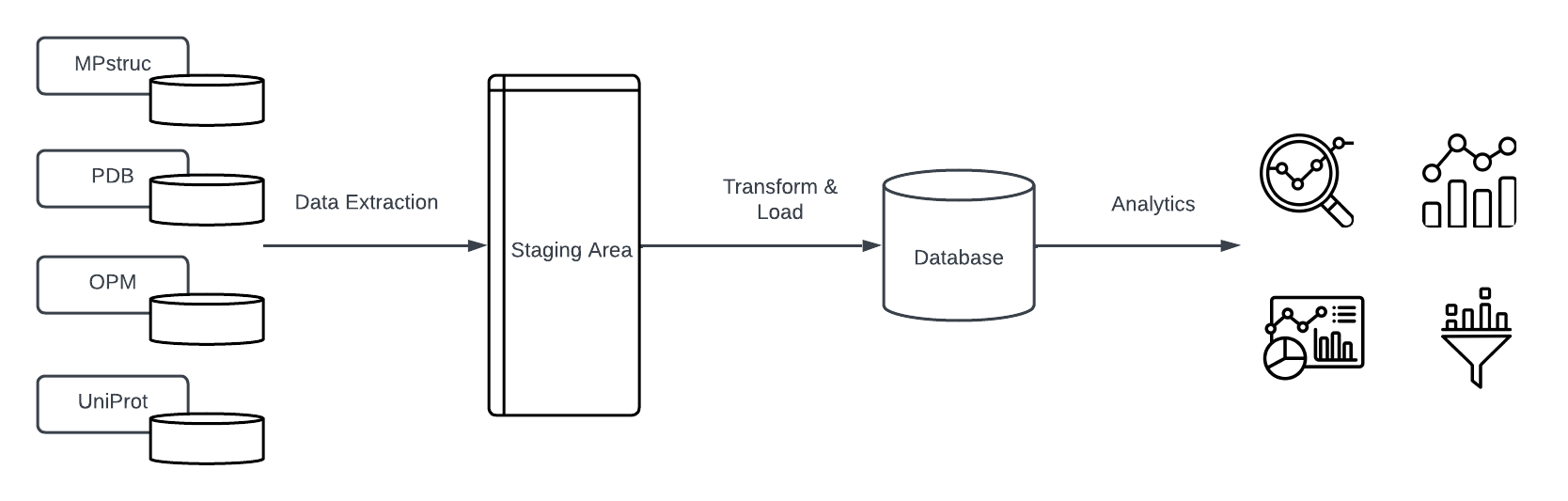}
\caption{
    \textbf{Diagram of the extract, transform, and load (ETL) data pipeline}.
    It illustrates the Extract, Transform, and Load processes that gather data from databases, including PDB, UniProt, and PubChem, convert it into a suitable format, and load it into MetaMP. 
    This pipeline ensures efficient and accurate data integration for subsequent analyses.
}
\label{fig:ETL}
\end{figure}

\begin{enumerate}[noitemsep]
    \item Data cleansing involves removing unnecessary spaces or special characters that may be invalid for our operations.
    \item Filtering selects specific columns/records that are essential for the analysis. For example, we retrieved only records from the PDB for MP structures listed in MPstruc.
    \item Data normalization standardizes data using rules and reference tables. For example, we ensured consistency in the organism expression system attribute, which can appear as variations such as \verb|"E. Colli"|, \verb|"E. Coli"|, or \verb|"Escherichia Coli"| extracted from MPstruc.
    \item Verification of data transfer ensures the successful data transfer from the staging area to intermediate tables within the MetaMP database.
    \item Data restructuring splits complex columns into multiple columns (column expansion). For example, we split the attribute \verb|"exptl_crystal_grow"| into four additional columns with the parent column name as a prefix.
    \item Combining data with Lookups merges data from multiple sources using reference tables and identifiers such as PDB CODE and UniProt ID for integration.
\end{enumerate}

By performing data transformations in the staging area, MetaMP minimizes the impact of performance issues. 
This approach also facilitates the early detection and correction of errors or inconsistencies in the extracted data. 
Once transformed, the data is efficiently loaded into the database to ensure optimal performance. 
In the event of a load failure, MetaMP includes recovery mechanisms to resume operations while maintaining data integrity.

\subsubsection*{Application Layer}
The application layer is designed to provide a robust application programmable interface (API) for seamless interaction with the enriched MetaMP database~\cite{bloch2006design}. 
This layer abstracts complex backend processes and provides a user-friendly interface that facilitates efficient data access, analysis, and visualization. 
Four key components are implemented at the application layer.
The four components include data access and retrieval, integration and interoperability, continuous database updates, and performance optimization.
They are briefly described below:

\begin{enumerate}[noitemsep]
\item {Data Access and Retrieval}: Search functionality allows users to search for proteins and related information based on various criteria such as pdb code and protein name.
Advanced filters allow users to refine their searches by applying filters such as groups, subgroups, membrane names, and functional annotations.
\item Integration and Interoperability: APIs and Web services provide APIs for programmatic access to data and services, enabling integration with other bioinformatics tools and platforms for further analysis.
Data export capabilities support the export of data in a variety of formats for further analysis in external applications.
These formats include structured data from the database or data visualizations from the user interface, (Comma-Separated Values) CSV files and (JavaScript Object Notation) JSON files, (Portable Network Graphics) PNG and (Scalable Vector Graphics) SVG, respectively.
\item {Continuous Database Updates}: The Continuous Database Updates section of MetaMP is responsible for keeping the four built-in databases up to date with the latest information. 
The key aspects of the continuous database update process in MetaMP include automated data retrieval, data synchronization, and incremental updates, all managed by Python scripts. These scripts are scheduled to run at regular intervals using services such as cron jobs for task scheduling. They connect to the databases via APIs to retrieve the latest data, handle API interactions, request updates, and handle issues such as network interruptions or API rate limits. The retrieved data is then parsed and transformed into a unified format suitable for MetaMP, ensuring consistency across all integrated datasets. The system performs incremental updates, modifying only the changed or new records to reduce processing time and minimize disruption, with the synchronized data integrated into MetaMP's internal database through controlled transactions to maintain data integrity.
In addition, version control and regular backups ensure security and provide rollback capabilities. Continuous monitoring through routine health checks verifies connectivity, data integrity and overall system performance, while detailed error logs are maintained for efficient troubleshooting and resolution of any issues. Through these meticulous methods, MetaMP maintains the accuracy and reliability of its integrated databases.
\item {Performance Optimization}: Caching mechanisms implement caching to improve the speed and efficiency of data retrieval and processing using Redis due to high throughput and low latency~\cite{priovolos2019escape}.
Containerization Using Docker streamlines scaling, deployment, and management processes, improves system reliability, and minimizes environmental issues ~\cite{merkel2014docker}.

\end{enumerate}

\noindent In summary, MetaMP is a web server developed using Python 3, Flask 2.2.5 and PostgreSQL, with frontend technologies including Vue.js, Bootstrap, HTML and CSS. 
It is compatible with all major browsers. 
The advanced PostgreSQL database is used for data management, while Docker containers provide consistency across deployment environments for easier scaling, deployment, and management. 
Docker also simplifies the development workflow and improves system robustness. 
MetaMP is scalable and can be integrated with Kubernetes to optimize performance.

\subsubsection*{Presentation Layer}
The presentation layer of MetaMP is a user interface (UI) that provides access to integrated data and analysis functionalities. 
It includes interactive data visualizations, intuitive navigation and compatibility with web standards. 
The presentation layer enables seamless interaction with the MetaMP database. 
It provides a rich and contextual representation of relevant information.
MetaMP provides eight different views, 
including the Overview, 
the Summary Statistics view, 
the Data Discrepancy view, 
the Outlier Detection view, 
the Database view,
the Exploration view, 
and the Grouping view, which provides comprehensive analysis and easy access to the data.

\begin{enumerate}[noitemsep]
    
    \item \textbf{MetaMP Overview}: The overview offers high-level information on the enriched MP structure data. 
    It comprises data visualizations of the MP structures and their associated metadata. 
    The data visualizations include the MP structures resolved by different experimental methods,
    the median resolution by experimental method over time, 
    the MP structures published by country (country of submission), the cumulative sum of resolved MP structures over time, categorized by taxonomic domain (Archaea, Bacteria, Eukaryota, Unclassified, and Viruses), and also categorized by group (monotopic, transmembrane alpha and beta). A screenshot of this view is provided in Supp.~Fig.~1.

    \item \textbf{Summary Statistics view}: 
    This view provides on-demand information about MP structure metadata to gain insight into its distribution and variability. 
    The main visualization presented in this view is a bar chart idiom. 
    It shows the cumulative sum of resolved MP structures, categorized by experimental method. Below the chart, a table provides a comprehensive list of all data points utilized in the interactive visualization.
    On-demand updates are available to examine the data distribution of various attributes, listed as follows: 
    by experimental method and molecular type, 
    by engineered source organisms,
    by expression system organism,
    by resolution,
    by software, 
    by space group,
    by molecular weight (structure),
    by atom count
    by groups,
    by journal and
    by growth method.
    Selecting an attribute updates the corresponding interactive visualization and table.  A screenshot of this view is provided in Supp.~Fig.~2.
    Interactive functionality in this view was evaluated during Task 1 of the task-Oriented user evaluation.    

    \item \textbf{Data Discrepancy view}: 

The Discrepancies View shown in Figure~\ref{figure:datadiscrepancyview} comprises two coordinated panels for rapid identification and resolution of metadata mismatches. 
The upper panel of Figure \ref{figure:datadiscrepancyviewchart} combines a line chart of annual inconsistency counts with an embedded, scrollable table of each discrepant entry. Every row lists the PDB code, the conflicting group assignments across OPM, MPstruc, TMbed and DeepTMHMM predictions, and expert labels, together with the year of structure determination and experimental method; each entry can be selected for in-depth review or submitted directly via the adjacent feedback form.

The lower panel (Table \ref{figure:datadiscrepancyviewtab}) presents the complete set of membrane-protein records, including expert-verified TM counts alongside TMbed and DeepTMHMM predictions. A real-time search box filters by PDB code, classification or TM count, and pagination controls ensure smooth navigation through larger datasets. By combining trend visualization with detailed records and integrated feedback, this two-panel layout makes every discrepancy both visible and immediately actionable.

    \item \textbf{Outlier Detection view}: 
    This view focuses on identifying and analyzing outliers within the MP structure data. 
    Outliers are data points that deviate significantly from the overall pattern and can provide valuable insights or indicate potential errors in the data. 
    The visualization comprises three charts, which are coordinated to provide a unified view.
    Initially, a Principal Component Analysis (PCA~\cite{ester1996density}) chart is employed, incorporating the DBSCAN~\cite{abdi2010principal} clustering algorithm to group data points effectively (blue points = inliers, orange points = outliers).
    Subsequently, a box plot is utilized to illustrate the locality, spread, and skewness of the selected attributes. 
    Accompanying this plot is a table that details outliers and their corresponding metadata for further examination.
    
    Lastly, a Scatter Plot Matrix (SPLOM) is presented, enabling users to identify outliers across various attributes. By default, the SPLOM is configured to display crystal density Matthews, resolution, and molecular weight. Users can interact with the visualizations through the brushing and linking technique to investigate specific outliers in greater detail. 
    
    This view helps to understand the variability in the data and identify potential anomalies that may warrant further investigation or correction.
    The Outlier Detection view corresponds has been evaluated and improved thanks to tasks 2 and 3 of the task-oriented user evaluation. A screenshot of this view is provided in Supp.~Figure~5.

    \item \textbf{Database view}: 
    The Database view provides a comprehensive and customizable tabular interface for exploring the enriched database provided by the MetaMP application. 
    This view is designed to provide advanced filtering capabilities, allowing users to refine the dataset according to specific criteria such as taxonomic domain, experimental method, and resolution.
    Users can sort and filter columns to focus on specific subsets of interest, facilitating detailed analysis and comparison. 
    In addition, the view supports exporting filtered data, allowing users to easily extract and use subsets for further analysis or reporting. 
    This functionality increases the accessibility and usability of data, enabling researchers to conduct precise, reproducible, and customized investigations. 
    A screenshot of this view is provided in Supp.~Figure~3.
    
    \item \textbf{Exploration view}:
    This view is designed to facilitate data-driven decision-making and hypothesis generation by allowing users to interactively explore MP structure data. 
    It features a dynamic dashboard with customizable filters and visualization options, allowing users to tailor their analysis to specific research questions or interests. 
    Key components include interactive charts and graphs that show relationships between attributes such as molecular type, experimental method, and taxonomic domain. 
    Users can apply various filters to focus on subsets of data, uncover patterns, and generate insights. 
    This exploratory approach allows researchers to identify trends, correlations, and potential areas for further investigation, enhancing their overall understanding of the enriched data. 
    This is illustrated in Supplementary Figure~6.

    \item \textbf{Grouping view}: 
    The grouping view leverages machine learning (ML) to assist experts in categorizing MP structures into predefined groups based on specified attributes.
    The target groups considered in this work are the three groups mentioned above, as inherited from the MPstruc database.
    While ML provides initial grouping suggestions, researchers actively review these classifications to ensure accuracy and relevance~\cite{dubey2024nested}.
    Therefore, this view allows for combining automated efficiency with potential expert oversight. This collaborative approach improves the analytical process of efficiently organizing data according to predefined criteria, enabling more nuanced data curation. A screenshot of this view is provided in Supplementary~Figure~4.
    \item \textbf{Single-Entry Structural view}: 
    This view combines an interactive 3D protein model with customizable controls and detailed annotations in one browser interface. On the left, users can adjust search type (e.g. PDB, Uniprot, OPM), background color, and toggle display options (e.g., sequence panel, landscape mode). The central canvas renders the structure in cartoon, supporting rotation, zooming, and snapshots. Beneath, a two‐card panel presents core metadata (accession, taxonomy, sequence) alongside computed features (helices, strands, active sites, transmembrane segments), topology predictions, and both expert and ML‐based classifications with confidence scores. 
    The Single-Entry Structural view is
shown in Figure~\ref{figure:singlePageViewFull}.
\end{enumerate}

Besides these views, the MetaMP Homepage serves as a dynamic gateway, providing a concise snapshot of the MetaMP database's composition.
It illustrates the exponential growth of unique MP structures through an interactive timeline, complemented by a trend analysis of experimental methods used over the years. 
Intuitive quick links and a powerful Google-like query field ensure seamless navigation for users of all expertise levels, providing a comprehensive yet accessible entry point to the world of MPs.

\subsubsection*{Data Visualization module}
\label{subsec: data-visualization}
The visualization module VIS of MetaMP uses the powerful Altair~\cite{vanderplas2018altair} plotting library to create interactive and informative visual representations of data. 
Known for its declarative approach to visualization, Altair enables the creation of a wide range of charts and graphs that effectively communicate complex patterns and relationships within the data set. 
This directly supports all of the above views.
MetaMP VIS goes beyond simple static plots to provide users with the ability to explore data through interactive visualizations and linked semantics across charts.
Table~\ref{tab:database-summary} shows the data summary for the VIS and ML modules of MetaMP.

\begin{table}[htbp]
\centering
\caption{
\textbf{Data Summary for the Visualization (VIS) and Machine Learning (ML) modules of MetaMP}. 
This table comprises the number of observations, nominal and quantitative attributes used in each of the two modules.
}
\begin{tabular}{lcccc}
\toprule
\textbf{Database} & \textbf{Rows/Observations} & \textbf{Attributes/Features} & \textbf{Nominal} & \textbf{Quantitative} \\
\midrule
MetaMP VIS & 3569 & 301 & 155 & 146 \\
MetaMP ML & 2849 & 5 & 2 & 3 \\
\bottomrule
\end{tabular}
\label{tab:database-summary}
\end{table}

\subsection*{Artificial Intelligence Modules in MetaMP}

\subsubsection*{AI‐assisted Transmembrane Segment Prediction}

Accurately determining the number and position of transmembrane (TM) segments is a critical precursor to functional annotation, topology-based classification, and database reconciliation. 
To establish a reliable baseline for topology inference, we applied two state-of-the-art AI-based predictors: \textit{TMbed} (v2.0) and \textit{DeepTMHMM} (v1.0.42). Each sequence in the expert-annotated reference set was processed using default parameters, and the predicted number of TM segments was extracted.

To evaluate agreement with expert annotations, we computed the difference:
$ \Delta\mathrm{TM} = \mathrm{TM}_{\mathrm{predicted}} - \mathrm{TM}_{\mathrm{expert}} $
as well as a binary agreement flag indicating exact segment count matches. Predictor performance was assessed via the exact-match rate (\(\Delta\mathrm{TM} = 0\)), mean absolute error (MAE), standard deviation of absolute differences, and correlation metrics (Spearman’s \(\rho\), Pearson’s \(r\)). These same metrics were also applied to pairwise comparisons of TMbed and DeepTMHMM predictions to quantify inter-predictor consistency. Full results are presented in Table~\ref{tab:benchmark} and Supplementary Table 14.

\subsubsection*{AI‐assisted Legacy Database Reconciliation and Topology-Based Classification}

To investigate inconsistencies in public repositories, we reconciled transmembrane segment predictions with historical annotations from OPM and MPstruc for 121 membrane proteins. Segment counts from TMbed and DeepTMHMM were compared to those stored in the legacy records. Discrepant entries—defined as having mismatched segment counts or class labels—were automatically flagged and highlighted in MetaMP’s Discrepancy interface. The Selection and Ranking Views allow users to filter and prioritize these entries for manual review, based on the magnitude of the discrepancy.

In parallel, we implemented a metadata-driven classification model to assign each protein to one of three MPstruc-defined classes: \textit{monotopic}, \textit{alpha-helical transmembrane}, and \textit{beta-barrel transmembrane}. The classifier was trained on structural attributes extracted from OPM, including helix tilt angle, membrane thickness, and subunit span. These features reflect the physical and geometric integration of each protein into the membrane and enable accurate topological classification independently of sequence-based predictors.

\subsubsection*{Machine Learning Module for Structural Group Classification}

To generalize topology-based classification across the full MetaMP dataset, we developed a dedicated machine learning (ML) module composed of four main stages: data preparation, feature selection, semi-supervised model training, and evaluation.

\paragraph{Data Preparation.} We curated a high-quality dataset of 2,849 membrane proteins by applying a structured preprocessing pipeline that included outlier removal, normalization, encoding of categorical variables, and removal of records with missing key attributes.

\paragraph{Feature Selection.} We used a hybrid approach combining manual curation and random forest (RF)-based selection. Non-informative features (e.g., bibliographic metadata) were removed manually. RF-based importance scores were then used to retain features most relevant to structural group classification. This process yielded six numerical and two categorical features. Feature interpretability was supported by Shapley Additive Explanations (SHAP)~\cite{zacharias2022designing}.

\paragraph{Semi-Supervised Learning.} We employed a self-training framework to iteratively expand labeled training data. An ensemble of classifiers—Logistic Regression, Decision Tree, Random Forest, K-Nearest Neighbors, Gradient Boosting, Gaussian Naive Bayes, and SVM—was trained on labeled data, then used to pseudo-label unlabeled entries. These pseudo-labels were reintegrated into the training set over multiple iterations, improving generalization and decision boundaries.

\paragraph{Evaluation.} Model performance was assessed using 5-fold cross-validation. Standard classification metrics (accuracy, precision, recall, F1 score) were computed, with special focus on F1 due to class imbalance. Additional model diagnostics and performance breakdowns are provided in the supplementary materials.

\subsection*{Task-oriented User Evaluation} 
\label{subsec:user-studies}

We conducted a task-oriented user study to evaluate the effectiveness, usability, and intuitiveness of the MetaMP platform. 
MetaMP is designed to integrate a range of functionalities such as summary statistics, outlier detection and identification, analysis of data discrepancies from databases such as OPM and MPstruc, and grouping of MP structures. 
The aim of this study was to test three hypotheses and to evaluate and improve the functionalities of the summary statistics view and the outlier detection view. 
This section comprises the hypothesis testing, the apparatus, the metrics and analysis, the tasks, and the procedure.

\subsubsection*{Hypothesis Testing}
Hypothesis 1. Learning Effectiveness: 
We hypothesize that the training phase effectively equips users with skills to perform better in the testing phase, assuming that familiarity with the tasks and the system leads to faster completion times. 
To evaluate this, we compared average task completion times between the training and test phases using a paired Student's t-test to determine statistical significance in time reduction.

Hypothesis 2. Task difficulty: 
This hypothesis suggests that task completion time will vary significantly based on task complexity, regardless of phase. 
To investigate, we calculated average completion times for each task and conducted an ANOVA test to determine statistically significant differences between tasks, analyzing three different tasks categorized into training and testing phases.

Hypothesis 3. Accuracy vs. Speed Trade-off: 
We propose a trade-off between speed and accuracy, where users who prioritize speed may be more error prone, while those who are more deliberate may achieve higher accuracy. 
To test this, we performed a logistic regression analysis assessing the relationship between task completion time and response accuracy, using time as the independent variable and accuracy as the dependent variable. 
A scatterplot with a regression line was used to visualize this relationship.

\subsubsection*{Apparatus}

The study was conducted entirely online using the MetaMP platform. 
To facilitate a thorough evaluation, we integrated a custom-built survey module into MetaMP. 
This customization allowed us to create a seamless experience for participants, covering all aspects of the study, including participant onboarding, socio-demographic data collection, training sessions, task execution, and usability evaluation.

Participants were asked to complete a series of three sequential tasks that included generating summary statistics and identifying outliers. 
These tasks were strategically designed to take advantage of MetaMP's intuitive features for analyzing membrane protein structures and to highlight the valuable insights that can be gained using the platform. 
MetaMP's interactive charts and tools have been specifically designed to appeal to both expert and non-expert users, ensuring that the platform remains accessible and user-friendly to a wide range of participants.

\subsubsection*{Metrics and Analysis}
We collected socio-demographic and task-relevant data for each participant.
The socio-demographic data included: gender, years of experience, current status (student or professional), and domain.
The task-relevant data included: System usability scale~(SUS)~\cite{hyzy2022system}, time to completion, number of clicks, and optional feedback.

\subsubsection*{Tasks}
The task-based evaluation consisted of three consecutive tasks, each with a training and testing phase.
MetaMP provides instructions, cues, hints, and sometimes screenshots for each training question to support user learning. 
There was no time limit for training or testing, and participants answered the questions with the help of data visualizations. 
The training questions were designed to help participants answer the test questions correctly. 
A workflow tour was also provided to familiarize participants with the layout and features of the MetaMP user evaluation module. 
Correct answers were provided for all of the training questions, but not for the test questions.
The list of questions used during the evaluation is given in Table~\ref{table:usability-tests}.

Task 1. Summary Statistics:
The first section of the test required participants to analyze interactive visualizations to assess the temporal growth of experimental methods used to resolve membrane protein structures. Participants were asked to identify trends over time, including the relative progress of different experimental methods. The second question focused on identifying the experimental method that is currently advancing the fastest, represented by a line graph showing the growth trajectories of different methods over the years.

Task 2. Outlier identification:
In the outlier identification section, participants were presented with four questions - two designed as training problems and two designed as test problems. The first training question involved identifying outliers in the resolution of MPs in terms of their groups, such as monotopic, alpha-helical transmembrane proteins, and beta-barrel transmembrane proteins. The second training task required participants to identify a specific MP within the monotopic group that was an outlier compared to other data points in the group using a box plot visualization.

Task 3. Outlier detection:
The final section involved comparing three different visualization methods - scatter plot, SPLOM plot, and box plot- to evaluate the accuracy of outlier identification. Participants were asked to check whether the outlier detection plot matched the points highlighted in the SPLOM plot based on the selected features or attributes. The interactive nature of the task allowed participants to click and drag over specific areas of the graphs to highlight and compare data points between the different visualizations. This comparison was essential for evaluating the effectiveness of MetaMP's outlier detection algorithm.

\begin{table}[htbp]
\centering
\caption{
\textbf{List of the six different questions used in the task-based user evaluation}. 
Each task followed a two-step process: training, then testing.}
\label{table:usability-tests}
\begin{tabularx}{\textwidth}{Xc}
\toprule
\textbf{Questions} & \textbf{Type}\\
\midrule
\textbf{Task 1. Summary Statistics} & \\
1. Which method appears to be most used? & Training  \\
2. Which experimental method appears to be growing faster now? & Test \\
\hline
\textbf{Task 2. Outlier Identification} & \\
3. Identify membrane protein structure groups that contain outliers. & Training \\
4. Study the variations in resolution values using electron microscopy (EM), specifically focusing on the initial group in the box plot illustrating Monotopic Membrane Protein Structures. How many outliers are evident within this context? & Test \\
\textbf{Task 3. Outlier Detection} & \\

5. How many outliers in the scatter plot matrix (SPLOM) were not identified by MetaMP using the DBSCAN? & Training \\
6. Do you observe any outliers in the SPLOM that MetaMP failed to detect using DBSCAN? & Test \\
\bottomrule
\end{tabularx}
\end{table}

\subsubsection*{Procedure}
All participants P1 to P24 were formally invited by e-mail. 
Two case studies were defined using MetaMP to design the task-oriented user evaluation. 
Each resulted in its own enriched dataset and was integrated as part of the user evaluation. 
Participants were asked to answer both training and test questions.
At the end of the study, participants were asked to complete a post-study questionnaire.
This questionnaire included a demographic and experience form as well as the System Usability Scale (SUS)~\cite{hyzy2022system}. 
The SUS section consisted of 14 questions (see the supplementary Material for the SUS questions) with responses on a scale of 1-5, where 1 indicates ``strongly disagree'' and 5 indicates ``strongly agree'' with the statements. The System Usability Scale was used to assess the subjective usability of the summary statistics view and the outlier detection view. 
The System Usability Scale (SUS) score is calculated using the following steps: For each question \(i\) from 1 to 10, the score \(S_i\) is calculated as \(Q_i - 1\) if \(i\) is even, and as \(5 - Q_i\) if \(i\) is odd. The sums for even and odd questions are then computed as \(\text{Sum}_E = \sum_{\text{even } i} (Q_i - 1)\) and \(\text{Sum}_O = \sum_{\text{odd } i} (5 - Q_i)\), respectively. The System Usability Scale (SUS) score is given by \(\text{SUS Score} = 2.5 \times (\text{Sum}_E - 5 \times |E| + 25 - \text{Sum}_O)\), where the number of even and odd questions (\(|E|\) and \(|O|\)) is 5.
This formula converts the individual responses into a single SUS score ranging from 0 to 100. The usability scores were categorized as follows: a score of less than 50 was considered ``poor,'' 50 to 69 was considered ``acceptable,'' 70 to 84 was considered ``good,'' and a score above 84 was considered ``excellent.

\subsection*{Discussion}

This paper presents a web-based computer application that provides researchers with access to a range of visualizations, thereby facilitating the maintenance of data integrity and increasing the reliability of scientific results. 

The investigation of missing records in the Protein Data Bank (PDB) revealed instances where records were either under review or had been updated.
This is confirmed by an alternative PDB endpoint (\href{https://www.rcsb.org/structure/removed/5W7L}{PDB entry for 5W7L}). 
These findings and other examples of affected proteins are detailed in the supplementary material.
In addition, we observed entries that remained unchanged or unallocated, marked as ``unreleased depositions withdrawn (WDRN)'', a discrepancy that may conflict with the information presented in the MPstruc database. 
Notable examples of membrane proteins affected by these discrepancies include 7UUV and 7ROW, which can be reviewed at (\href{https://www.rcsb.org/structure/unreleased/7ROW}{Unreleased PDB Entry for 7ROW}).

Second analyses on MetaMP confirm the advances in cryo-EM resolution often referred to as the "resolution revolution"~\cite{kuhlbrandt2014resolution, hong2023cryo, burley2022electron}. 
This revolution has been driven by advances in transmission electron microscope optics, direct detector technology, image processing algorithms, and grid preparation methods~\cite{kuhlbrandt2014resolution}. 
Cryo-EM has become the dominant method for resolving membrane protein structures, surpassing traditional methods such as nuclear magnetic resonance (NMR) and X-ray crystallography, as shown in Figure~\ref{fig:homepage} and the supplementary material.

Third, MP classification varies widely across databases and domain experts. 
MetaMP adopted the three MP types from MPstruc for machine learning to assist domain experts in the task of classification. However, there is a need for general agreement in the research community on refined classifications.
For example, the OPM database classifies quaternary complexes based on their major domains within membranes, using information from SCOP and TCDB, but with notable differences~\cite{uniprot2023uniprot}. 
It organizes these complexes into four hierarchical levels: Type, Class, Superfamily, and Family. 
The type level includes categories such as transmembrane proteins, monotopic proteins, and membrane-active peptides. 
The class level includes structural classifications such as all-\(\alpha\), all-\(\beta\), and mixed structures. 
The superfamily level groups proteins with similar 3D structures based on evolutionary relatedness, and the family level groups proteins with detectable sequence homology.
Such groupings make a lot of sense and support specific tasks. 
Based on the results of the AI use cases involving TMbed and DeepTMHMM, we observed that the results generally align well with the expert annotations, with a few exceptions.
These minor discrepancies may reflect differences in algorithmic interpretation or limitations in the original expert annotations rather than inherent ambiguity about the protein classification.

Fourth, the semi-supervised learning model will perform better with an expanded training dataset. 
Regular updates to our database will further improve its predictive accuracy. 
However, due to inconsistencies between databases such as OPM and MPstruc, it is important to encourage and coordinate communication between all membrane protein databases to increase the reliability of our predictions and overall performance. 
The use of machine learning can also help categorize proteins into specific subgroups or taxonomic domains, thereby streamlining the data curation process and assisting domain experts in their efforts.
Although cross-validation was used, there is no golden standard.
In fact, certain MP structures can be either monotopic or bitopic depending of their environment.
The current cross-validation results are considered sufficient, and in the future, multiple human experts in the loop is very much needed.

Fifth, many MP structures are currently under- or over-represented in the database, because of the disease-related variable. It is currently possible to search for specific diseases using the Google-like query field on the MetaMP homepage. However, this is by no means comprehensive and is inherited from the UniProt database. 
Further work will be required in future versions to integrate databases such as Orphanet for rare diseases, thus increasing the interest of MetaMP to a wider audience.

Sixth, information on surfactant usage for MP structure determination has not been included. 
The reason for this is that we will get a biased representation of MP structures resolved by X-ray, as this is the only method for which we have data.
Other lists will have to be included, for example for NMR.

Seventh, MPs are not always resolved from the first to the last amino acid.
We currently do not record this sequence resolution information. 
However, MetaMP provides the size from both the PDB and UniProt databases, which are known to differ.

Eighth, the advancement of MP research hinges on the development of more comprehensive and integrative databases that incorporate critical metadata, including for instance information on protein folding and misfolding after production. 
By fusing fluorescent proteins such as mCherry or mVenus to MPs, scientists can follow their entire life-cycle in real time, from synthesis and insertion into membranes to degradation. 
This technique allows the visualization of critical processes such as protein trafficking, localization and interactions within living cells and within cell populations~\cite{goulian2000tracking, hattab2018novel}.
This integrative approach is essential for solving MP folding problems, as misfolded proteins can disrupt cellular function. 
A comprehensive database would improve the prediction and manipulation of MP behavior, potentially transforming drug discovery.

Ninth and last, we can predict that more applications will be powered by artificial intelligence and machine learning to assist human experts in their tasks, and possibly even make suggestions to users about the discrepancies they encounter during data curation.

\section*{Conclusion}
In summary, the MetaMP platform has demonstrated significant potential to improve the integrity and reliability of MP structure analysis through its various modules. 
The task-oriented user evaluation and database audits have highlighted the critical need for continued refinement of these tools to further establish MetaMP as an indispensable resource in the scientific community.
Our findings underscore the importance of rigorous data validation and collaboration among existing databases. 
This collaborative effort is essential to maintaining data consistency and fostering a more robust scientific research process.
%
Continuous improvements and new methods will meet the evolving needs of the scientific community. 
MetaMP will integrate other MP databases for further enrichment and community feedback. 
Underpinning MetaMP's success is a transparent and open culture that encourages expert feedback and feature requests via the landing page. 
This commitment to continuous improvement and user-driven development ensures that MetaMP remains at the forefront of membrane protein structure analysis, driving advances in structural biology and its applications beyond.

\section*{Supplementary Materials}

MetaMP is a comprehensive web application built using Vue.js for dynamic front-end development and Altair for effective data visualization~\cite{lavanya2023assessing}. 
The backend code and data for MetaMP can be accessed publicly on GitHub at \url{https://github.com/Ebenco36/MetaMP-Server}, while the front-end codebase is available at \url{https://github.com/Ebenco36/MetaMP.git}. 
A standalone prototype of \textbf{MetaMP} can be deployed using Docker as explained on the Github repository of MetaMP: \url{https://github.com/Ebenco36/MetaMP-Server?tab=readme-ov-file#installation-and-running}.
A minimal prototype of \textbf{MetaMP} can be viewed online at \url{https://mpvisualization-1w5i.onrender.com/}.
All supplementary materials, including source code, datasets, data generation scripts, and detailed instructions, are also provided.

\section*{Acknowledgments}
We would like to thank all the members of the Center for Artificial Intelligence, at the Robert Koch Institute, and the Laboratory of Physical and the Biochemistry of Membrane Proteins at the French National Center for Scientific Research (CNRS), in Paris, France, for providing the facilities, insights, and resources essential to this study. 
The authors acknowledge support from the French National
Research Agency (ANR) through LABEX DYNAMO (ANR-11-LABX 0011).
We also acknowledge the financial support of the Center for Artificial Intelligence and the collegial support of the Visualization Group members.

We would also like to thank the following individuals for their specific contributions to this research by participating in the survey: Dr.~Aleksandar Anžel, Ana Paula Gomes Ferreira, Akshat Dubey, Andre Jatmiko Wijaya, Dr.~Tunde Asiyanbi, Isaac Dunga, Dr.~Fowotade Itunuoluwa, Dr.~Zewen Yang, Oluwatobiloba Davies, and Blessing Makaraba. Special recognition is given to Dr.~Aleksandar Anžel, as well as to the peer reviewers, whose constructive feedback significantly improved the quality of this paper.


\section*{Author contributions statement}
\noindent Ebenezer Awotoro: Responsible for implementation and writing of the initial draft.
Georges Hattab: Developed the research concept, supervised the project, edited and wrote sections of the manuscript.
Chisom Ezekannagha: Conducted a paired review of the manuscript.
Dominik Heider: Conducted a review of the manuscript, providing expert feedback.
Katharina Ladewig: Conducted a review of the manuscript, providing expert feedback.
Christel Le Bon: Conducted a review of the manuscript, providing expert feedback.
Karine Moncoq: Conducted a review of the manuscript, providing expert feedback.
Bruno Miroux: Conducted a review of the manuscript, providing expert feedback.
Johannes Tauscher: Initiated the project during their B.Sc. studies by developing a minimal working example of a front-end specifically for MPstruc metadata as a direct foundation to the Exploration view.
Florian Schwarz: Enhanced the project by integrating the Protein Data Bank (PDB) with MPstruc, and wrote a Jupyter notebook to facilitate this integration.

\section*{Data availability}
The data supporting the findings of this study are available at  
\href{https://drive.google.com/file/d/1QXfQYis08zKc_PdqQtQroegcRbradIrR/view?usp=sharing}{MetaMP Datasets}.

%
\section*{Conflicts of Interest}

The authors declare no conflicts of interest. The sponsors had no involvement in the conception or design of the study, data acquisition, analysis, or interpretation, the drafting or revision of the manuscript, or the decision to submit the work for publication.

\bibliography{main}


\end{document}


\vspace*{7cm}
\section*{Supplementary Material. \\
\\
Enabling AI Applications for Membrane Proteins with a Unified Framework for Machine Learning, Visualization, and Metadata Enrichment
\\Awotoro \textit{et al. }}

\begin{figure}
\centering
\includegraphics[width=1\textwidth]{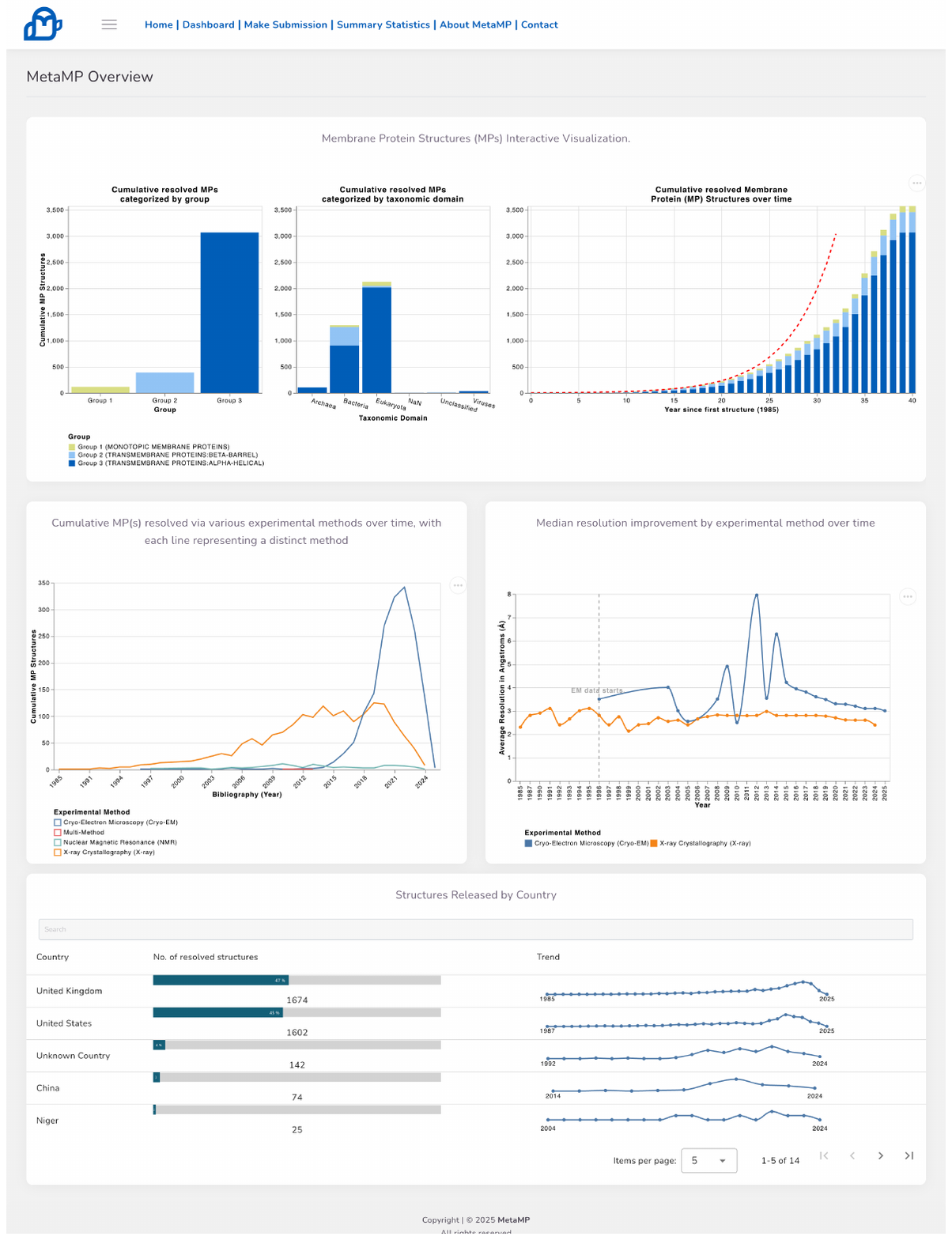}

\caption{\textbf{Overview}
    \\
    This view provides a detailed and interactive visualization of membrane protein (MP) structures. It categorizes cumulative MP data by groups, taxonomic domains, and historical trends, providing insights into structural progress and advancements in experimental methods, including median resolution improvements over time. The overview also presents country-specific contributions with trend analyses, serving as a comprehensive resource for researchers to explore MP datasets, identify patterns, and track technological milestones.
  }
\label{figure:dashboard}
\end{figure}

\begin{figure}
\centering
\includegraphics[width=.9\textwidth]{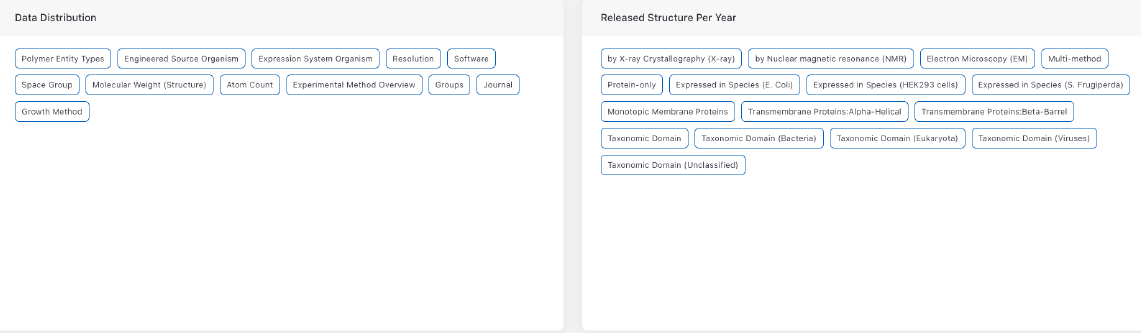}
\par
\includegraphics[width=.9\textwidth]{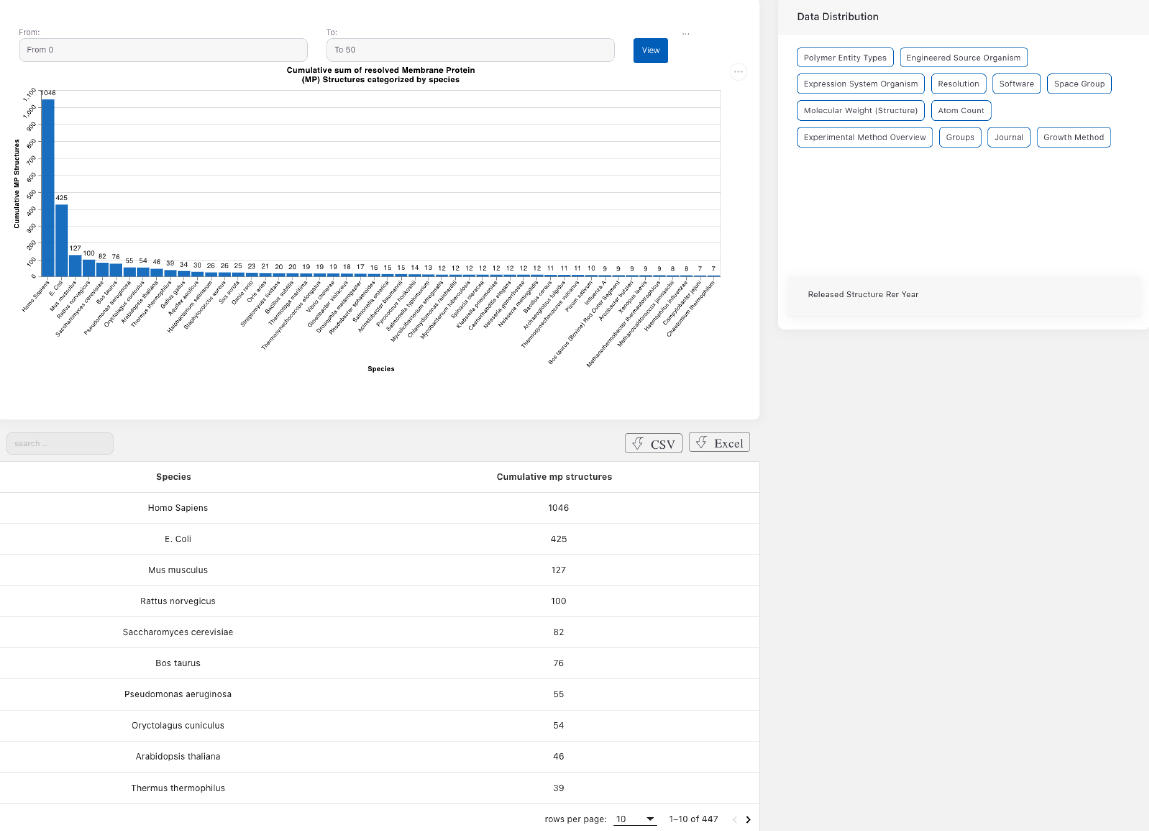}
\caption{\textbf{Summary Statistics View}. This view provides an organized breakdown of membrane protein (MP) data, facilitating exploration through two main sections. First, the Data Distribution panel categorizes key attributes such as polymer entity types, experimental methods, molecular structure details, and growth methods. Second, the Released Structure Per Year panel highlights structural data trends over time, segmented by experimental methods, taxonomic domains, and protein-specific attributes. Together, these sections offer a comprehensive overview of MP dataset statistics, enabling detailed insights into data characteristics and trends.}
\label{figure:summary-view}
\end{figure}

\begin{figure}
\centering
\includegraphics[width=.9\textwidth]{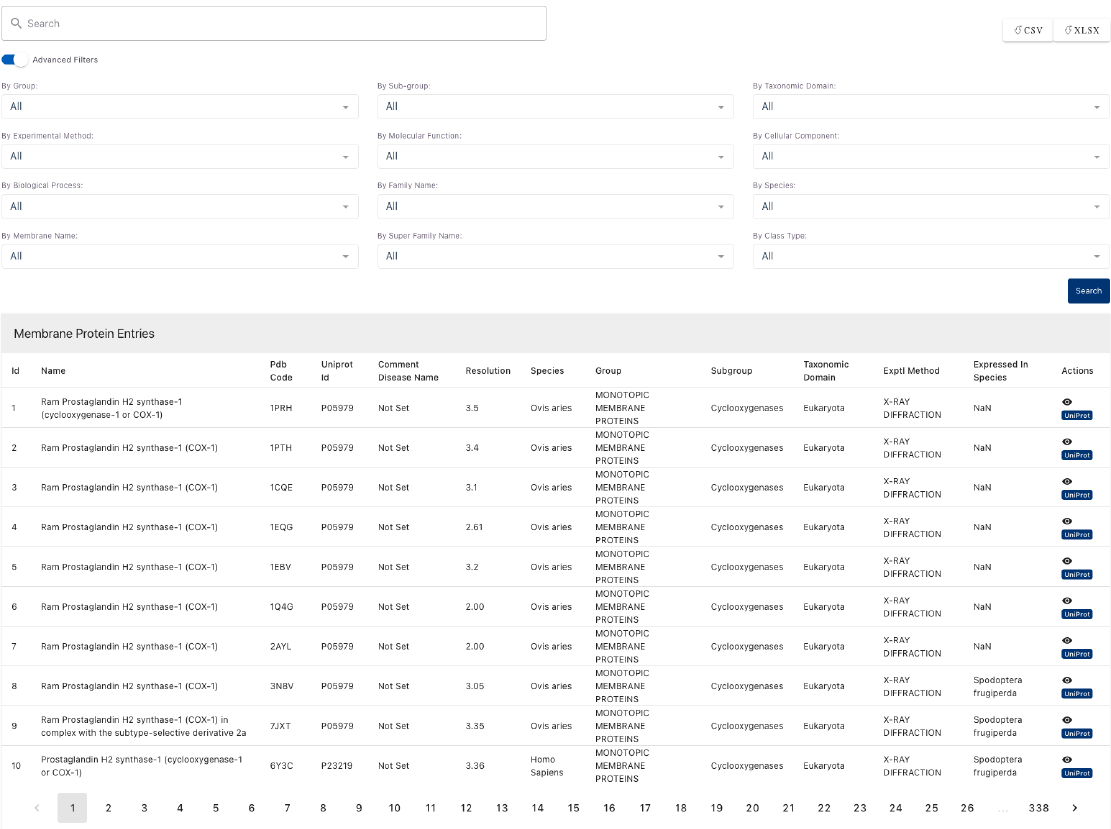}
\caption{\textbf{Database View}. This view provides a detailed table of membrane protein (MP) entries, allowing researchers to explore data with advanced search and filter options. Each entry includes key attributes such as protein name, PDB code, UniProt ID, resolution, species, group, subgroup, taxonomic domain, and experimental method. Users can export the data in CSV or XLSX formats and access related information through direct links to UniProt. The intuitive pagination and filtering features ensure efficient navigation of the extensive dataset, supporting comprehensive data exploration and analysis.}
\label{figure:database-view}
\end{figure}

\begin{figure}
\centering
\includegraphics[width=.9\textwidth]{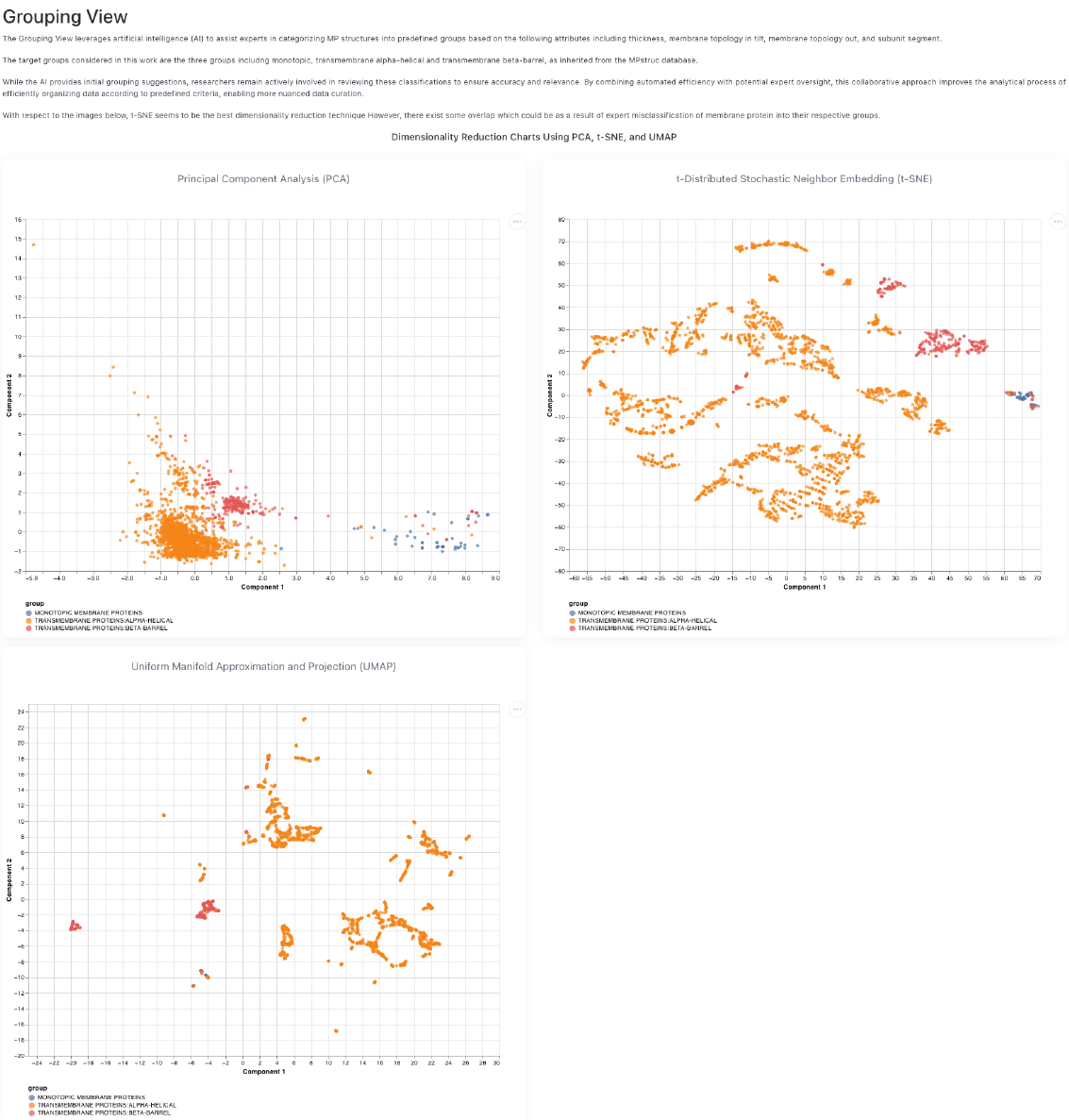}
\caption{\textbf{Grouping View}. This view offers a comprehensive analysis of membrane protein structures using dimensionality reduction techniques, like Principal Component Analysis (PCA), t-Distributed Stochastic Neighbor Embedding (t-SNE), and Uniform Manifold Approximation and Projection (UMAP). These methods categorize membrane proteins into groups such as "Transmembrane Membrane Proteins Alpha Helical", "Transmembrane Membrane Proteins Beta Barrel", and "Monotopic Membrane Protein." The visualizations reveal clusters and separations within the data, providing valuable insights into structural relationships.}
\label{figure:grouping-view}
\end{figure}

\begin{figure}
\centering
\includegraphics[width=.85\textwidth]{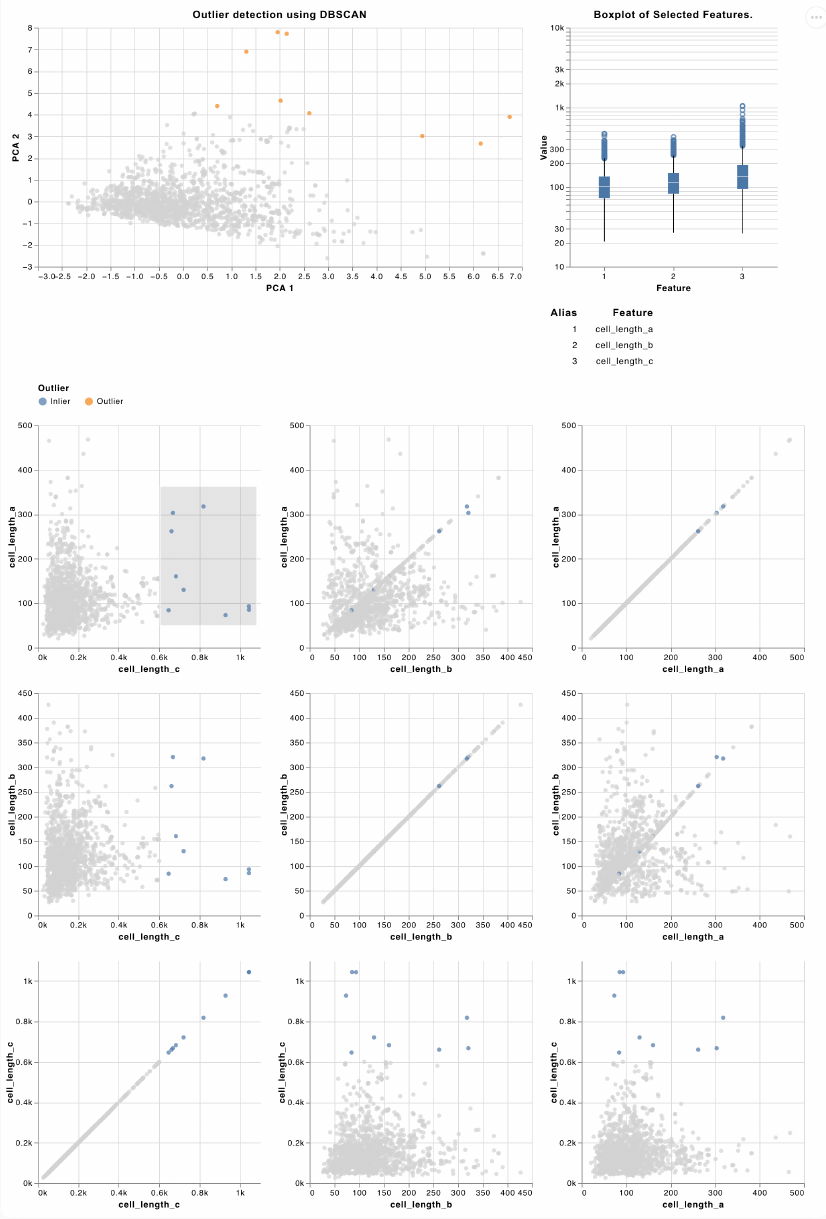}
\caption{\textbf{Outlier Detection View}. This view provides visualizations for analyzing patterns and identifying outliers across dimensions, including a DBSCAN-based scatter plot with PCA components, box plots of feature distributions, and pairwise scatter plots (e.g., \texttt{cell\_length\_a}, \texttt{cell\_length\_b}, \texttt{cell\_length\_c}), offering insights into data structure and outlier behavior.}
\label{figure:outlier-detection-view}
\end{figure}

\begin{figure}[htbp]
\centering
\includegraphics[width=\textwidth]{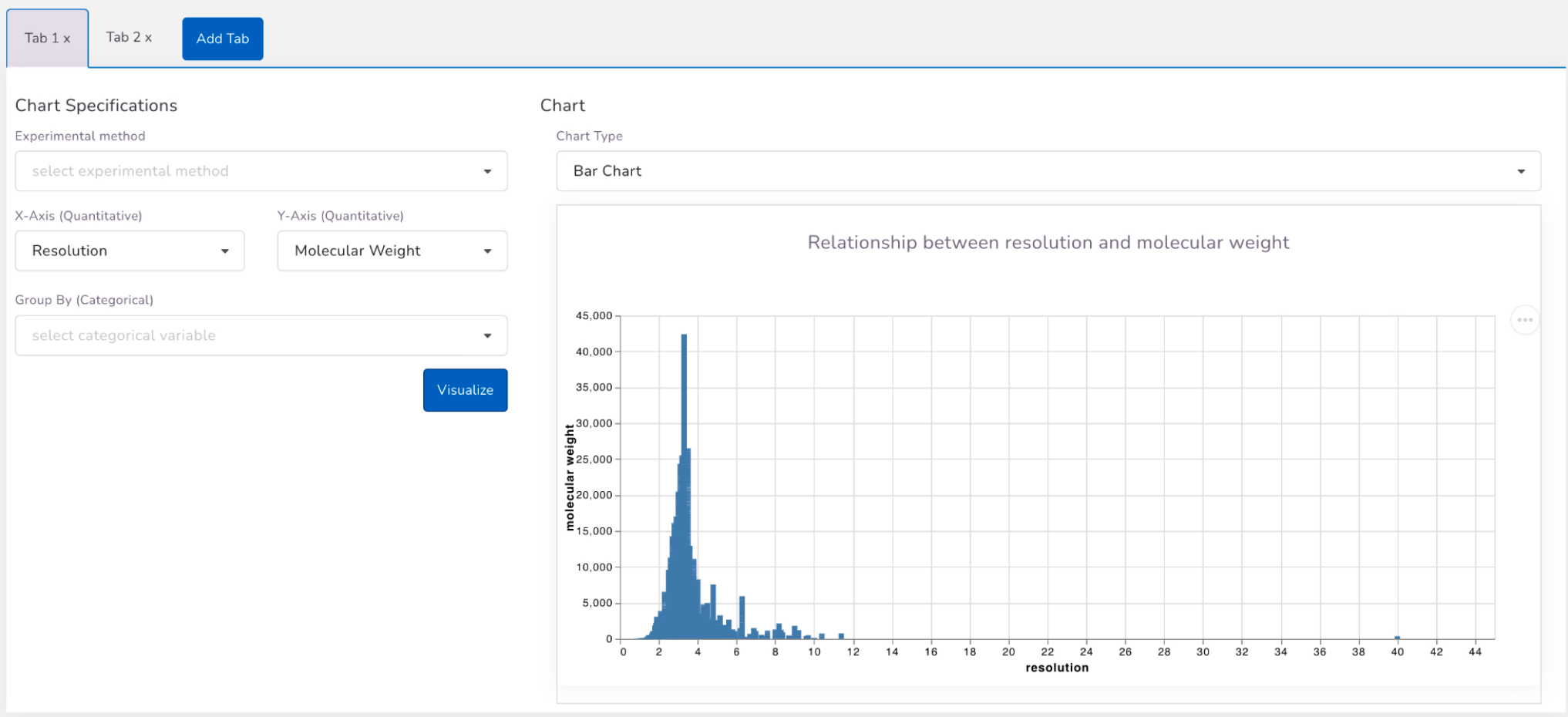}
\caption{
\textbf{Screenshot of the Exploration view. }
One of the eight views provided in MetaMP, the Exploration view facilitates the analysis of membrane protein metadata through an interactive dashboard. 
It enables researchers to visualize relationships between key attributes such as molecular classification, experimental methods, and taxonomic domains. 
This example shows the molecular weight as a function of resolution using a bar graph.
MetaMP dynamic filtering capabilities allow focused examination of specific subsets of data, facilitating subsequent steps such as the identification of statistically significant patterns and correlations. 
By providing tools for hypothesis testing, this view supports evidence-based research strategies.
and enhances the potential for new insights in membrane protein research. 
The view design prioritizes data integrity, reproducibility, and user interaction to ensure data-driven exploration.
}
\label{figure:explorationview}
\end{figure}

\begin{table}
    \centering
    \caption{\textbf{Mean resolution of cryo-EM and X-ray crystallographic membrane protein structures}.}
    \begin{tabular}{ccc}
        \toprule
        Year & Cryo-Electron Microscopy (Å) & X-ray Crystallography (Å) \\
        \midrule
        1985 & N/A & 2.3±(n = 1) \\
        1987 & N/A & 2.8±(n = 1) \\
        1990 & N/A & 2.9±(n = 1) \\
        1991 & N/A & 3.1±(n = 1) \\
        1992 & N/A & 2.4±0.6 \\
        1993 & N/A & 2.65±0.07 \\
        1994 & N/A & 2.86±0.56 \\
        1995 & N/A & 3.1±0.25 \\
        1996 & 3.5±(n = 1) & 2.64±0.66 \\
        1997 & N/A & 2.58±0.52 \\
        1998 & N/A & 2.72±0.45 \\
        1999 & N/A & 2.39±0.69 \\
        2000 & N/A & 2.44±0.49 \\
        2001 & N/A & 2.48±0.51 \\
        2002 & N/A & 2.6±0.67 \\
        2003 & 4.0±(n = 1) & 2.74±0.7 \\
        2004 & 3.0±(n = 1) & 2.6±0.53 \\
        2005 & 2.55±0.92 & 2.53±0.63 \\
        2006 & N/A & 2.68±0.65 \\
        2007 & N/A & 2.77±0.64 \\
        2008 & 3.5±(n = 1) & 2.76±0.7 \\
        2009 & 4.9±2.97 & 2.72±0.71 \\
        2010 & 2.5±(n = 1) & 2.79±0.66 \\
        2011 & N/A & 2.69±0.67 \\
        2012 & 7.95±2.47 & 2.72±0.53 \\
        2013 & 3.54±0.37 & 2.91±0.65 \\
        2014 & 6.59±3.08 & 2.83±0.6 \\
        2015 & 4.25±0.93 & 2.66±0.63 \\
        2016 & 4.69±1.52 & 2.8±0.7 \\
        2017 & 4.12±1.12 & 2.79±0.56 \\
        2018 & 3.81±0.98 & 2.77±0.71 \\
        2019 & 3.68±0.99 & 2.7±0.55 \\
        2020 & 3.47±0.89 & 2.71±0.63 \\
        2021 & 3.51±2.27 & 2.64±0.78 \\
        2022 & 3.23±0.62 & 2.64±0.7 \\
        2023 & 3.22±0.68 & 2.5±0.63 \\
        2024 & 3.17±0.39 & N/A \\
        \bottomrule
    \end{tabular}
    \label{table:1}
\end{table}

\begin{table}
    \centering
    \caption{\textbf{MetaMP Updates to PDB Codes}. This table shows the automatic update made to a set of PDB accession codes.}
    \begin{tabular}{cc}
        \toprule
        Old PDB code & New PDB code \\
        \midrule
        5W7L & 8G1N \\
        3WXV & 6KS0 \\
        3J8E & 5TB0 \\
        3HGC & 4NYK \\
        4UPC & 5A63 \\
        6FFV & 8C7P \\
        6AN7 & 6OIH \\
        5TSI & 5UAR \\
        4J05 & 7SP5 \\
        3B8C & 5KSD \\
        4P6V & 8ACY \\
        1FUM & 1L0V \\
        3BZ1 & 4V62 \\
        3ARC & 3WU2 \\
        3CJU & 3EGV \\
        5G1J & 7PDC \\
        \bottomrule
    \end{tabular}
    \label{table:5}
\end{table}

\clearpage

\begin{figure}
\centering
\includegraphics[width=\textwidth]{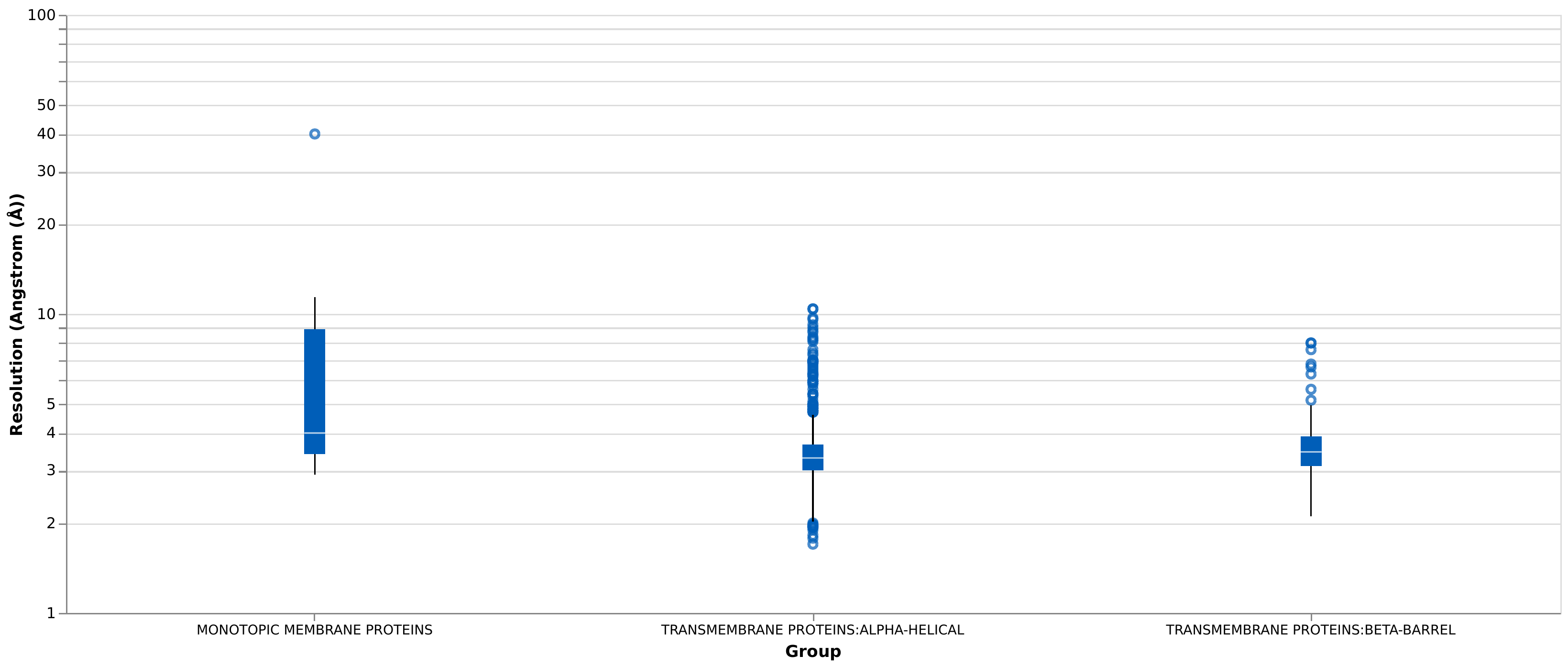}
\caption{\textbf{Outliers in Cryo-electron microscopy}.
The boxplot illustrates outliers in the Cryo-electron microscopy (EM) analysis. Notably, the protein coded 6ZG5 is identified as an outlier among MONOTOPIC MEMBRANE PROTEINS, displaying an unusually high resolution of 40 Ångströms (Å). This measurement markedly exceeds the typical resolution range observed in the dataset, highlighting the exceptional nature of this protein's resolution.}
\label{figure:EMOutlier}
\end{figure}

\begin{figure}
\centering
\includegraphics[width=\textwidth]{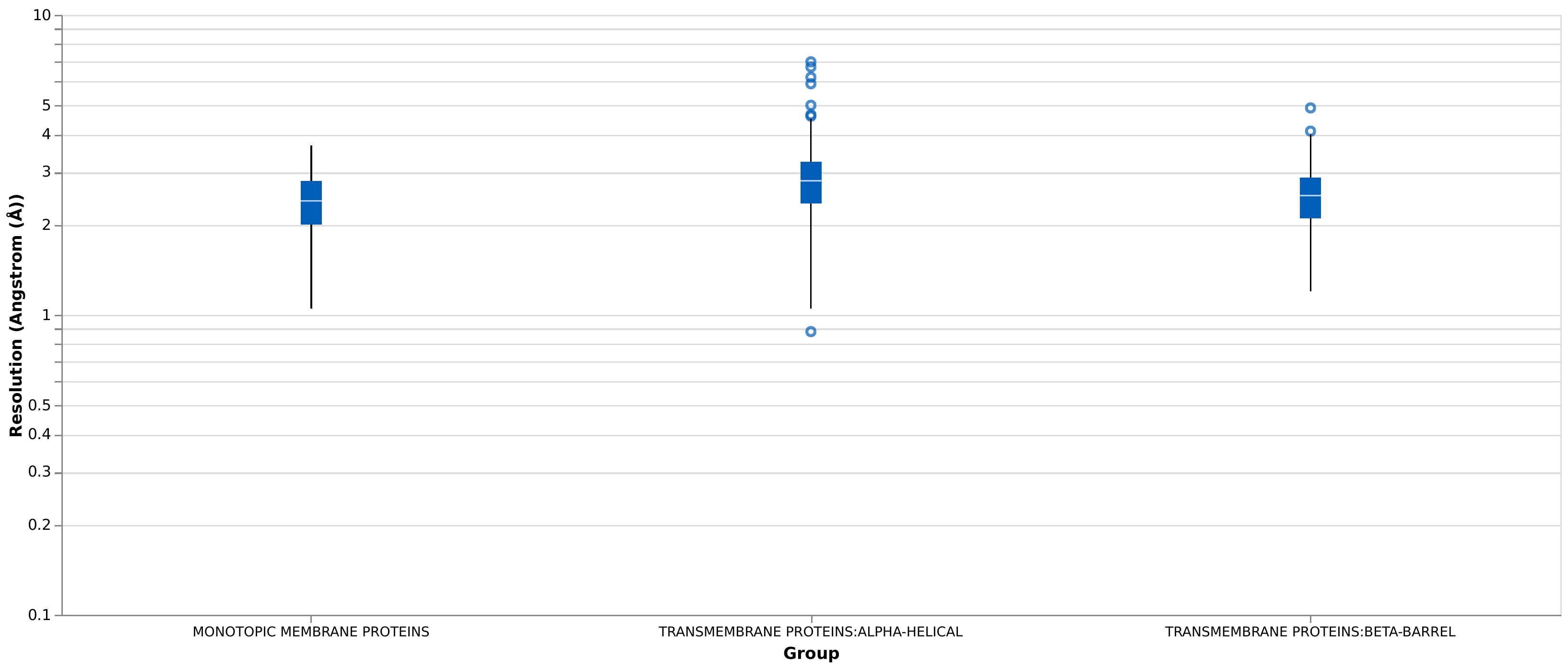}
\caption{\textbf{Outliers in X-ray crystallography}. The boxplot shows the presence of numerous outliers in X-ray crystallography data for both TRANSMEMBRANE ALPHA-HELICAL and TRANSMEMBRANE BETA-BARREL proteins. The analysis highlights several instances where the resolution measurements significantly deviate from the typical range observed in these MP groups.}
\label{figure:X-rayOutlier}
\end{figure}

\clearpage

\begin{figure}
\centering
\includegraphics[width=\textwidth]{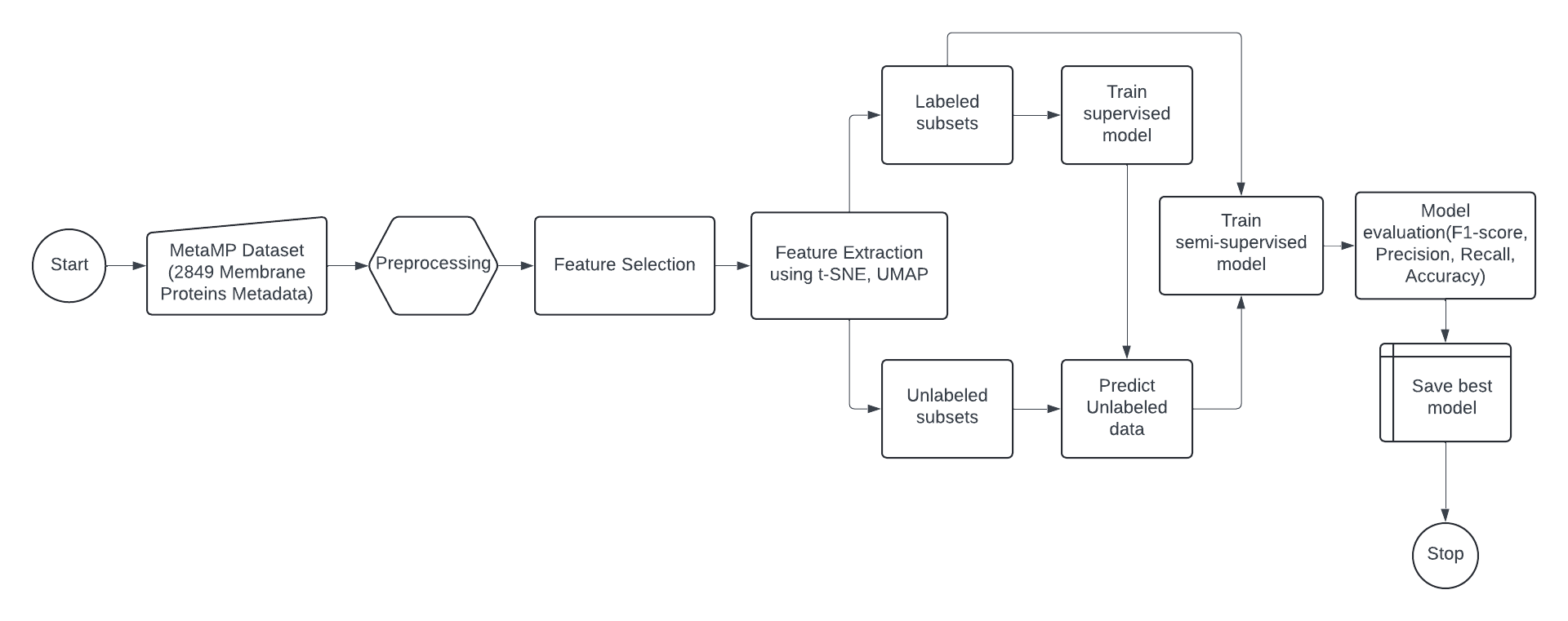}
\caption{\textbf{Semi-supervision learning pipeline}. 
The pipeline integrates labeled and unlabeled data to improve the learning model. It details the use of supervised learning techniques for training the model while employing unsupervised methods for dimensionality reduction to enable visualization in a 2D space. 
Additionally, feature selection is utilized to identify the most relevant features for prediction. This approach enhances model performance by effectively combining the strengths of both labeled and unlabeled data, optimizing feature relevance, and facilitating clearer data visualization.}
\label{fig:Semi-supervised-learning}
\end{figure}

\begin{table}
\centering
\caption{\textbf{Model Performance}.
Performance metrics for various classifiers, presented as mean values with standard deviations, based on 5-fold cross-validation. The metrics include mean accuracy, F1-score, precision, and recall, along with their respective standard deviations. The Random Forest model achieves the highest mean accuracy and F1-score, and also shows strong precision and recall, highlighting its superior performance and consistency compared to other classifiers.}
\begin{tabular}{lcccccccc}
\toprule
Classifier & Accuracy (mean $\pm$ std) & F1-Score (mean $\pm$ std) & Precision (mean $\pm$ std) & Recall (mean $\pm$ std) \\
\midrule
Logistic Regression & $0.959 \pm 0.008$ & $0.946 \pm 0.008$ & $0.935 \pm 0.008$ & $0.959 \pm 0.008$ \\
Decision Tree & $0.971 \pm 0.007$ & $0.971 \pm 0.007$ & $0.971 \pm 0.007$ & $0.971 \pm 0.007$ \\
Random Forest & $0.977 \pm 0.005$ & $0.976 \pm 0.005$ & $0.977 \pm 0.005$ & $0.977 \pm 0.005$ \\
KNeighbors Classifier & $0.974 \pm 0.011$ & $0.974 \pm 0.010$ & $0.976 \pm 0.011$ & $0.974 \pm 0.011$ \\
Gradient Boosting Classifier & $0.975 \pm 0.004$ & $0.974 \pm 0.004$ & $0.974 \pm 0.004$ & $0.975 \pm 0.004$ \\
Gaussian NB & $0.949 \pm 0.006$ & $0.935 \pm 0.007$ & $0.926 \pm 0.005$ & $0.949 \pm 0.006$ \\
SVM & $0.958 \pm 0.008$ & $0.947 \pm 0.008$ & $0.937 \pm 0.009$ & $0.958 \pm 0.008$ \\
\bottomrule
\end{tabular}
\label{tab:performance_metrics}
\end{table}

\begin{table}[ht]
  \centering
  \caption{Model Performance. Supervised vs Semi-Supervised Machine Learning Performance Metrics.}
  \label{table:4}
  \begin{tikzpicture}
    \matrix (tbl) [%
      matrix of nodes,
      nodes in empty cells,
      nodes={inner sep=1ex, align=center},      
      column sep=2em,                            
      row sep=-\pgflinewidth,                   
      column 2/.style={nodes={align=left,anchor=west}}
    ] {
      \bfseries Category       & \bfseries Classifier            & \bfseries Accuracy & \bfseries Precision & \bfseries Recall & \bfseries F1-score \\
      Supervised               & Logistic Regression            & 0.956 & 0.936 & 0.956 & 0.945 \\
                              & Decision Tree                  & 0.974 & 0.978 & 0.974 & 0.976 \\
                              & Random Forest                  & 0.973 & 0.975 & 0.973 & 0.974 \\
                              & KNeighbors Classifier          & 0.983 & 0.985 & 0.983 & 0.984 \\
                              & Gradient Boosting Classifier   & 0.974 & 0.976 & 0.974 & 0.975 \\
                              & Gaussian NB                    & 0.956 & 0.937 & 0.956 & 0.945 \\
                              & SVM                            & 0.966 & 0.973 & 0.966 & 0.968 \\
      Semi-Supervised          & Logistic Regression            & 0.955 & 0.931 & 0.955 & 0.942 \\
                              & Decision Tree                  & 0.970 & 0.978 & 0.970 & 0.972 \\
                              & Random Forest                  & 0.985 & 0.986 & 0.985 & 0.985 \\
                              & KNeighbors Classifier          & 0.985 & 0.985 & 0.985 & 0.985 \\
                              & Gradient Boosting Classifier   & 0.985 & 0.988 & 0.985 & 0.986 \\
                              & Gaussian NB                    & 0.950 & 0.927 & 0.950 & 0.936 \\
                              & SVM                            & 0.975 & 0.976 & 0.975 & 0.975 \\
    };

    \draw[line width=1pt]   (tbl-1-1.north west) -- (tbl-1-6.north east);
    \draw[line width=1pt]   (tbl-1-1.south west) -- (tbl-1-6.south east);
    \draw[line width=0.5pt] (tbl-8-1.south west) -- (tbl-8-6.south east);
    \draw[line width=1pt]   (tbl-15-1.south west) -- (tbl-15-6.south east);

    \draw[red,thick,rounded corners]
      (tbl-9-1.north west) rectangle (tbl-15-6.south east);
  \end{tikzpicture}
\end{table}

\begin{table}
    \centering
    \caption{\textbf{Task metrics}. Table showing the mean, median, and variance metrics for each task.}
    \label{table:4}
    \begin{tabular}{|l|c|c|c|}
        \hline
        \textbf{Task} & \textbf{Average Time (minutes) $\pm$ std} & \textbf{Median} & \textbf{Variance} \\
        \hline
        1 (Training) & 2.024 $\pm$ 2.364 & 1.176 & 5.588 \\
        2 (Testing) & 1.111 $\pm$ 2.017 & 0.507 & 4.067 \\
        3 (Training) & 0.682 $\pm$ 0.700 & 0.480 & 0.490 \\
        4 (Testing) & 1.456 $\pm$ 2.389 & 0.821 & 5.707 \\
        5 (Training) & 2.993 $\pm$ 2.960 & 2.317 & 8.761 \\
        6 (Testing) & 1.083 $\pm$ 1.261 & 0.768 & 1.591 \\
        \hline
    \end{tabular}
\end{table}

\begin{table}
    \caption{
\textbf{Participant demographics}. 
    This table provides a comprehensive overview of the demographic characteristics of the participants involved in the study. 
    It includes data on gender, years of experience in their respective domains, student status, and domain of expertise.
Thirteen participants indicated that they are not students, while eleven indicated that they are students. 
In terms of domain expertise, approximately 46\% (11 participants or ps) were from computer science, 29\% (7 ps) from various other fields, 16\% (4 ps) identified as bioinformaticians, 4\% (1 participant) from biochemistry, and another 4\% from structural biology.
Additionally, the analysis indicated that approximately 71\% (17 ps) had less than one year of experience, 8\% (2 ps) had between one and three years, 8\% (2 ps) had three to five years, 8 percent (2 ps) had five to ten years, and 8 percent (2 ps) had more than ten years of experience.
    }
\resizebox{\textwidth}{!}{%
    \footnotesize
    \centering
    \begin{tabular}{l p{1.5cm} p{3cm} p{1.3cm} p{3cm}}
        \toprule
        \# & Gender & Years of Experience & Is Student & Domain \\
        \midrule
        P1  & \MALE   & < 1        & Yes & Computer Sci.  \\
        P2  & \FEMALE & < 1    & Yes  & Bioinformatics \\
        P3  & \FEMALE & > 10    & No  & Structural Biology  \\
        P4  & \FEMALE   & < 1    & Yes  & Bioinformatics  \\
        P5  & \textbf{NA}  & < 1        & No & Computer Sci.  \\
        P6  & \MALE   & < 1        & Yes & Computer Sci.  \\
        P7  & \MALE & 5 -- 10   & No  & Computer Sci.  \\
        P8  & \FEMALE & < 1   & Yes  & Computer Sci.  \\
        P9  & \MALE   & < 1    & No  & Others \\
        P10  & \MALE & < 1    & Yes  & Computer Sci. \\
        P11 & \FEMALE   & 1 -- 3    & No  & Others \\
        P12 & \MALE   & 3 -- 5   & No  & Others \\
        P13 & \MALE & < 1    & No  & Computer Sci.   \\
        P14 & \MALE & < 1    & No  & Bioinformatics  \\
        P15 & \MALE   & > 10    & No  & Biochemistry  \\
        P16 & \MALE   & < 1        & No & Others \\
        P17 & \FEMALE   & < 1    & No  & Bioinformatics \\
        P18 & \FEMALE & < 1        & No & Computer Sci. \\
        P19 & \MALE & 3 -- 5   & No  & Computer Sci. \\
        P20 & \MALE   & < 1        & Yes & Computer Sci. \\
        P21 & \FEMALE & < 1        & No & Others \\
        P22 & \FEMALE & 1 -- 3        & No & Computer Sci. \\
        P23 & \FEMALE & < 1        & Yes & Others \\
        P24 & \MALE & < 1        & No & Others \\
        \bottomrule
    \end{tabular}%
    }
    \label{tab: participants}
\end{table}

\clearpage

\begin{table}
\centering
\caption{\textbf{The system usability scale (SUS) results: Usefulness}.
}
\begin{tabular}{p{6cm}ccccc}
    \toprule
    \textbf{Description} & \textbf{Strongly Agree} & \textbf{Agree} & \textbf{Neutral} & \textbf{Disagree} & \textbf{Strongly Disagree} \\
    \midrule
This tool would be useful for classifying membrane proteins into groups and sub-groups. & 5 & 14 & 3 & 2 & 1 \\
This tool would be useful for the study of membrane proteins. & 5 & 15 & 3 & 2 & 0 \\
Do you agree that this system accurately detects outliers based on the analysis of scatter plot matrix (SPLOM) ?
 & 6 & 11 & 8 & 0 & 0 \\
Do you agree that this system accurately detects outliers based on the analysis of the scatter plot (outlier detection plot) using DBSCAN ?
  & 6 & 12 & 6 & 0 & 1 \\
Do you agree that this system accurately detects outliers based on the analysis of the boxplot ? & 4 & 14 & 6 & 0 & 1 \\
\bottomrule
\end{tabular}
\label{table:Usefulness}
\end{table}

\begin{table}
\centering
\caption{\textbf{The system usability scale (SUS) results: Intuitiveness}.
P7 and P20 highlighted that ``\textit{everything looks well-developed and quite straightforward to follow}'' and that the questionnaire was ``\textit{extremely user-friendly with a visually appealing design.}'' Although the majority of participants found the system to be generally intuitive and well-designed, an issue was noted about the task instructions by only one participant. P4 reported problems related to ``\textit{resetting the page and selecting/deselecting clusters}'', which are functionalities covered in the provided instructions. 
This suggests that not all participants may have carefully read with the available instructions.
This may indicate a need for shorter instructions or clearer instructional cues to guide participants.}
\begin{tabular}{p{6cm}ccccc}
    \toprule
    \textbf{Description} & \textbf{Strongly Agree} & \textbf{Agree} & \textbf{Neutral} & \textbf{Disagree} & \textbf{Strongly Disagree} \\
\midrule
I think I would like to use this system frequently and most people would learn to use it very quickly. & 4 & 9 & 9 & 2 & 1 \\
The system is intuitive, facilitating easy navigation to locate desired information.
    & 6 & 11 & 4 & 3 & 1 \\
I found the system unnecessarily complex and very cumbersome to use. & 1 & 2 & 6 & 10 & 6 \\
I would need the support of a technical person to use this system. & 1 & 5 & 2 & 9 & 8 \\
I thought there was too much inconsistency in this system. & 0 & 3 & 3 & 14 & 5 \\

I felt confident using the system and am very satisfied with the overall user experience. & 3 & 12 & 6 & 3 & 1 \\
The system loads quickly.  & 12 & 13 & 0 & 0 & 0 \\
    
\bottomrule
\end{tabular}
\label{table:intuitiveness}
\end{table}

\begin{table}
\centering
\caption{\textbf{The system usability scale (SUS) results: Design choices}. }
\begin{tabular}{p{6cm}ccccc}
    \toprule
    \textbf{Description} & \textbf{Strongly Agree} & \textbf{Agree} & \textbf{Neutral} & \textbf{Disagree} & \textbf{Strongly Disagree} \\
    \midrule
I found it easy to interact with the charts, identify outliers using the box-plot, and use the interactive elements. & 7 & 11 & 5 & 2 & 0 \\
The layout and organization of graphical elements are intuitive. & 9 & 9 & 6 & 1 & 0 \\
    \bottomrule
\end{tabular}
\label{table:design-choices}
\end{table}

\begin{table}
    \centering
    \caption{\textbf{Integrated attributes from the MPstruc database}.
    The 10 attributes from the Membrane Proteins of Known Structure (MPstruc) database included in the MetaMP database.}
    \begin{tabular}{ll}

\toprule
\textbf{Group} & \textbf{Attribute} \\
\midrule
Protein Classification & \begin{tabular}[t]{@{}l@{}}
group \
subgroup \
is\_master\_protein
\end{tabular} \\
Protein Identification & \begin{tabular}[t]{@{}l@{}}
pdb\_code \
name \
description
\end{tabular} \\
Taxonomic Information & \begin{tabular}[t]{@{}l@{}}
species \
taxonomic\_domain \
expressed\_in\_species
\end{tabular} \\
Structural Information & resolution \\
        
        \bottomrule
    \end{tabular}
\end{table}

\begin{table}
    \centering
    \caption{\textbf{Integrated attributes from the OPM database}.
    The 28 attributes from the Orientations of Proteins in Membranes (OPM) database included in the MetaMP database.}
    \begin{tabular}{ll}
        \toprule

Group & Attribute \\
\midrule
Identification & \begin{tabular}[t]{@{}l@{}}
id, \
pdbid, \
pdb\_code
\end{tabular} \\
Protein Information & \begin{tabular}[t]{@{}l@{}}
name, \
comments, \
family\_name, \
family\_name\_cache,
\end{tabular} \\
Structural Properties & \begin{tabular}[t]{@{}l@{}}
resolution, \
thickness, \
thicknesserror, \\
subunit\_segments, \
tilt, \
tilterror, \
gibbs
\end{tabular} \\
Classification & \begin{tabular}[t]{@{}l@{}}
family\_superfamily\_name \\
family\_superfamily\_tcdb \\
family\_superfamily\_classtype\_name \\
famsupclasstype\_ superfamilies\_count \\
famsupclasstype\_type\_name \\
famsupclasstype\_type\_classtypes\_count
\end{tabular} \\
Species Information & \begin{tabular}[t]{@{}l@{}}
species\_name, \
species\_name\_cache, \
species\_description
\end{tabular} \\
Membrane Properties & \begin{tabular}[t]{@{}l@{}}
membrane\_name, \
membrane\_name\_cache, \
membrane\_short\_name, \\
membrane\_topology\_in, \
membrane\_topology\_out
\end{tabular} \\

        \bottomrule
    \end{tabular}
\end{table}

\begin{table}
    \centering
    \caption{\textbf{Integrated attributes from the UniProt database}. 
    The 28 attributes from the Universal Protein Resource (UniProt) included in the MetaMP database.}
    \begin{tabular}{ll}
        \toprule

Group & Attribute \\
\midrule
Identification & \begin{tabular}[t]{@{}l@{}}
id, \
uniprot\_id, \
pdb\_code, \
secondary\_accession
\end{tabular} \\
Metadata & \begin{tabular}[t]{@{}l@{}}
info\_type, \
info\_created, \\
info\_modified, \
info\_sequence\_update, \
annotation\_score
\end{tabular} \\
Organism Information & \begin{tabular}[t]{@{}l@{}}
organism\_scientific\_name, \
organism\_common\_name, \
organism\_lineage,
\end{tabular} \\
Protein Information & \begin{tabular}[t]{@{}l@{}}
protein\_recommended\_name, \
protein\_alternative\_name, \\
associated\_genes \
gene\_names
\end{tabular} \\
Sequence Information & \begin{tabular}[t]{@{}l@{}}
sequence\_length, \
sequence\_mass, \
sequence\_sequence,
\end{tabular} \\
Functional Information & \begin{tabular}[t]{@{}l@{}}
molecular\_function, \
cellular\_component, \
biological\_process
\end{tabular} \\
Comments & \begin{tabular}[t]{@{}l@{}}
comment\_function, \
comment\_interactions, \
comment\_catalytic\_activity, \\
comment\_subunit, \
comment\_ptm, \\
comment\_caution, \
comment\_subcellular\_locations, \\
comment\_alternative\_products, \
comment\_disease\_name, \\
comment\_disease, \
comment\_similarity
\end{tabular} \\
Additional Information & \begin{tabular}[t]{@{}l@{}}
features, \
references, \
keywords, \
extra\_attributes, \
cross\_references,
\end{tabular} \\ 
   
        \bottomrule
    \end{tabular}
\end{table}

\clearpage 

\begin{longtable}{p{4cm}p{11cm}}
\caption{\textbf{Integrated attributes from the PDB database}. The 220 attributes from the Protein Data Bank (PDB) database included in the MetaMP database.} \\

\toprule
\textbf{Group} & \textbf{Attribute Names} \\
\midrule
\endfirsthead

\multicolumn{2}{c}{\tablename\ \thetable{} -- Continued from previous page} \\
\toprule
\textbf{Group} & \textbf{Attribute Names} \\
\midrule
\endhead

\midrule \multicolumn{2}{r}{{Continued on next page}} \\
\endfoot

\bottomrule
\endlastfoot

Crystal Structure & cell\_angle\_alpha, cell\_angle\_beta, cell\_angle\_gamma, cell\_length\_a, cell\_length\_b, cell\_length\_c, cell\_zpdb, symmetry\_int\_tables\_number, symspagroup\_name\_hm \\

Database Status & pdbx\_database\_status\_pdb\_format\_compatible, pdbx\_database\_status\_recvd\_initial\_deposition\_date, pdbx\_database\_status\_status\_code, pdbx\_database\_status\_process\_site, pdbx\_database\_status\_deposit\_site, pdbx\_database\_status\_status\_code\_sf \\

Accession Information & rcsb\_accession\_info\_deposit\_date, rcsb\_accession\_info\_has\_released\_experimental\_data, rcsb\_accession\_info\_initial\_release\_date, rcsb\_accession\_info\_major\_revision, rcsb\_accession\_info\_minor\_revision, rcsb\_accession\_info\_revision\_date, rcsb\_accession\_info\_status\_code \\

Entry Information & rcsentinfo\_assembly\_count, rcsb\_entry\_info\_* (all remaining attributes) \\

Citation Information & rcsb\_primary\_citation\_country, rcsb\_primary\_citation\_journal\_abbrev, rcsb\_primary\_citation\_journal\_volume, rcsb\_primary\_citation\_page\_first, rcsb\_primary\_citation\_page\_last, rcsb\_primary\_citation\_rcsb\_journal\_abbrev, citation\_country \\

Refinement Statistics & refine\_ls\_rfactor\_rfree, refine\_ls\_rfactor\_rwork, refine\_ls\_rfactor\_obs, refine\_ls\_dres\_high, refine\_ls\_dres\_low, refine\_ls\_number\_reflns\_obs, refine\_pdbx\_ls\_sigma\_f, refine\_ls\_percent\_reflns\_rfree, refine\_ls\_percent\_reflns\_obs, refine\_biso\_mean, refine\_ls\_number\_reflns\_rfree, refine\_ls\_number\_reflns\_all, refine\_correlation\_coeff\_fo\_to\_fc, refine\_correlation\_coeff\_fo\_to\_fc\_free, refine\_overall\_suml \\

Refinement Methods & refine\_pdbx\_rfree\_selection\_details, refine\_pdbx\_method\_to\_determine\_struct, refine\_solvent\_model\_details, refine\_pdbx\_stereochemistry\_target\_values, refine\_pdbx\_starting\_model, refine\_overall\_sub, refine\_pdbx\_overall\_esurfree, refine\_pdbx\_solvent\_ion\_probe\_radii, refine\_pdbx\_solvent\_shrinkage\_radii, refine\_pdbx\_solvent\_vdw\_probe\_radii, refine\_details \\

Anisotropic B-factors & refine\_aniso\_b11, refine\_aniso\_b12, refine\_aniso\_b13, refine\_aniso\_b22, refine\_aniso\_b23, refine\_aniso\_b33 \\

\text{Refinement Details} & \begin{tabular}[t]{@{}l@{}}
\text{refine\_pdbx\_overall\_phase\_error} \\
\text{refine\_biso\_max, refine\_biso\_min} \\
\text{refine\_hist\_d\_res\_high, refine\_hist\_d\_res\_low} \\
\text{refine\_hist\_number\_atoms\_solvent} \\
\text{refine\_hist\_number\_atoms\_total} \\
\text{refine\_hist\_pdbx\_number\_atoms\_ligand} \\
\text{refine\_hist\_pdbx\_number\_atoms\_nucleic\_acid} \\
\text{refine\_hist\_pdbx\_number\_atoms\_protein}
\end{tabular} \\

\text{Diffraction Experiment} & \begin{tabular}[t]{@{}l@{}}
\text{diffrn\_ambient\_temp} \\
\text{diffrn\_pdbx\_serial\_crystal\_experiment} \\
\text{diffrn\_radiation\_pdbx\_scattering\_type} \\
\text{diffrn\_radiation\_pdbx\_monochromatic\_or\_laue\_ml} \\
\text{diffrn\_radiation\_monochromator} \\
\text{diffrn\_radiation\_pdbx\_diffrn\_protocol} \\
\text{diffrn\_detector\_pdbx\_collection\_date} \\
\text{diffrn\_detector\_detector} \\
\text{diffrn\_detector\_type}
\end{tabular} \\

\text{Reflection Data} & \begin{tabular}[t]{@{}l@{}}
\text{reflns\_d\_resolution\_high, reflns\_d\_resolution\_low} \\
\text{reflns\_number\_obs, refobscriterion\_sigma\_i} \\
\text{reflns\_pdbx\_rmerge\_iobs, reflns\_pdbx\_ordinal} \\
\text{reflns\_pdbx\_redundancy, reflns\_percent\_possible\_obs} \\
\text{reflns\_number\_all, reflns\_observed\_criterion\_sigma\_f} \\
\text{reflns\_pdbx\_net\_iover\_sigma\_i} \\
\text{reflns\_biso\_wilson\_estimate, reflns\_pdbx\_cchalf}
\end{tabular} \\

\text{Diffraction Source} & \begin{tabular}[t]{@{}l@{}}
\text{diffrn\_source\_source, diffrn\_source\_type} \\
\text{diffrn\_source\_pdbx\_synchrotron\_beamline} \\
\text{diffrn\_source\_pdbx\_synchrotron\_site} \\
\text{diffrn\_source\_pdbx\_wavelength\_list}
\end{tabular} \\

\text{Crystal Growth} & \begin{tabular}[t]{@{}l@{}}
\text{exptl\_crystal\_grow\_method} \\
\text{expcrygrow\_p\_h, expcrygrow\_pdbx\_details} \\
\text{exptl\_crystal\_grow\_temp} \\
\text{exptl\_crystal\_grow\_method1} \\
\text{exptl\_crystal\_grow\_method2}
\end{tabular} \\

\text{Reflection Shells} & \begin{tabular}[t]{@{}l@{}}
\text{refshed\_res\_high, refshed\_res\_low} \\
\text{reflns\_shell\_pdbx\_ordinal} \\
\text{reflns\_shell\_rmerge\_iobs} \\
\text{refshepercent\_possible\_all} \\
\text{reflns\_shell\_pdbx\_redundancy} \\
\text{refshemean\_iover\_sig\_iobs} \\
\text{refshenumber\_unique\_obs} \\
\text{reflns\_shell\_pdbx\_cchalf}
\end{tabular} \\

\text{Related Database Information} & \begin{tabular}[t]{@{}l@{}}
\text{pdbx\_database\_related\_content\_type} \\
\text{pdbx\_database\_related\_db\_name}
\end{tabular} \\

\text{Initial Refinement Model} & \begin{tabular}[t]{@{}l@{}}
\text{pdbx\_initial\_refinement\_model\_accession\_code} \\
\text{pdbx\_initial\_refinement\_model\_source\_name} \\
\text{pdbx\_initial\_refinement\_model\_type} \\
\text{pdbx\_initial\_refinement\_model\_details}
\end{tabular} \\

\text{Audit Support} & \begin{tabular}[t]{@{}l@{}}
\text{pdbx\_audit\_support\_country} \\
\text{pdbx\_audit\_support\_funding\_organization} \\
\text{pdbx\_audit\_support\_grant\_number} \\
\text{pdbx\_audit\_support\_ordinal}
\end{tabular} \\

\text{Electron Microscopy Fitting} & \begin{tabular}[t]{@{}l@{}}
\text{em3d\_fitting\_ref\_protocol} \\
\text{em3d\_fitting\_ref\_space}
\end{tabular} \\

\text{EM 3D Reconstruction} & \begin{tabular}[t]{@{}l@{}}
\text{em3d\_reconstruction\_num\_particles} \\
\text{em3d\_reconstruction\_resolution} \\
\text{em3d\_reconstruction\_resolution\_method} \\
\text{em3d\_reconstruction\_symmetry\_type}
\end{tabular} \\

\text{EM Imaging} & \begin{tabular}[t]{@{}l@{}}
\text{em\_ctf\_correction\_type} \\
\text{em\_imaging\_recording\_average\_exposure\_time} \\
\text{em\_imaging\_recording\_avg\_electron\_dose\_per\_image} \\
\text{em\_imaging\_recording\_film\_or\_detector\_model} \\
\text{em\_imaging\_recording\_detector\_mode} \\
\text{em\_imaging\_recording\_num\_real\_images} \\
\text{em\_imaging\_accelerating\_voltage} \\
\text{em\_imaging\_electron\_source} \\
\text{em\_imaging\_illumination\_mode} \\
\text{em\_imaging\_microscope\_model} \\
\text{em\_imaging\_mode} \\
\text{em\_imaging\_nominal\_defocus\_max} \\
\text{em\_imaging\_nominal\_defocus\_min} \\
\text{em\_imaging\_cryogen} \\
\text{em\_imaging\_nominal\_cs} \\
\text{em\_imaging\_specimen\_holder\_model} \\
\text{em\_imaging\_nominal\_magnification}
\end{tabular} \\

\text{EM Particle Selection} & \begin{tabular}[t]{@{}l@{}}
\text{em\_particle\_selection\_num\_particles\_selected}
\end{tabular} \\

\text{EM Specimen Preparation} & \begin{tabular}[t]{@{}l@{}}
\text{em\_specimen\_embedding\_applied} \\
\text{em\_specimen\_shadowing\_applied} \\
\text{em\_specimen\_staining\_applied} \\
\text{em\_specimen\_vitrification\_applied} \\
\text{em\_specimen\_concentration}
\end{tabular} \\

\text{EM Vitrification} & \begin{tabular}[t]{@{}l@{}}
\text{em\_vitrification\_cryogen\_name} \\
\text{em\_vitrification\_chamber\_temperature} \\
\text{em\_vitrification\_humidity} \\
\text{em\_vitrification\_instrument}
\end{tabular} \\

\text{EM Particle Symmetry} & \begin{tabular}[t]{@{}l@{}}
\text{emsinparticle\_entity\_point\_symmetry}
\end{tabular} \\

\text{Bibliography} & \begin{tabular}[t]{@{}l@{}}
\text{bibliography\_year}
\end{tabular} \\

\text{Resolution} & \begin{tabular}[t]{@{}l@{}}
\text{Resolution}
\end{tabular} \\

\text{Miscellaneous} & \begin{tabular}[t]{@{}l@{}}
\text{struct\_pdbx\_descriptor}
\end{tabular} \\

\end{longtable}

\begin{figure}
\centering
\includegraphics[width=\textwidth]{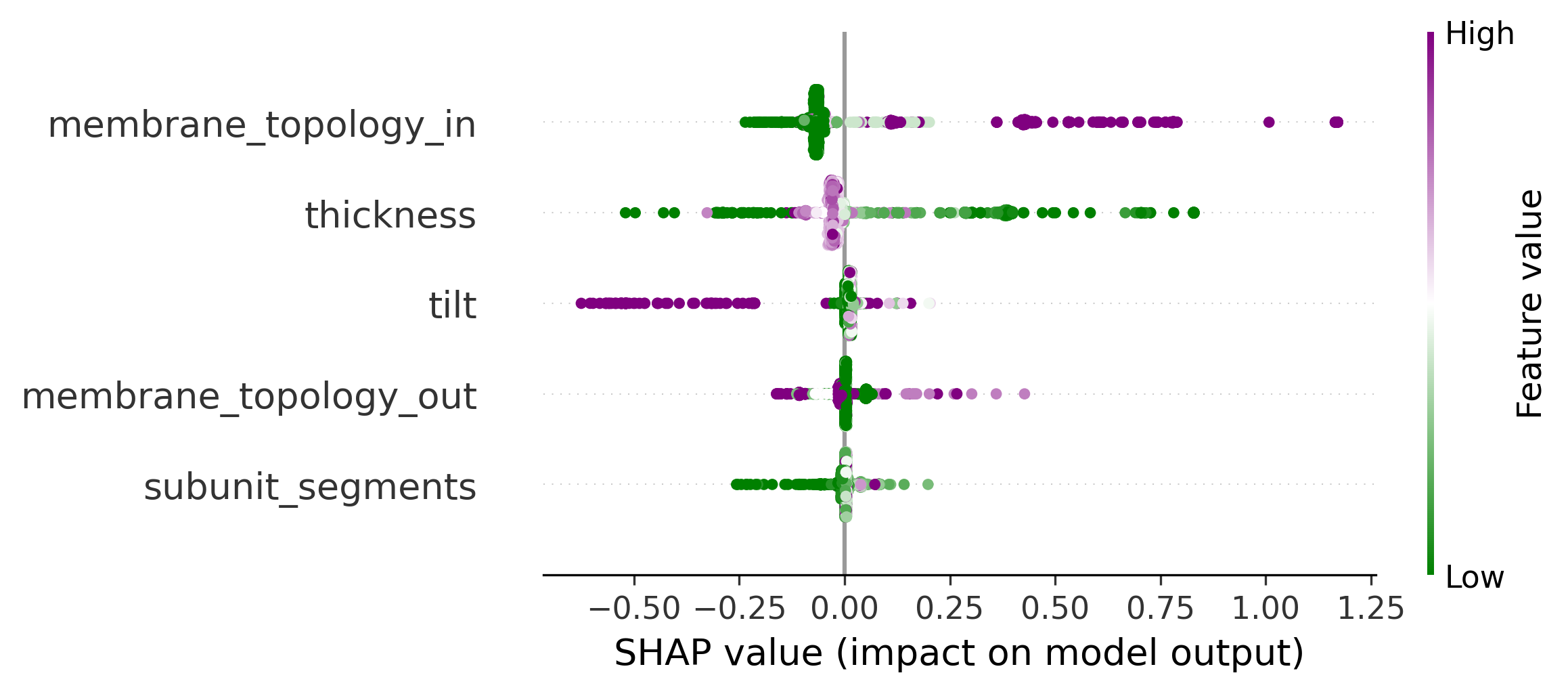}
\caption{\textbf{SHAP summary chart}. This chart visualizes the contribution of each feature to the prediction model, illustrating the feature importance and their respective impact on the target variable. Each dot represents a single observation in the dataset, where the position along the x-axis shows the SHAP value (effect on the prediction), and the color gradient indicates the feature value (from low to high). Features with higher SHAP values have a more substantial influence on the prediction. This plot not only ranks the features by importance but also provides insights into how different values of each feature drive model predictions. (green = low, purple = high)}
\label{figure:shap}
\end{figure}

\clearpage

\begin{figure}[ht]
  \centering
  \subcaptionbox{Pearson $r$\label{fig:pearson}}
    {\includegraphics[width=0.48\linewidth]{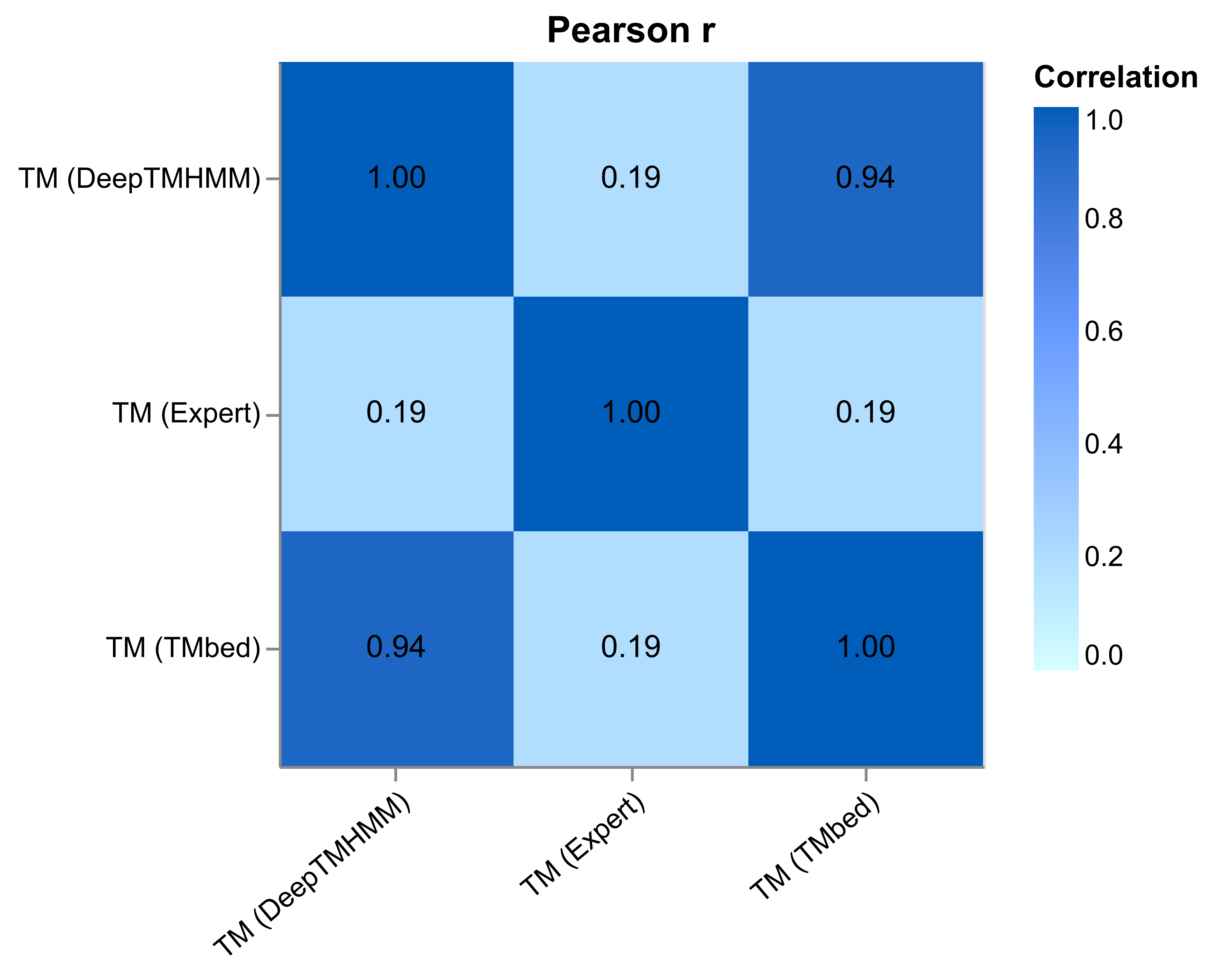}}
  \hfill
  \subcaptionbox{Spearman $\rho$\label{fig:spearman}}
    {\includegraphics[width=0.48\linewidth]{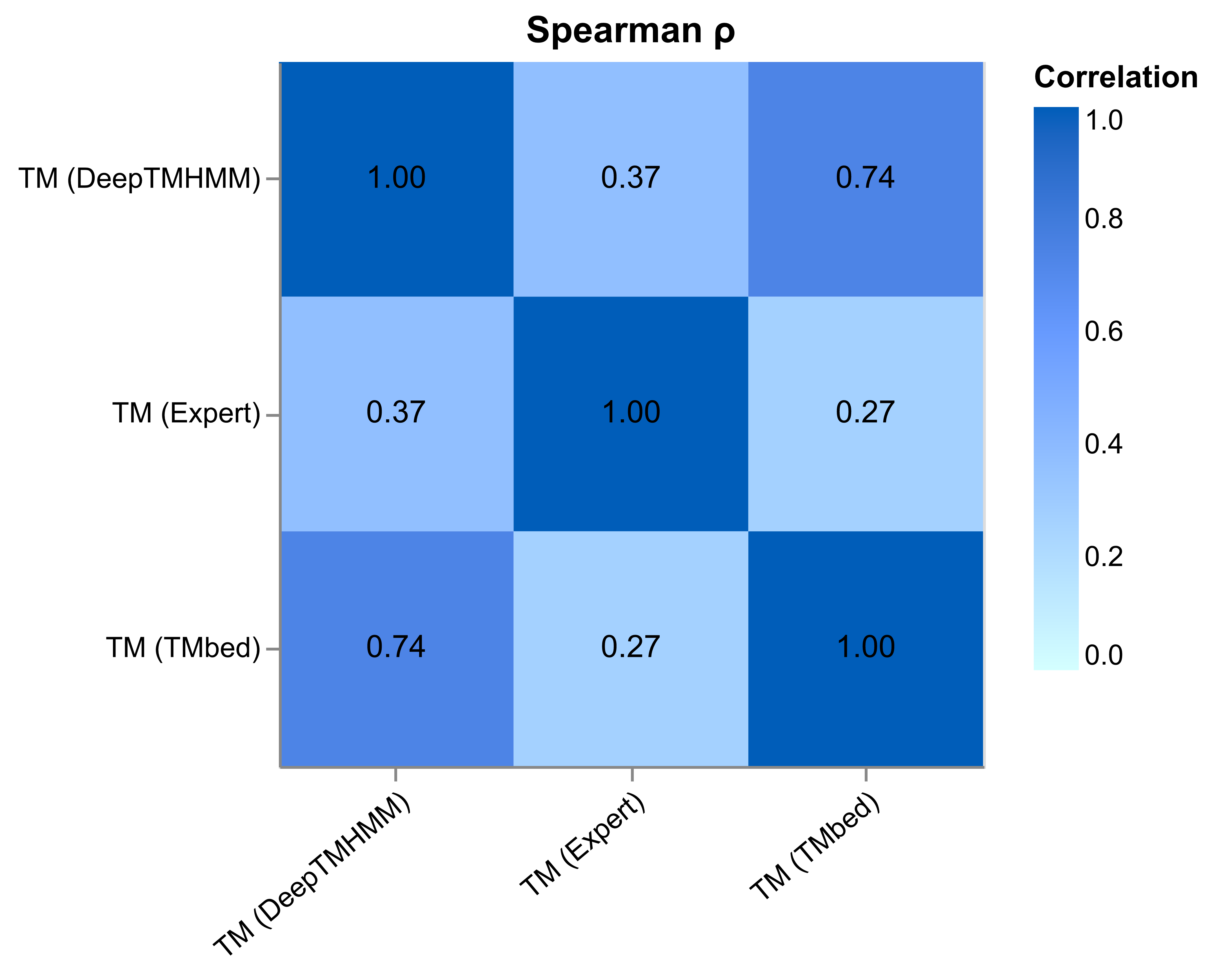}}
  \caption{
    \textbf{Pairwise correlations between expert TM-segment counts and automatic predictors.}
    (A) Pearson’s product--moment coefficients ($r$) and (B) Spearman’s rank-order coefficients ($\rho$) calculated for all 121 proteins in the benchmark set.
    Axes list the reference annotation \textbf{TM (Expert)} and the two predictions
    \textbf{TMbed (Predicted)} and \textbf{DeepTMHMM (Predicted)}.
    Each cell is shaded from light blue (no correlation) to deep blue (\#005DB9; $|$coefficient$|=1$)
    and labelled with the coefficient to two decimal places.
    TMbed and DeepTMHMM correlate strongly with one another ($r = 0.94$, $\rho = 0.74$)
    but only weakly with the expert counts (maximum $r \approx 0.19$, $\rho \approx 0.37$),
    underscoring their shared prediction bias relative to the curated reference.
  }
  \label{fig:correlation_heatmaps}
\end{figure}

\begin{figure}[ht]
    \centering
    \includegraphics[width=\linewidth]{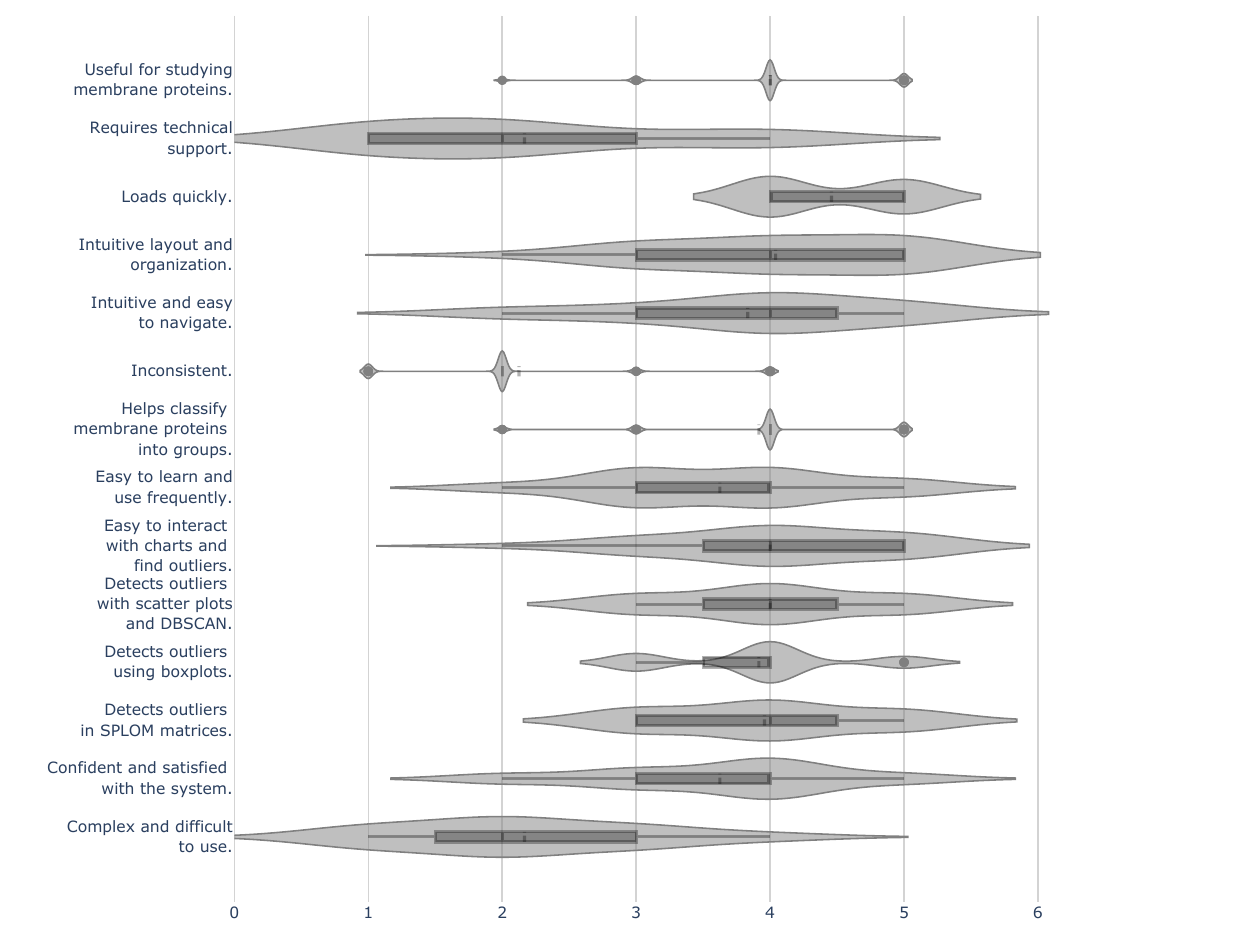}
    \caption{
    \textbf{System Usability Scale (SUS) scores}. 
    The majority of participants evaluated the system as highly usable, well integrated, and loads quickly.
    P7 noted, ``\textit{All in all, it's a great system, fast and reliable.}'' 
The integration of different visualization techniques was recognized as valuable and contributed to an overall positive user experience.
In addition, during Task 3, some participants found the process of comparing outlier detection between SPLOM and the scatterplot DBSCAN (Density-Based Spatial Clustering of Applications with Noise) complex. 
P4 commented, ``\textit{Comparing outlier detection between SPLOM and DBSCAN by manually hovering over all data points seems unnecessarily complicated.}'' 
To remedy this, we have changed the layout of the graphs for intuitive visual linking of the relevant visualizations, reduced the need for scrolling, and improved the interactive idiom of brushing and linking.
Another criticism was that it was difficult to interpret certain graphs, especially those related to count data. 
P7 noted, ``\textit{The different aspect of the experiment was not really showing on the graph for easy interpretation on only 1 part. It was the graph relating to the count. 
You can see the count but unable to see which experiment is in blue or red when hovered, making it not too clear for the user.}'' 
In this version, we have updated the most evident graphical marks that were superimposed on one another by changing the visual channel encodings. 
Furthermore, we are currently adding annotations with the objective of enhancing clarity. 
The feedback on the interactive features revealed both positive aspects and potential areas for further enhancement.
For instance, one participant (P9) requested to improve a Drop List Field-related interaction and make the Apply Change button ``\textit{more visually prominent}''. 
We made this adjustment according to the feedback.
Another instance was a request from a couple of participants to improve the real-time brushing and linking during Task 3. Outlier Detection: ``\textit{When selecting data points in the SPLOM, the corresponding points should be dynamically displayed in the DBSCAN clustering scatterplot.}''. 
We have addressed this feature request by making interactivity bi-directional across charts. 
This update improved the intuitiveness with which users can find and analyze outliers.
    }
    \label{fig:SUS_violin_results}
\end{figure}

\begin{figure}[ht]
\centering
\includegraphics[width=\textwidth]{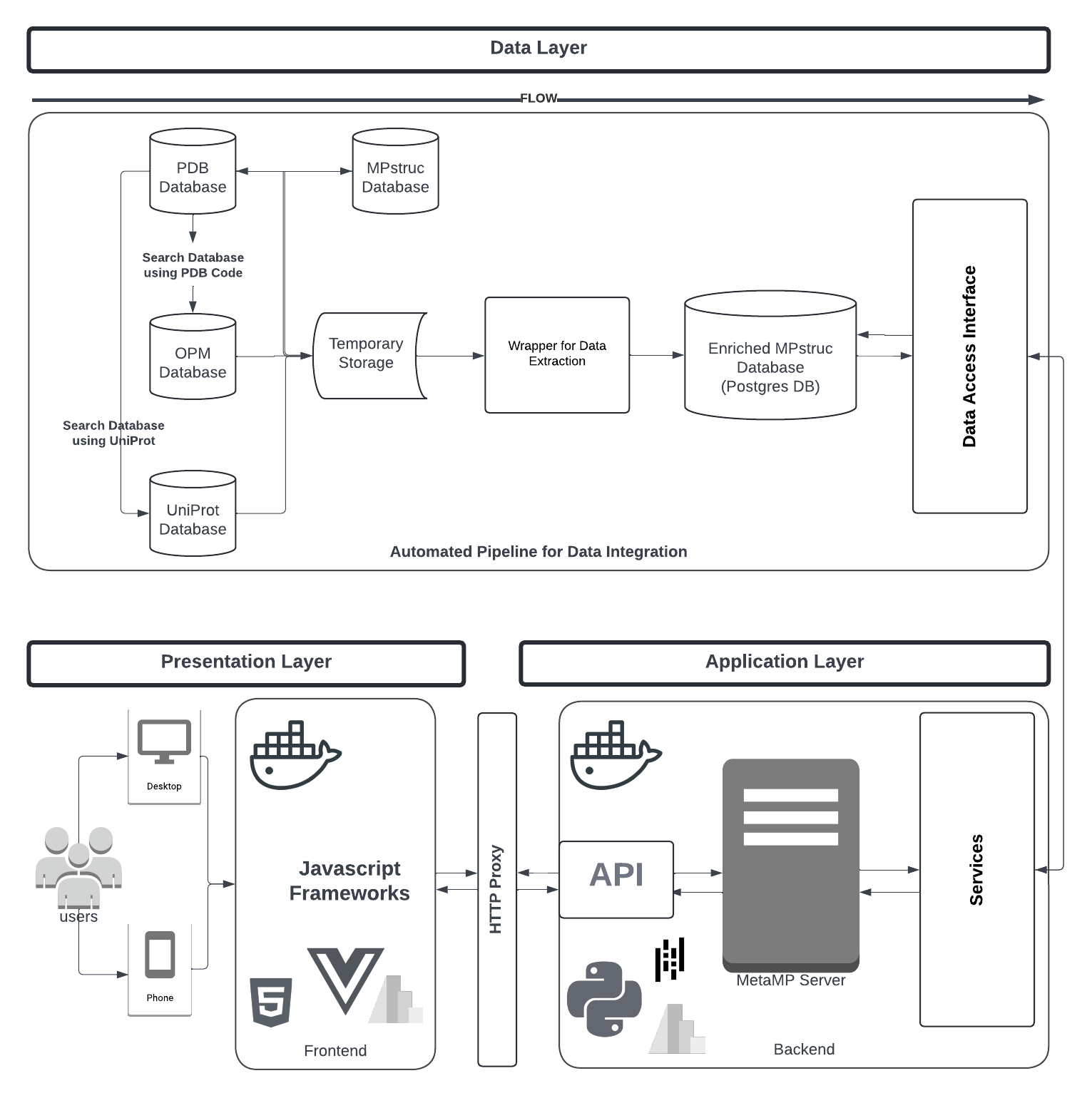}
\caption{
    \textbf{The three-tiered architecture of MetaMP}. 
    It comprises the Data, Application, and Presentation layers. 
    The Data Layer integrates various experimental and computational sources. 
    The Application Layer handles data processing and analysis.
    The Presentation Layer provides visualization and user interaction. 
    This structure ensures robust, scalable, and comprehensive meta-analysis capabilities.
}
\label{fig:architectural_design_for_MetaMP}
\end{figure}

\begin{longtable}{p{1cm}p{1cm}p{2cm}p{2cm}p{2cm}p{2cm}p{1cm}p{1cm}p{1cm}}
\caption{\textbf{Data discrepancy resolution}. 
Comparison between OPM database, MPstruc database, model predictions, and expert feedback.
In the columns Group (OPM), Group (MPstruc) and Group (Predicted), three groups inherited from the MPstruc database are considered: Monotopic (1), transmembrane alpha (2), and transmembrane beta (3).
The Group (Expert) proposes a fourth group Bitopic (4) to distinguish mono-, bi-, from poly-topic membrane proteins.
A bitopic protein or single-pass membrane protein, also known as single-spanning protein, is a transmembrane protein that spans the lipid bilayer only once; in contrast to poly-topic membrane proteins.
This distinction adds more granularity to the classes; hence, transmembrane alpha and beta are characterized as polytopic alpha-helical and polytopic beta-sheet proteins, respectively.
Trans membrane (TM) segments are reported by the expert to support decision-making.
(x) indicates the presence of subunits in the membrane protein. For example, (1x2) indicates a dimer with 1 TM per chain.
(*) indicates no or not all TM  present in the structure 
(**) indicates beta-barrel pore forming in its soluble form (no TM). This is specific to CDC pore forming family.
The CDC pore-forming family: 1PFO, 3NSJ, 4HSC, 4CDB, 5IMW, 5IMY. 
} \\
\toprule
\textbf{Year} & \textbf{PDB Code} & \textbf{Group (OPM)} & \textbf{Group (MPstruc)} & \textbf{Group (Predicted)} & \textbf{Group (Expert)} & \textbf{TM (Expert)} & \textbf{TM (TMbed)} & \textbf{TM (DeepTMHMM)} \\
\midrule
\endfirsthead

\multicolumn{7}{c}{\tablename\ \thetable{} -- Continued from previous page} \\
\toprule
\textbf{Year} & \textbf{PDB Code} & \textbf{Group (OPM)} & \textbf{Group (MPstruc)} & \textbf{Group (Predicted)} & \textbf{Group (Expert)} & \textbf{TM (Expert)} & \textbf{TM (TMbed)} & \textbf{TM (DeepTMHMM)} \\
\midrule
\endhead

\midrule \multicolumn{7}{r}{{Continued on next page}} \\
\endfoot

\bottomrule
\endlastfoot

1997 & 1PFO & Monotopic membrane proteins & Transmembrane proteins:beta-barrel       & Transmembrane proteins:beta-barrel       & Transmembrane proteins:beta-barrel       & 0** & 2  & 0  \\
1997 & 1FDM & Bitopic proteins                & Transmembrane proteins:alpha-helical     & Transmembrane proteins:alpha-helical     & Bitopic                                 & 1   & 1  & 1  \\
1997 & 1AFO & Bitopic proteins                & Transmembrane proteins:alpha-helical     & Transmembrane proteins:alpha-helical     & Bitopic                                 & 1   & 1  & 1  \\
1998 & 2CPB & Bitopic proteins                & Transmembrane proteins:alpha-helical     & Transmembrane proteins:alpha-helical     & Bitopic                                 & 1   & 1  & 1  \\
1998 & 1B12 & Transmembrane proteins:alpha-helical & Monotopic membrane proteins         & Monotopic membrane proteins         & Monotopic membrane proteins         &     & 2  & 2  \\
2002 & 1GOS & Bitopic proteins                & Monotopic membrane proteins             & Monotopic membrane proteins             & Bitopic                                 & 1   & 0  & 0  \\
2002 & 1MT5 & Bitopic proteins                & Monotopic membrane proteins             & Monotopic membrane proteins             & Monotopic membrane proteins         &     & 0  & 0  \\
2002 & 1KN9 & Transmembrane proteins:alpha-helical & Monotopic membrane proteins         & Monotopic membrane proteins         & Monotopic membrane proteins         &     & 2  & 2  \\
2003 & 1OJA & Bitopic proteins                & Monotopic membrane proteins             & Monotopic membrane proteins             & Bitopic                                 & 1   & 0  & 0  \\
2003 & 1MZT & Bitopic proteins                & Transmembrane proteins:alpha-helical    & Transmembrane proteins:beta-barrel      & Bitopic                                 & 1   & 1  & 1  \\
2004 & 1O5W & Bitopic proteins                & Monotopic membrane proteins             & Transmembrane proteins:alpha-helical    & Bitopic                                 & 1   & 0  & 0  \\
2004 & 1PJF & Bitopic proteins                & Transmembrane proteins:alpha-helical    & Transmembrane proteins:alpha-helical    & Bitopic                                 & 1   & 1  & 1  \\
2004 & 1UUM & Bitopic proteins                & Monotopic membrane proteins             & Monotopic membrane proteins             & Bitopic                                 & 1*  & 0  & 0  \\
2004 & 1T7D & Transmembrane proteins:alpha-helical & Monotopic membrane proteins         & Monotopic membrane proteins         & Monotopic membrane proteins         &     & 2  & 2  \\
2005 & 2BXR & Bitopic proteins                & Monotopic membrane proteins             & Transmembrane proteins:alpha-helical    & Bitopic                                 & 1   & 0  & 0  \\
2005 & 1YGM & Monotopic membrane proteins     & Transmembrane proteins:alpha-helical    & Transmembrane proteins:alpha-helical    & Not a Membrane Protein             &     & 0  & 0  \\
2005 & 1ZLL & Bitopic proteins                & Transmembrane proteins:alpha-helical    & Transmembrane proteins:alpha-helical    & Bitopic                                 & 1   & 1  & 1  \\
2006 & 2GMH & Monotopic membrane proteins     & Transmembrane proteins:alpha-helical    & Transmembrane proteins:alpha-helical    & Monotopic membrane proteins         &     & 0  & 0  \\
2006 & 2J58 & Bitopic proteins                & Transmembrane proteins:alpha-helical    & Transmembrane proteins:beta-barrel      & Bitopic                                 & 1x8 & 1  & 0  \\
2006 & 2HAC & Bitopic proteins                & Transmembrane proteins:alpha-helical    & Transmembrane proteins:alpha-helical    & Bitopic                                 & 1   & 1  & 1  \\
2006 & 2J7A & Bitopic proteins                & Transmembrane proteins:alpha-helical    & Transmembrane proteins:alpha-helical    & Bitopic                                 & 1x2 & 0  & 0  \\
2007 & 2OLV & Bitopic proteins                & Monotopic membrane proteins             & Monotopic membrane proteins             & Bitopic                                 & 1*  & 1  & 1  \\
2007 & 2JO1 & Bitopic proteins                & Transmembrane proteins:alpha-helical    & Transmembrane proteins:alpha-helical    & Bitopic                                 & 1   & 1  & 1  \\
2007 & 2OQO & Bitopic proteins                & Monotopic membrane proteins             & Monotopic membrane proteins             & Bitopic                                 & 1*  & 1  & 1  \\
2008 & 2QCU & Monotopic membrane proteins     & Transmembrane proteins:alpha-helical    & Monotopic membrane proteins             & Monotopic membrane proteins         &     & 0  & 0  \\
2008 & 2PRM & Bitopic proteins                & Monotopic membrane proteins             & Monotopic membrane proteins             & Bitopic                                 & 1*  & 0  & 0  \\
2008 & 2Z5X & Bitopic proteins                & Monotopic membrane proteins             & Transmembrane proteins:alpha-helical    & Bitopic                                 & 1   & 0  & 0  \\
2008 & 2VQG & Monotopic membrane proteins     & Transmembrane proteins:alpha-helical    & Transmembrane proteins:beta-barrel      & Transmembrane proteins:alpha-helical & 3   & 0  & 0  \\
2008 & 2JWA & Bitopic proteins                & Transmembrane proteins:alpha-helical    & Transmembrane proteins:alpha-helical    & Bitopic                                 & 1x2 & 1  & 1  \\
2008 & 2K1L & Bitopic proteins                & Transmembrane proteins:alpha-helical    & Transmembrane proteins:alpha-helical    & Bitopic                                 & 1x2 & 1  & 1  \\
2008 & 3BKD & Bitopic proteins                & Transmembrane proteins:alpha-helical    & Transmembrane proteins:alpha-helical    & Bitopic                                 & 1x4 & 1  & 1  \\
2008 & 2RLF & Bitopic proteins                & Transmembrane proteins:alpha-helical    & Transmembrane proteins:alpha-helical    & Bitopic                                 & 1x4 & 1  & 1  \\
2009 & 3HYW & Monotopic membrane proteins     & Transmembrane proteins:alpha-helical    & Monotopic membrane proteins             & Monotopic membrane proteins         &     & 0  & 0  \\
2009 & 3I65 & Bitopic proteins                & Monotopic membrane proteins             & Monotopic membrane proteins             & Bitopic                                 & 1*  & 0  & 0  \\
2009 & 2KNC & Bitopic proteins                & Transmembrane proteins:alpha-helical    & Transmembrane proteins:alpha-helical    & Bitopic                                 & 1   & 1  & 1  \\
2009 & 3HD7 & Bitopic proteins                & Transmembrane proteins:alpha-helical    & Transmembrane proteins:alpha-helical    & Bitopic                                 & 1x2 & 1  & 1  \\
2009 & 2KOG & Bitopic proteins                & Transmembrane proteins:alpha-helical    & Transmembrane proteins:alpha-helical    & Bitopic                                 & 1x1 & 1  & 1  \\
2009 & 2KIH & Bitopic proteins                & Transmembrane proteins:alpha-helical    & Transmembrane proteins:alpha-helical    & Bitopic                                 & 1x4 & 1  & 1  \\
2009 & 2KIX & Bitopic proteins                & Transmembrane proteins:alpha-helical    & Transmembrane proteins:alpha-helical    & Bitopic                                 & 1x4 & 0  & 0  \\
2009 & 3VMA & Bitopic proteins                & Monotopic membrane proteins             & Transmembrane proteins:alpha-helical    & Bitopic                                 & 1   & 1  & 1  \\
2009 & 3JQO & Transmembrane proteins:alpha-helical & Transmembrane proteins:beta-barrel & Transmembrane proteins:beta-barrel & Bitopic & 1x14& 2  & 1  \\
2010 & 3NSJ & Monotopic membrane proteins     & Transmembrane proteins:beta-barrel      & Monotopic membrane proteins             & Transmembrane proteins:beta-barrel & 0**& 0  & 0  \\
2010 & 2KSJ & Bitopic proteins                & Transmembrane proteins:alpha-helical    & Transmembrane proteins:alpha-helical    & Bitopic                                 & 1   & 1  & 1  \\
2010 & 2KS1 & Bitopic proteins                & Transmembrane proteins:alpha-helical    & Transmembrane proteins:alpha-helical    & Bitopic                                 & 1   & 1  & 1  \\
2010 & 2K9Y & Bitopic proteins                & Transmembrane proteins:alpha-helical    & Transmembrane proteins:alpha-helical    & Bitopic                                 & 1   & 1  & 1  \\
2010 & 2L35 & Bitopic proteins                & Transmembrane proteins:alpha-helical    & Transmembrane proteins:alpha-helical    & Bitopic                                 & 1x3 & 1  & 1  \\
2010 & 2L0J & Bitopic proteins                & Transmembrane proteins:alpha-helical    & Transmembrane proteins:alpha-helical    & Bitopic                                 & 1x4 & 1  & 1  \\
2010 & 3LBW & Bitopic proteins                & Transmembrane proteins:alpha-helical    & Transmembrane proteins:alpha-helical    & Bitopic                                 & 1   & 1  & 1  \\
2011 & 3ML3 & Monotopic membrane proteins     & Transmembrane proteins:beta-barrel      & Transmembrane proteins:beta-barrel      & Transmembrane proteins:beta-barrel & 15*& 0  & 0  \\
2011 & 3PRW & Monotopic membrane proteins     & Transmembrane proteins:beta-barrel      & Transmembrane proteins:beta-barrel      & Monotopic membrane proteins         &     & 0  & 0  \\
2011 & 3P1L & Monotopic membrane proteins     & Transmembrane proteins:beta-barrel      & Transmembrane proteins:beta-barrel      & Monotopic membrane proteins         &     & 0  & 0  \\
2011 & 3Q7M & Monotopic membrane proteins     & Transmembrane proteins:beta-barrel      & Transmembrane proteins:beta-barrel      & Monotopic membrane proteins         &     & 0  & 0  \\
2011 & 2YH3 & Monotopic membrane proteins     & Transmembrane proteins:beta-barrel      & Transmembrane proteins:beta-barrel      & Monotopic membrane proteins         &     & 0  & 0  \\
2011 & 3LIM & Monotopic membrane proteins     & Transmembrane proteins:alpha-helical    & Transmembrane proteins:beta-barrel      & Bitopic                                 & 1   & 0  & 0  \\
2011 & 2KPF & Bitopic proteins                & Transmembrane proteins:alpha-helical    & Transmembrane proteins:alpha-helical    & Bitopic                                 & 1x2 & 1  & 1  \\
2011 & 2L9U & Bitopic proteins                & Transmembrane proteins:alpha-helical    & Transmembrane proteins:alpha-helical    & Bitopic                                 & 1x2 & 1  & 1  \\
2011 & 2LJB & Bitopic proteins                & Transmembrane proteins:alpha-helical    & Transmembrane proteins:alpha-helical    & Bitopic                                 & 1x4 & 1  & 1  \\
2011 & 2KYV & Bitopic proteins                & Transmembrane proteins:alpha-helical    & Transmembrane proteins:alpha-helical    & Bitopic                                 & 1x5 & 1  & 1  \\
2012 & 3VMT & Bitopic proteins                & Monotopic membrane proteins             & Transmembrane proteins:alpha-helical    & Bitopic                                 & 1   & 1  & 1  \\
2012 & 3Q54 & Monotopic membrane proteins     & Transmembrane proteins:beta-barrel      & Transmembrane proteins:beta-barrel      & Monotopic membrane proteins         &     &    &     \\
2012 & 2LCX & Bitopic proteins                & Transmembrane proteins:alpha-helical    & Transmembrane proteins:alpha-helical    & Bitopic                                 & 1x2 & 1  & 1  \\
2013 & 2LZL & Bitopic proteins                & Transmembrane proteins:alpha-helical    & Transmembrane proteins:alpha-helical    & Bitopic                                 & 1x2 & 1  & 1  \\
2013 & 2M8R & Bitopic proteins                & Transmembrane proteins:alpha-helical    & Transmembrane proteins:alpha-helical    & Bitopic                                 & 1   & 1  & 1  \\
2013 & 2LY0 & Bitopic proteins                & Transmembrane proteins:alpha-helical    & Transmembrane proteins:alpha-helical    & Bitopic                                 & 1x4 & 1  & 1  \\
2013 & 2M3B & Bitopic proteins                & Transmembrane proteins:alpha-helical    & Transmembrane proteins:alpha-helical    & Bitopic                                 & 1x5 & 1  & 1  \\
2014 & 4LXJ & Bitopic proteins                & Monotopic membrane proteins             & Transmembrane proteins:alpha-helical    & Bitopic                                 & 1   & 1  & 0  \\
2014 & 4HSC & Monotopic membrane proteins     & Transmembrane proteins:beta-barrel      & Transmembrane proteins:beta-barrel      & Transmembrane proteins:beta-barrel & 0**& 1  & 0  \\
2014 & 4CDB & Monotopic membrane proteins     & Transmembrane proteins:alpha-helical    & Transmembrane proteins:beta-barrel      & Transmembrane proteins:beta-barrel & 0**& 1  & 0  \\
2014 & 2MFR & Bitopic proteins                & Transmembrane proteins:alpha-helical    & Transmembrane proteins:alpha-helical    & Bitopic                                 & 1   & 1  & 1  \\
2014 & 2M59 & Bitopic proteins                & Transmembrane proteins:alpha-helical    & Transmembrane proteins:alpha-helical    & Bitopic                                 & 1x2 & 1  & 1  \\
2015 & 4TSY & Monotopic membrane proteins     & Transmembrane proteins:alpha-helical    & Transmembrane proteins:alpha-helical    & Bitopic                                 & 1   & 0  & 0  \\
2015 & 5EH4 & Bitopic proteins                & Transmembrane proteins:alpha-helical    & Transmembrane proteins:alpha-helical    & Bitopic                                 & 1x2 & 1  & 1  \\
2015 & 4WOL & Bitopic proteins                & Transmembrane proteins:alpha-helical    & Transmembrane proteins:alpha-helical    & Bitopic                                 & 1x3 & 1  & 1  \\
2015 & 4QKC & Bitopic proteins                & Transmembrane proteins:alpha-helical    & Transmembrane proteins:alpha-helical    & Bitopic                                 & 1x4 & 1  & 1  \\
2016 & 5B49 & Bitopic proteins                & Monotopic membrane proteins             & Monotopic membrane proteins             & Monotopic membrane proteins         &     & 0  & 0  \\
2016 & 5IMW & Monotopic membrane proteins     & Transmembrane proteins:beta-barrel      & Transmembrane proteins:beta-barrel      & Transmembrane proteins:beta-barrel & 0**& 2  & 0  \\
2016 & 5IMY & Monotopic membrane proteins     & Transmembrane proteins:beta-barrel      & Transmembrane proteins:beta-barrel      & Transmembrane proteins:beta-barrel & 0**& 2  & 0  \\
2016 & 2MIC & Bitopic proteins                & Transmembrane proteins:alpha-helical    & Transmembrane proteins:alpha-helical    & Bitopic                                 & 1x2 & 1  & 1  \\
2016 & 2N2A & Bitopic proteins                & Transmembrane proteins:alpha-helical    & Transmembrane proteins:alpha-helical    & Bitopic                                 & 1x2 & 1  & 1  \\
2016 & 5HK1 & Bitopic proteins                & Transmembrane proteins:alpha-helical    & Transmembrane proteins:alpha-helical    & Bitopic                                 & 1x3 & 1  & 1  \\
2017 & 5LY6 & Monotopic membrane proteins     & Transmembrane proteins:beta-barrel      & Transmembrane proteins:beta-barrel      & Transmembrane proteins:beta-barrel & 4   & 4  & 0  \\
2017 & 5NUO & Transmembrane proteins:beta-barrel & Transmembrane proteins:alpha-helical & Transmembrane proteins:beta-barrel & Transmembrane proteins:beta-barrel & 16x3&16 &16 \\
2017 & 5LV6 & Bitopic proteins                & Transmembrane proteins:alpha-helical    & Transmembrane proteins:alpha-helical    & Bitopic                                 & 1x2 & 1  & 1  \\
2017 & 5JOO & Bitopic proteins                & Transmembrane proteins:alpha-helical    & Transmembrane proteins:alpha-helical    & Bitopic                                 & 1x4 & 1  & 1  \\
2018 & 6BFG & Bitopic proteins                & Monotopic membrane proteins             & Monotopic membrane proteins             & Monotopic membrane proteins         &     & 0  & 0  \\
2018 & 6MLU & Transmembrane proteins:alpha-helical & Monotopic membrane proteins         & Monotopic membrane proteins         & Transmembrane proteins:alpha-helical & 2   & 2  & 2  \\
2018 & 6DLW & Monotopic membrane proteins     & Transmembrane proteins:beta-barrel      & Transmembrane proteins:beta-barrel      & Transmembrane proteins:beta-barrel & 88  & 0  & 0  \\
2018 & 6H03 & Monotopic membrane proteins     & Transmembrane proteins:beta-barrel      & Transmembrane proteins:beta-barrel      & Transmembrane proteins:beta-barrel & 72  & 0  & 0  \\
2018 & 6F2D & Transmembrane proteins:alpha-helical & Transmembrane proteins:beta-barrel & Transmembrane proteins:alpha-helical & Transmembrane proteins:alpha-helical & 34  & 4  & 4  \\
2018 & 6HJR & Bitopic proteins                & Transmembrane proteins:alpha-helical    & Transmembrane proteins:alpha-helical    & Bitopic                                 & 1x3 & 1  & 1  \\
2018 & 6E10 & Bitopic proteins                & Transmembrane proteins:alpha-helical    & Transmembrane proteins:alpha-helical    & Bitopic                                 & 1x7 & 0  & 0  \\
2018 & 6BKK & Bitopic proteins                & Transmembrane proteins:alpha-helical    & Transmembrane proteins:alpha-helical    & Bitopic                                 & 1x4 & 1  & 1  \\
2019 & 6NYF & Monotopic membrane proteins     & Transmembrane proteins:alpha-helical    & Transmembrane proteins:alpha-helical    & Transmembrane proteins:beta-barrel & 12* & 0  & 0  \\
2019 & 6MQU & Bitopic proteins                & Transmembrane proteins:alpha-helical    & Transmembrane proteins:alpha-helical    & Bitopic                                 & 1x5 &    &    \\
2019 & 6JXR & Bitopic proteins                & Transmembrane proteins:alpha-helical    & Transmembrane proteins:alpha-helical    & Bitopic                                 & 1x8 & 1  & 1  \\
2019 & 6MTI & Monotopic membrane proteins     & Transmembrane proteins:alpha-helical    & Monotopic membrane proteins             & Bitopic                                 & 1*  & 1  & 1  \\
2019 & 6MJH & Bitopic proteins                & Transmembrane proteins:alpha-helical    & Transmembrane proteins:alpha-helical    & Bitopic                                 & 1x4 & 2  & 1  \\
2020 & 6S3S & Transmembrane proteins:alpha-helical & Transmembrane proteins:beta-barrel & Transmembrane proteins:alpha-helical & Transmembrane proteins:alpha-helical & 34  & 4  & 4  \\
2020 & 6S3R & Transmembrane proteins:alpha-helical & Transmembrane proteins:beta-barrel & Transmembrane proteins:alpha-helical & Transmembrane proteins:alpha-helical & 36  & 4  & 4  \\
2020 & 7K7A & Bitopic proteins                & Transmembrane proteins:alpha-helical    & Transmembrane proteins:alpha-helical    & Bitopic                                 & 1x3 & 1  & 1  \\
2020 & 7BV6 & Bitopic proteins                & Transmembrane proteins:alpha-helical    & Monotopic membrane proteins             & Bitopic                                 & 1*  & 1  & 1  \\
2020 & 6NV1 & Bitopic proteins                & Transmembrane proteins:alpha-helical    & Transmembrane proteins:alpha-helical    & Bitopic                                 & 1x4 & 1  & 1  \\
2020 & 6PVR & Bitopic proteins                & Transmembrane proteins:alpha-helical    & Transmembrane proteins:alpha-helical    & Bitopic                                 & 1x4 &    &    \\
2020 & 7K3G & Bitopic proteins                & Transmembrane proteins:alpha-helical    & Transmembrane proteins:alpha-helical    & Bitopic                                 & 1x5 & 1  & 1  \\
2020 & 6Z0G & Bitopic proteins                & Transmembrane proteins:alpha-helical    & Monotopic membrane proteins             & Bitopic                                 & 1   & 1  & 1  \\
2021 & 7OFM & Bitopic proteins                & Monotopic membrane proteins             & Transmembrane proteins:alpha-helical    & Bitopic                                 & 1   & 1  & 0  \\
2021 & 7AGX & Transmembrane proteins:alpha-helical & Transmembrane proteins:beta-barrel & Transmembrane proteins:alpha-helical & Transmembrane proteins:alpha-helical & 34  & 4  & 4  \\
2021 & 7OKN & Transmembrane proteins:alpha-helical & Transmembrane proteins:beta-barrel & Transmembrane proteins:beta-barrel & Bitopic                                 & 1x17& 3  & 1  \\
2021 & 7KN0 & Bitopic proteins                & Transmembrane proteins:alpha-helical    & Transmembrane proteins:alpha-helical    & Bitopic                                 & 1x2 & 1  & 1  \\
2021 & 7LQ6 & Bitopic proteins                & Monotopic membrane proteins             & Transmembrane proteins:alpha-helical    & Bitopic                                 & 1   & 1  & 1  \\
2021 & 6LKD & Bitopic proteins                & Transmembrane proteins:alpha-helical    & Transmembrane proteins:alpha-helical    & Bitopic                                 & 1x2 & 0  & 1  \\
2022 & 7RSL & Transmembrane proteins:alpha-helical & Monotopic membrane proteins         & Transmembrane proteins:alpha-helical & Transmembrane proteins:alpha-helical & 2   & 2  & 2  \\
2022 & 8A1D & Monotopic membrane proteins     & Transmembrane proteins:beta-barrel      & Transmembrane proteins:alpha-helical    & Transmembrane proteins:beta-barrel & 64  & 2  & 1  \\
2022 & 7WSO & Bitopic proteins                & Transmembrane proteins:alpha-helical    & Transmembrane proteins:alpha-helical    & Bitopic                                 & 1x4 & 1  & 1  \\
2022 & 7XT6 & Bitopic proteins                & Transmembrane proteins:alpha-helical    & Transmembrane proteins:alpha-helical    & Bitopic                                 & 1x4 & 1  & 1  \\
2022 & 7XQ8 & Bitopic proteins                & Transmembrane proteins:alpha-helical    & Transmembrane proteins:alpha-helical    & Bitopic                                 & 1x4 & 1  & 1  \\
2022 & 7W2B & Bitopic proteins                & Transmembrane proteins:alpha-helical    & Transmembrane proteins:alpha-helical    & Bitopic                                 & 1x3 & 1  & 0  \\
2022 & 7FJD & Bitopic proteins                & Transmembrane proteins:alpha-helical    & Transmembrane proteins:alpha-helical    & Bitopic                                 & 1x8 & 1  & 1  \\
2022 & 7VU5 & Bitopic proteins                & Transmembrane proteins:alpha-helical    & Transmembrane proteins:alpha-helical    & Bitopic                                 & 1x2 & 1  & 1  \\
2022 & 7MPA & Bitopic proteins                & Transmembrane proteins:alpha-helical    & Transmembrane proteins:alpha-helical    & Bitopic                                 & 1   & 1  & 1  \\
2023 & 8GI1 & Bitopic proteins                & Transmembrane proteins:beta-barrel      & Monotopic membrane proteins             & Bitopic                                 & 1x6 & 1  & 1  \\

\end{longtable}

\begin{figure}[htbp]
\centering
\includegraphics[width=.8\textwidth]{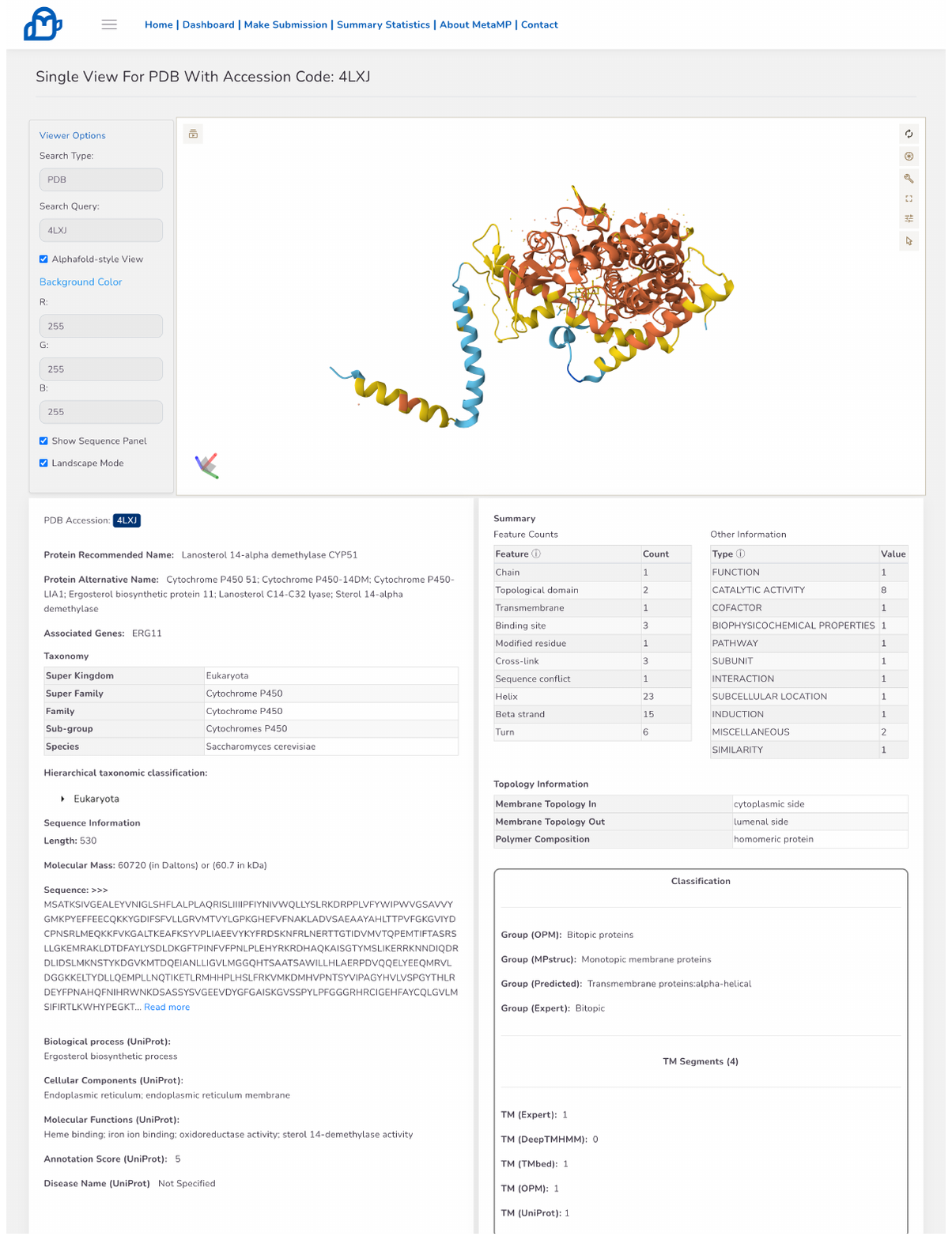}
    \caption{\textbf{Single-Entry Structural view}. This view displays a combined representation of protein structural, functional, and sequence information. The central panel shows a 3D molecular structure rendered as a ribbon diagram, highlighting secondary structure elements and overall folding. Summary tables on the right provide a count of annotated structural features (e.g.,~signal peptides, transmembrane regions, domains, active sites, secondary structure elements) and functional annotations (e.g.,~catalytic activity, cofactors, subunit composition, subcellular localization). Additional panels summarize the protein’s taxonomy, sequence characteristics, topology (membrane orientation and polymer composition), and curated biological annotations, including molecular functions, biological processes, and cellular components. This interface integrates diverse annotations to facilitate comprehensive interpretation of protein features within structural and biological contexts.}
    \label{figure:singleview}
\end{figure}